\documentclass[12pt, a4paper]{report}
\usepackage[a4paper, total={6.25in, 8.25in}]{geometry} 
\usepackage[T1]{fontenc}
\usepackage[latin1]{inputenc}
\usepackage[english]{babel}
\usepackage{siunitx}
\usepackage{graphicx}
\usepackage{tipa} 
\usepackage{graphics} 
\usepackage{amsmath}
\usepackage{latexsym}
\usepackage{amscd}
\usepackage{mathrsfs} 
\usepackage{relsize}
\usepackage{delarray}

\usepackage{graphicx} 

\usepackage{datatool}

\usepackage{xcolor} 
\usepackage[nottoc]{tocbibind} 

\usepackage{glossaries}

\usepackage{pstricks}
\usepackage{theorem}
\usepackage{changepage}
\usepackage{euscript}
\usepackage{textcomp}
\usepackage{esvect}
\usepackage{parskip}
\usepackage{placeins}
\usepackage{subcaption}
\usepackage{array}
\usepackage{delarray}
\usepackage{stmaryrd}
\usepackage{fancyhdr}
\usepackage{graphpap}
\usepackage{makeidx}
\usepackage{enumerate}
\usepackage{esint}
\usepackage{datetime}
\usepackage{caption}
\usepackage{smartdiagram}
\usesmartdiagramlibrary{additions}
\usepackage{abstract}
\usepackage{color}
\definecolor{igreen}{rgb}{0.0, 0.56, 0.0}
\usepackage{xcolor, colortbl}
\colorlet{gred}{-red!75!green!65!}
\colorlet{mamber}{-red!75!green!15!blue!50!}
\colorlet{grown}{-red!75!blue!20!green}
\colorlet{bled}{-red!85!blue!40!green!45!}
\colorlet{waters}{cyan!25} 
\colorlet{water}{cyan!25!green!20!} 
\definecolor{grin}{HTML}{00F9DE}
\usepackage{rotating}
\providecommand{\keywords}[1]{\textbf{\textit{Keywords---}} #1}

\usepackage{arydshln}
\usepackage{booktabs}
\usepackage{graphicx}
\usepackage{tikz}
\usepackage{tikz-3dplot}
\usetikzlibrary{
arrows,
arrows.meta,
automata,
backgrounds,
calc,
decorations,
decorations.pathmorphing,
decorations.pathreplacing,
decorations.fractals,
external,
fit,
matrix,
petri,
positioning,
shadows,
shapes,
shapes.multipart,
topaths,
intersections
}
\usepackage{eso-pic}
\def\ba{\begin{array}}
\def\ea{\end{array}}
\def\beann{\begin{eqnarray*}}
\def\eeann{\end{eqnarray*}}
\def\bea{\begin{eqnarray}}
\def\eea{\end{eqnarray}}

\makeatletter
\newlength\qvec@height
\newlength\qvec@depth
\newlength\qvec@width
\newcommand{\qvec}[2][]{
    \settoheight{\qvec@height}{$#2$}
    \settodepth{\qvec@depth}{$#2$}
    \settowidth{\qvec@width}{$#2$}
  \def\qvec@arg{#1}
  \raisebox{.2ex}{\raisebox{\qvec@height}{\rlap{%
    \kern.05em
    \begin{tikzpicture}[scale=1,shorten >=-3pt,shorten <=-3pt]
    \pgfsetroundcap
    \coordinate (Stx) at (.05em,0) ;
		\coordinate (Arx) at (\qvec@width-.05em,0) ;
    \draw[->](Stx) to[bend left] (Arx);
    \end{tikzpicture}
  }}}
  #2
}
\makeatother
\makeatletter
\newlength\pvec@height
\newlength\pvec@depth
\newlength\pvec@width
\newcommand{\pvec}[2][]{
    \settoheight{\pvec@height}{$#2$}
    \settodepth{\pvec@depth}{$#2$}
    \settowidth{\pvec@width}{$#2$}
  \def\pvec@arg{#1}
  \raisebox{.2ex}{\raisebox{\pvec@height}{\rlap{%
    \kern.05em
    \begin{tikzpicture}[scale=1,shorten >=-3pt,shorten <=-3pt]
    \pgfsetroundcap
    \coordinate (Stx) at (.05em,0) ;
		\coordinate (Arx) at (\pvec@width-.05em,0) ;
    \draw[->](Stx) to[bend right] (Arx);
    \end{tikzpicture}
  }}}
  #2
}
\makeatother
\makeatletter
\newlength\vvec@height%
\newlength\vvec@depth%
\newlength\vvec@width%
\newcommand{\vvec}[2][]{%
  \ifmmode%
    \settoheight{\vvec@height}{$#2$}%
    \settodepth{\vvec@depth}{$#2$}%
    \settowidth{\vvec@width}{$#2$}%
  \else 
    \settoheight{\vvec@height}{#2}%
    \settodepth{\vvec@depth}{#2}%
    \settowidth{\vvec@width}{#2}%
  \fi%
  \def\vvec@arg{#1}%
  \def\vvec@dd{:}%
  \def\vvec@d{.}%
  \raisebox{.2ex}{\raisebox{\vvec@height}{\rlap{%
    \kern.05em%
    \begin{tikzpicture}[scale=1]
    \pgfsetroundcap
    \draw (.05em,0)--(\vvec@width-.05em,0);
    \draw (\vvec@width-.05em,0)--(\vvec@width-.15em, .075em);
    \draw (\vvec@width-.05em,0)--(\vvec@width-.15em,-.075em);
    \ifx\vvec@arg\vvec@d%
      \fill(\vvec@width*.45,.5ex) circle (.5pt);%
    \else\ifx\vvec@arg\vvec@dd%
      \fill(\vvec@width*.30,.5ex) circle (.5pt);%
      \fill(\vvec@width*.65,.5ex) circle (.5pt);%
    \fi\fi%
    \end{tikzpicture}%
  }}}%
  #2%
}
\makeatother
\def\ba{\begin{array}}
\def\ea{\end{array}}
\def\beann{\begin{eqnarray*}}
\def\eeann{\end{eqnarray*}}
\def\bea{\begin{eqnarray}}
\def\eea{\end{eqnarray}}

\usepackage{titlesec}
\usepackage{multirow}
\usepackage{hyperref}
\usepackage{lipsum}
\usepackage{tikz-cd}
\usepackage{float}
\usepackage{titling}
\usepackage{epigraph}
\usepackage[title, titletoc]{appendix}
\titleformat{\chapter}{\normalfont\LARGE}{\thechapter\,$\vert$}{20pt}{\LARGE}{\setcounter{chapter}{0}}
\titlespacing*{\chapter}{0pt}{-70pt}{40pt} 
\makeatletter
\newcommand\BackgroundPicturea[3]{
	\setlength{\unitlength}{1pt}
	\put(0,\strip@pt\paperheight){
		\parbox[t]{\paperwidth}{
			\vspace{#2}\hspace{#3}
			\mbox{\includegraphics[scale=0.5]{#1}}
}}}
\newcommand\BackgroundPictureb[3]{
	\setlength{\unitlength}{1pt}
	\put(0,\strip@pt\paperheight){
		\parbox[t]{\paperwidth}{
			\vspace{#2}\hspace{#3}
			\mbox{\includegraphics[scale=0.3]{#1}}
}}}
\makeatother

\usepackage{setspace}
\addto\captionsenglish{
	\renewcommand{\contentsname}%
	{Table of Contents}
	\setcounter{tocdepth}{3}
	\setcounter{secnumdepth}{3}
	}

\usepackage{xcolor}
\usepackage{listings}
\usepackage{xcolor}

\definecolor{codegreen}{rgb}{0,0.6,0}
\definecolor{codegray}{rgb}{0.5,0.5,0.5}
\definecolor{codepurple}{rgb}{0.58,0,0.82}
\definecolor{backcolour}{rgb}{0.95,0.95,0.92}

\lstdefinestyle{mystyle}{
  backgroundcolor=\color{backcolour}, commentstyle=\color{codegreen},
  keywordstyle=\color{magenta},
  numberstyle=\tiny\color{codegray},
  stringstyle=\color{codepurple},
  basicstyle=\ttfamily\footnotesize,
  breakatwhitespace=false,         
  breaklines=true,                 
  captionpos=b,                    
  keepspaces=true,                 
  numbers=left,                    
  numbersep=5pt,                  
  showspaces=false,                
  showstringspaces=false,
  showtabs=false,                  
  tabsize=2
}

\usepackage{amsfonts}
\newcommand{\R}{\mathbb{R}}

\newcommand{\I}{\mathbb{I}}

\newcommand{\sortitem}[1]{%
  \DTLnewrow{list}
  \DTLnewdbentry{list}{description}{#1}
}
\newenvironment{sortedlist}{%
  \DTLifdbexists{list}{\DTLcleardb{list}}{\DTLnewdb{list}}
}{%
  \DTLsort{description}{list}
  \begin{itemize}%
    \DTLforeach*{list}{\theDesc=description}{%
      \item[] \theDesc}
  \end{itemize}%
}

\usepackage{algorithm}
\usepackage{algpseudocode}

\usepackage{pgfgantt}

\usepackage{makecell} 

\usepackage{lscape}

\usepackage{verbatim}

\usepackage{multicol}
\hypersetup{pdftitle = Project Report, pdfauthor = {First Last}, pdfstartview=FitH, pdfkeywords = essay, pdfpagemode = FullScreen, colorlinks, anchorcolor = black, citecolor = blue, urlcolor = blue, filecolor = green, linkcolor = blue, plainpages = false}
\fancyheadoffset{8pt}
\fancyfootoffset{8pt}

\makeglossaries

\newglossaryentry{latex}
{
    name=Latex,
    text=latex,
    description={Is a markup language specially suited 
    for scientific documents, this term is printed in conclusion }
}
\newglossaryentry{raster}
{
    name=Raster,
    text=raster,
    description={ images are compiled using pixels, or tiny dots, containing unique color and tonal information that come together to create the image }
}
\newglossaryentry{gradient descent}
{
    name=Gradient descent,
    text=gradient descent,
    description={Is a naive optimization method which consists of steepest descent down the gradient of the given cost function}
}

\newglossaryentry{Gauss-Newton}
{
    name=Gauss-Newton,
    description={Is a Newton-like method for solving a non-linear least squares problem, in which the Hessian $H$ is approximated by $H \approx J^T WJ$, where $J$ is the design matrix and $W$ is the weights. The normal equations are the resulting prediction equations given as \\ $(J^TWJ) \delta x = -(JW \Delta z)$}
}

\newglossaryentry{Conjugate Gradient}
{
    name=Conjugate Gradient,
    text=conjugate gradient,
    description={Is an accelerated first order iterative process for solving positive definite linear systems or minimizing a non linear cost function.}
}

\newglossaryentry{Jacobian}
{
    name=Jacobian,
    description={Matrix of partial differentials of the cost function $J = \frac{df}{dx}$}
}

\newglossaryentry{Hessian}
{
    name=Hessian,
    description={Matrix of second partial differentials of the cost function $H = \frac{d^2f}{dx^2}$}
}
\newglossaryentry{Gradient}
{
    name=Gradient,
    description={First differential $g = \frac{df}{dx}$}
}
\newglossaryentry{epipolar plane}
{
    name=epipolar plane,
    description={The plane containing the intersection line joining the camera centres with the image plane.}
}
\newglossaryentry{linear least squares}
{ 
    name =Linear Least Squares,
    description ={Least squares approximation of linear functions to data, by minimizing residuals.
     $E_{LS} = \sum_i{||\hat{x'_i}- \tilde{x'_i}||}$}
}
\newglossaryentry{RANdom Sample Consensus (RANSAC)}
{
    name = Random Sample Consensus (RANSAC),
    description = {An iterative method to estimate parameters of a mathematical model from a set of observed data which contains outliers.}
}

\date{September 2022}

\title{Semantic Validation in Structure from Motion}
\author{\\ \Large{Joseph Rowell}
\\ Supervisors: Prof. Simon Julier, Ziwen Lu
\\ Faculty of Engineering
\\ Department of Computer Science
\\ 
\\
\\ University College London
\\
A Project Report Presented in Partial Fulfillment of the Degree \\ \textit{MSc Robotics and Computation}
\\ \\
}
\begin{document}
\AddToShipoutPictureBG*{%
  \AtPageUpperLeft{%
    \raisebox{-\height}{%
      \includegraphics[width=\paperwidth]{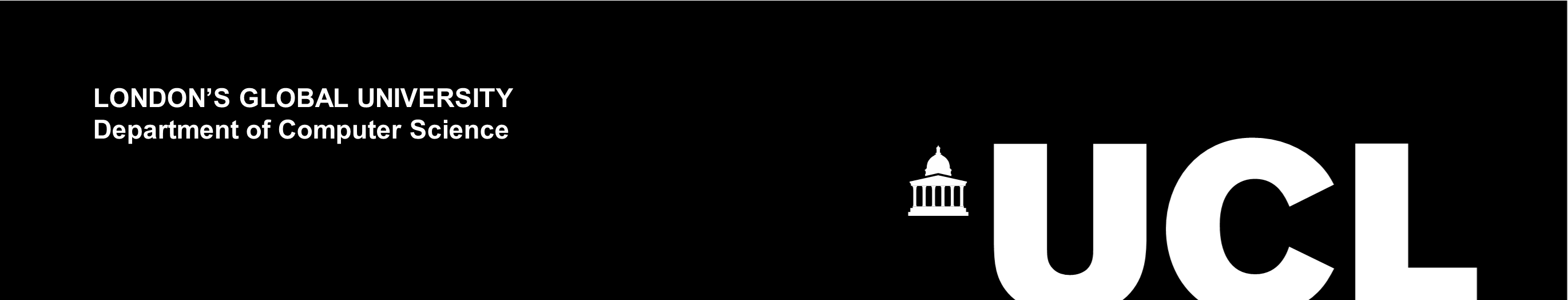}%
    }}
}
\AddToShipoutPicture*{%
      \parbox[t][\paperheight][t]{\paperwidth}{%
          \includegraphics[width=1.2\paperwidth]{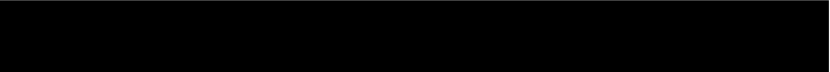}
      }}

\thispagestyle{headings}
\maketitle
\FloatBarrier
\pagenumbering{roman}

\thispagestyle{empty}
\begin{abstract}

The Structure from Motion (SfM) challenge in computer vision is the process of recovering the 3D structure of a scene from a series of projective measurements that are calculated from a collection of 2D images, taken from different perspectives. SfM consists of three main steps; feature detection and matching, camera motion estimation, and recovery of 3D structure from estimated intrinsic and extrinsic parameters and features. 

A problem encountered in SfM is that scenes lacking texture or with repetitive features can cause erroneous feature matching between frames.
Semantic segmentation offers a route to validate and correct SfM models by labelling pixels in the input images with the use of a deep convolutional neural network. The semantic and geometric properties associated with classes in the scene can be taken advantage of to apply prior constraints to each class of object. 
The SfM pipeline COLMAP and semantic segmentation pipeline DeepLab were used. This, along with planar reconstruction of the dense model, were used to determine erroneous points that may be occluded from the calculated camera position, given the semantic label, and thus prior constraint of the reconstructed plane.  Herein, semantic segmentation is integrated into SfM to apply priors on the 3D point cloud, given the object detection in the 2D input images. Additionally, the semantic labels of matched keypoints are compared and inconsistent semantically labelled points discarded. Furthermore, semantic labels on input images are used for the removal of objects associated with motion in the output SfM models. The proposed approach is evaluated on a data-set of 1102 images of a repetitive architecture scene. This project offers a novel method for improved validation of 3D SfM models.


\end{abstract}

\newpage
\thispagestyle{empty}
\vspace*{\fill}
\begin{center}
Copyright \copyright  \thinspace 2022 by Joseph Rowell \\ All Rights Reserved
\end{center}
\vspace*{\fill}
\newpage
\thispagestyle{empty}
\epigraph{Sizzling Saturn, we've got a lunatic robot on our hands.}{--- \textup{Isaac Asimov, I, Robot}}

\thispagestyle{empty}
\chapter*{Acknowledgements}
I would like to express my gratitude towards Professor Simon Julier and Ziwen Lu for their guidance and feedback throughout the project, and Ziwen Lu again for the acquisition of the Brighton data-set, Simon again for the use of his GPU.
I would also like to acknowledge Oliver Kingshott for his help and guidance in the project.

\thispagestyle{empty}
\chapter*{Disclaimer}
This report is submitted as part requirement for the MSc in Robotics and Computation at University College London. It is substantially the result of my own work except where explicitly indicated in the text. The report may be freely copied and distributed provided the source is explicitly acknowledged.
\begin{figure}[H]
\includegraphics{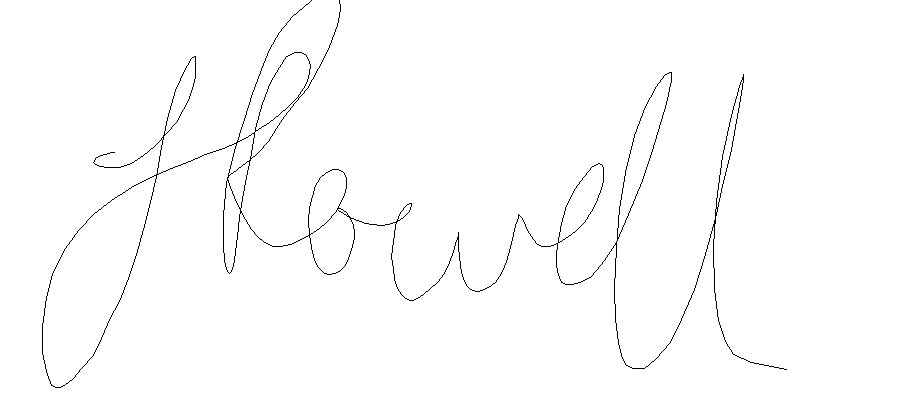}
\end{figure}
\vspace{-2cm}
\noindent\begin{tabular}{ll}
 & 12/09/22 \\
\makebox[2.5in]{\hrulefill} & \makebox[2.5in]{\hrulefill}\\
\textit{Signature} & \textit{Date}\\
\end{tabular}

\tableofcontents

\thispagestyle{plain}
\listoffigures
\listoftables
\listofalgorithms

\chapter*{List of Abbreviations}
\begin{sortedlist} 
  \sortitem{CNN:     Convolutional Neural Network}
  \sortitem{DCNN:    Deep Convolutional Neural Network}
  \sortitem{SLAM:    Simultaneous Localization And Mapping}   \sortitem{GNSS:    Global Navigation Satellite System}
  \sortitem{vSLAM:   visual Simultaneous Localisation and Mapping}
  \sortitem{MAP:     Maximum a Posteriori}
  \sortitem{ML:      Maximum Likelihood}
  \sortitem{GPS:     Global Positioning System}
  \sortitem{BA:      Bundle Adjustment}
  \sortitem{TRC:     TPU Research Cloud}
  \sortitem{TPU:     Tensor Processing Unit}
  \sortitem{SfM:     Structure from Motion}
  \sortitem{MVS:     Multi-view Stereo}
  \sortitem{SIFT:    Scale Invariant Feature Transform}
  \sortitem{SURF:    Speeded-Up Robust Features}
  \sortitem{ORB:     Oriented FAST Rotated BRIEF}
  \sortitem{FAST:    Features from Accelerated Segment Test}
  \sortitem{BRIEF:   Binary Robust Independent Elementary Features}
  \sortitem{RANSAC:  Random Sample Consensus}
  \sortitem{SSfM:    Semantic Structure from Motion}
  \sortitem{ISO:     International Standardization Organisation}
  \sortitem{DOG:     Difference Of Gaussians}
  \sortitem{YOLO:    You Only Look Once}
  \sortitem{GUI:     Graphical User Interface}
  \sortitem{CLI:     Command Line Interface}
  \sortitem{mAP:     Mean Average Precision}
  \sortitem{DOF:     Degrees Of Freedom}
  \sortitem{SDF:     Signed Distance Function }   
  \sortitem{SVD:     Singular Value Decomposition}
  \sortitem{EXIF:    Exchangeable Image File Format}
  \sortitem{IMU:     Inertial Measurement Unit}
  \end{sortedlist}
\chapter{Project Plan} \label{Chap1}
\section{Problem}
\begin{figure}[H]
     \centering
     \begin{subfigure}{0.33\textwidth}
         \centering
         \includegraphics[width=\textwidth]{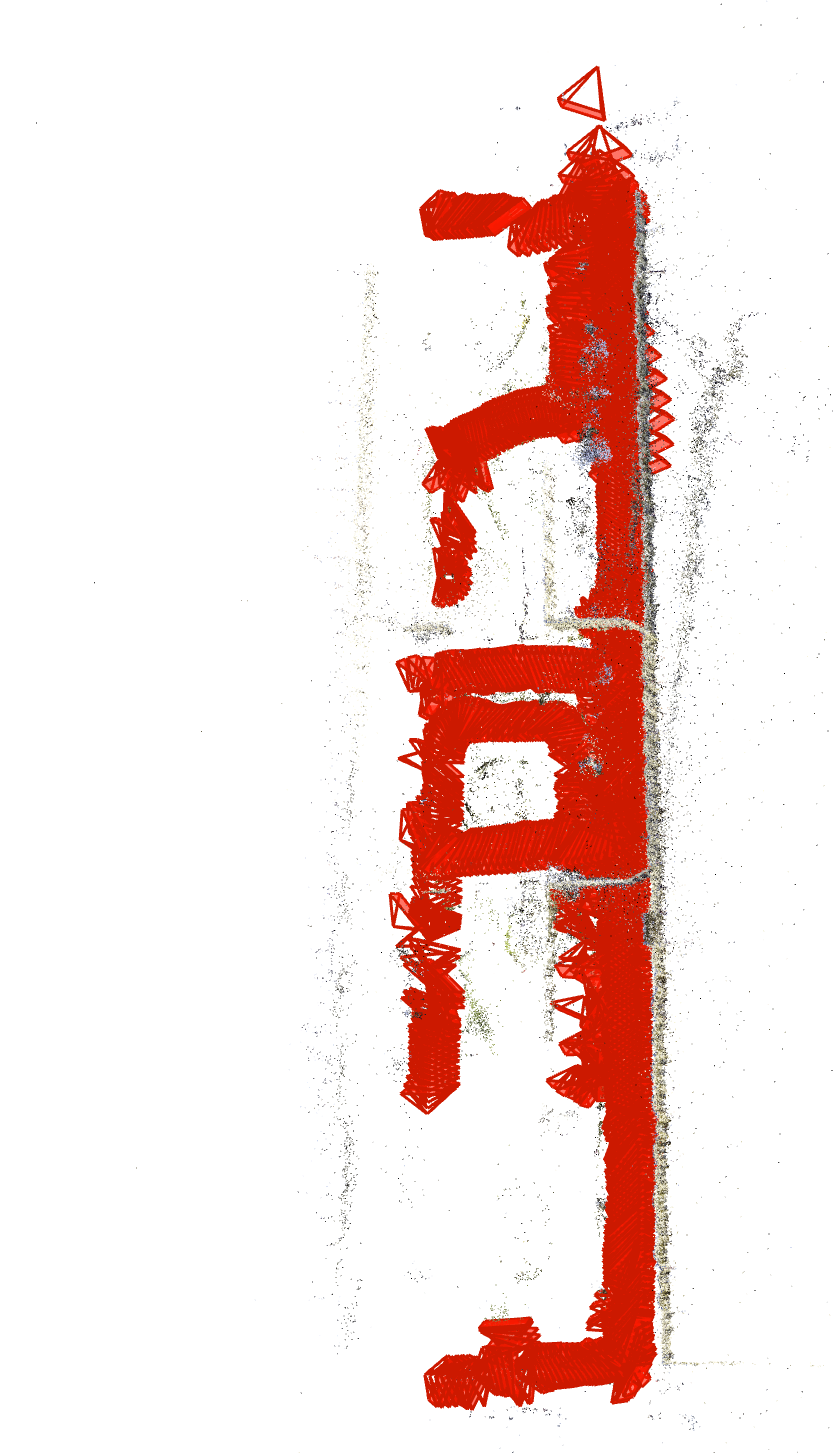}
         \caption{Reconstruction 1}
         \label{fig:recon1}
     \end{subfigure}
     \begin{subfigure}{0.33\textwidth}
         \centering
         \includegraphics[width=\textwidth]{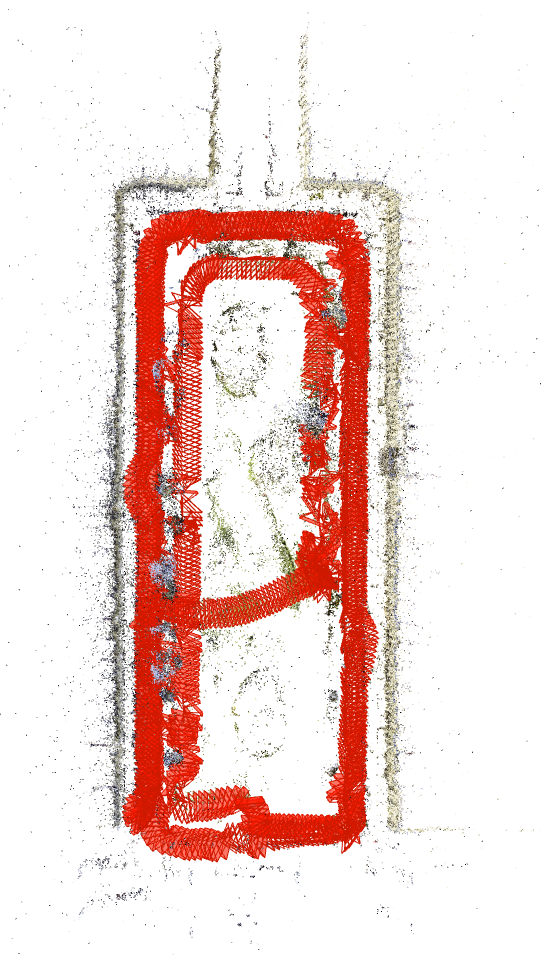}
         \caption{Reconstruction 2}
         \label{fig:recon2}
     \end{subfigure}
     \begin{subfigure}{0.3\textwidth}
         \centering
         \includegraphics[width=\textwidth]{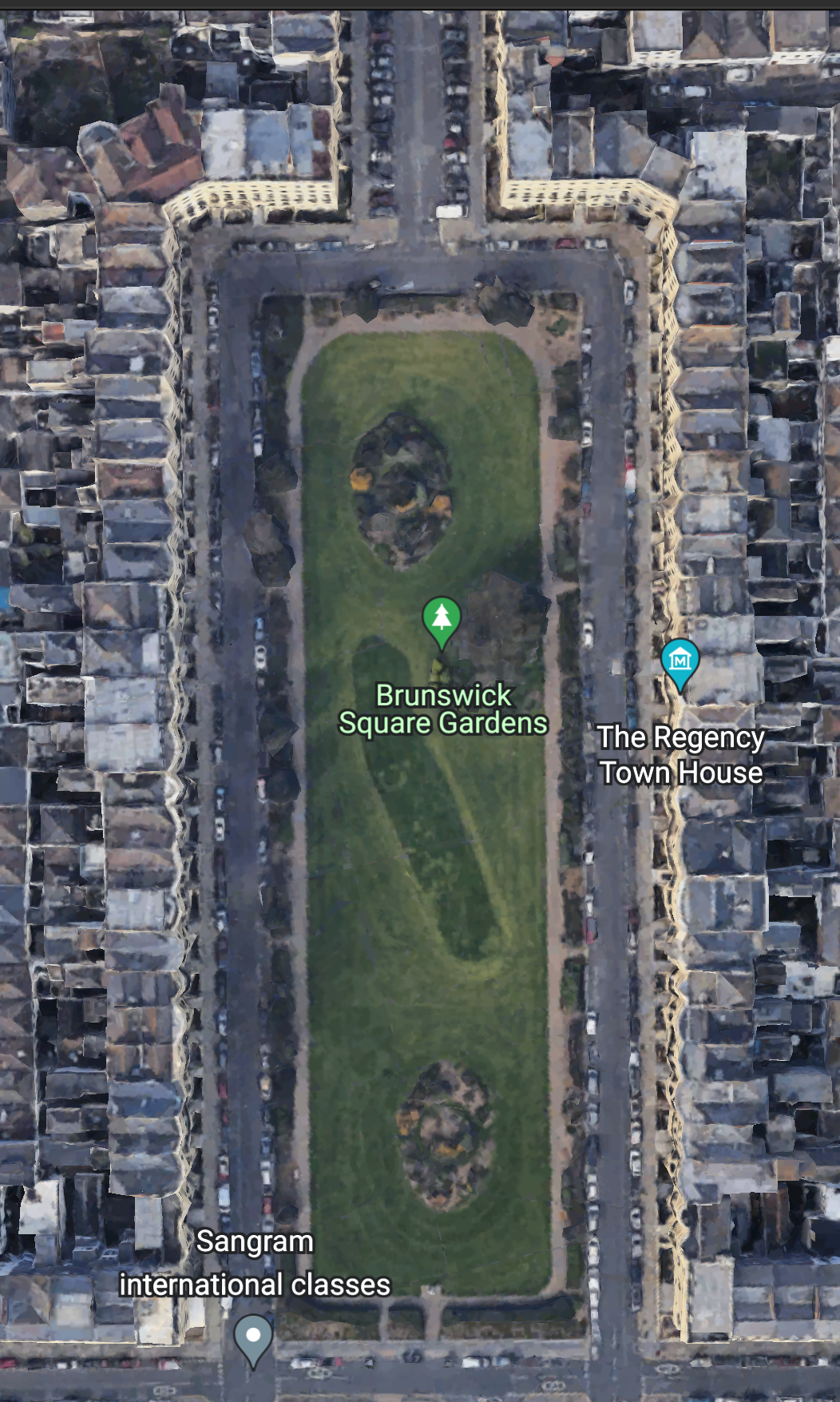}
         \caption{Brunswick Square aerial view}
         \label{fig:aerialview}
     \end{subfigure}
    \caption[Brunswick Square, Brighton. Erroneous feature matching.]{COLMAP SfM sparse reconstructions of Brunswick Square, Brighton top down views. Reconstruction 1 shows poor performance due to erroneous feature matching between frames, causing both sides of the square to be mapped onto one side. Reconstruction 2 shows a \emph{better} sparse point cloud reconstruction. Aerial view of Brunswick Square \cite{GoogleGoogleEarth} shows the ground truth.}
        \label{fig:brunswicksquareoverview}
\end{figure}

In the field of computer vision and visual perception, Structure from Motion (SfM) is a photogrammetric imaging technique for estimating three-dimensional structures from two-dimensional image sequences through matching features between frames \cite{Ullman1979TheMotion}. The feature matching between repetitive features can be erroneous, and cause loop closure errors. Semantic, topologic and geometric information in the images can be used to (in)validate the 3D SfM model. An example of the result of erroneous feature matching can be seen in Fig. \ref{fig:brunswicksquareoverview}, where Brunswick Square, Brighton has repetitive architecture and has caused two sides of the Square to be mapped onto the same side. There is also great variability in the output of SfM software COLMAP, with the same input data, as can be observed in the differences between Fig. \ref{fig:recon1} and Fig. \ref{fig:recon2}, thus an autonomous validation and correction pipeline is necessary.

\section{Work Plan}
\begin{enumerate}
    \item Literature review and review of state-of-the-art in SfM and semantic segmentation
    \item Running SfM system on a repetitive feature data-set and determining erroneous measurements and scenarios
    \item Develop Python based solution
    \item Test and validate on repetitive feature data-set
    \item Develop plugin for open-source SfM system COLMAP
    \item Develop real-time solution to correct and validate erroneous models 
\end{enumerate}

\section{Deliverables}
The project will focus on correcting erroneous measurements in SfM loop closure in the case of repetitive or similar visual features. The project will be conducted using COLMAP, and initially solve the problem as a Structure for Motion (SfM) problem, without consideration for real time SLAM. A plugin will be developed and validated against a data-set with repeating visual structure (e.g. brick wall or repetitive architecture). By incorporating semantic understanding into SfM, we can improve the 3D models robustness and correctness. The approach taken will be to evaluate the semantic consistency between matched feature points in two view geometries, and to ray trace between cameras and points to determine opaque semantic labelled planar occlusions. In addition, motion removal in SfM will be explored.

\section{Evaluation}
The project requires video data-sets of public places with repetitive architecture and features, and either open-source benchmark data-sets or self-acquired data-sets.
  
  

\chapter{Introduction and Background} \label{Chap2}
\pagenumbering{arabic}
The problem of estimating camera pose and reconstructing three dimensional models of the surrounding environment has drawn significant attention over the last few decades \cite{Risqi201837Survey}. The main techniques for solving this problem are Structure from Motion (SfM) and Visual Simultaneous Localisation and Mapping (vSLAM). Standard SfM and vSLAM aim to estimate 3D structure of the scene and corresponding camera poses through feature correspondences observed in multiple images. The estimation problem in these methods can be solved by filter-based approaches such as the Kalman Filter \cite{Risqi201837Survey}, or by bundle adjustment. This project will aim to improve upon the state-of-the-art in a SfM problem that uses bundle adjustment, by integrating semantic understanding of the environment.
Conventional Structure from Motion (SfM) addresses the problem of recovering intrinsic and extrinsic camera parameters (motion) and 3D locations (structure) of detected feature points \cite{Bao2011SemanticMotion}, without prior knowledge of location from sensors like Global Positioning System (GPS), Global Navigation Satellite System (GNSS) or Inertial Measurement Units (IMU). The images are acquired by sensors such as RGB-D cameras. A problem that can arise from this method of acquisition is noisy data \cite{Holz2012Real-timeCameras}. The combination of data noise, no access to prior information about location, and repetitive similar features in the input images can cause erroneous feature matching between frames. This can cause erroneous loop closure, and render the final model incorrect. The proposed solution to this is to utilise the semantic, geometric and topologic information available in the images, to (in)validate the model. For example, when modelling a structure with repetitive features such as a brick wall, the consistency of the model can be checked against a prior constraint such as - brick wall is planar, and opaque. By using semantic segmentation information, the feature point matching accuracy can be improved.
Assumptions are made in SfM \cite{Agapito2022COMP0130SLAM}, such as there is sufficient (and similar frame by frame) illumination in the environment, the environment has a dominance of static scenes over moving objects, there is enough texture to allow apparent motion to be extracted, and there is high visual overlap between consecutive frames to allow for reconstruction. Furthermore, due to the random element of Random Sample Consensus (RANSAC) in feature matching between frames, the 3D model output can be different for each run given the same input data necessitating a validation step. 
\section{Thesis Outline}
The remainder of this report is organised as follows:
\begin{itemize}
    \item[] \hyperref[Chap2]{\textbf{Chapter 2}} --- Introduces the related work and literature review in preparation for this project.
    \item[] \hyperref[Chap3]{\textbf{Chapter 3}} --- Introduces the Methodology used in creating the semantic validation algorithm, including technical challenges encountered.
    \item[] \hyperref[Chap4]{\textbf{Chapter 4}} --- Introduces the Results and Analysis of the performed method.
    \item[] \hyperref[Chap5]{\textbf{Chapter 5}} --- Introduces the Discussion of the results and possible future work.
    \item[] \hyperref[Chap6]{\textbf{Chapter 6}} --- Concludes and reflects on the project and its outcomes overall.
 \end{itemize}
\section{Background}
\subsection{Structure From Motion} \label{sec:sfm}
Perceiving the 3D semantic and spatial structure of a complex scene from images is a critical capability of an intelligent perception system.
Structure from Motion (SfM) is the process of reconstructing 3D structures from an unordered series of images to determine projections to camera centres. The input of SfM is a set of overlapping images of the same structure, from different perspectives, to output a 3D reconstruction of the structure. Another output of the SfM pipeline is the the reconstructed intrinsic and extrinsic camera parameters of all of the inputted images.
SfM can be divided into three main steps: 
\begin{itemize}
    \item Feature detection and extraction from input images, frame by frame
    \item Feature matching and geometric verification between frames
    \item Structure and motion reconstruction
\end{itemize}
Included in the first stage of SfM is correspondence search, which finds scene overlaps in the input images $\mathbb{I}$, and identifies the projections of the identified points in the overlapping images.
Initial feature detection is performed by an algorithm such as Scale Invariant Feature Transform (SIFT) as used in the COLMAP pipeline. The features identified should be invariant under radiometric and geometric alterations, to allow for SfM to uniquely and accurately recognise and match them. In the case of repetitive architectural features, or texture-less scenes, there can be erroneous feature matching between frames.

\subsubsection{Feature Descriptors} \label{subsubsec:Feature descriptors}
\begin{figure}[H]
    \centering
    \includegraphics[width = \textwidth]{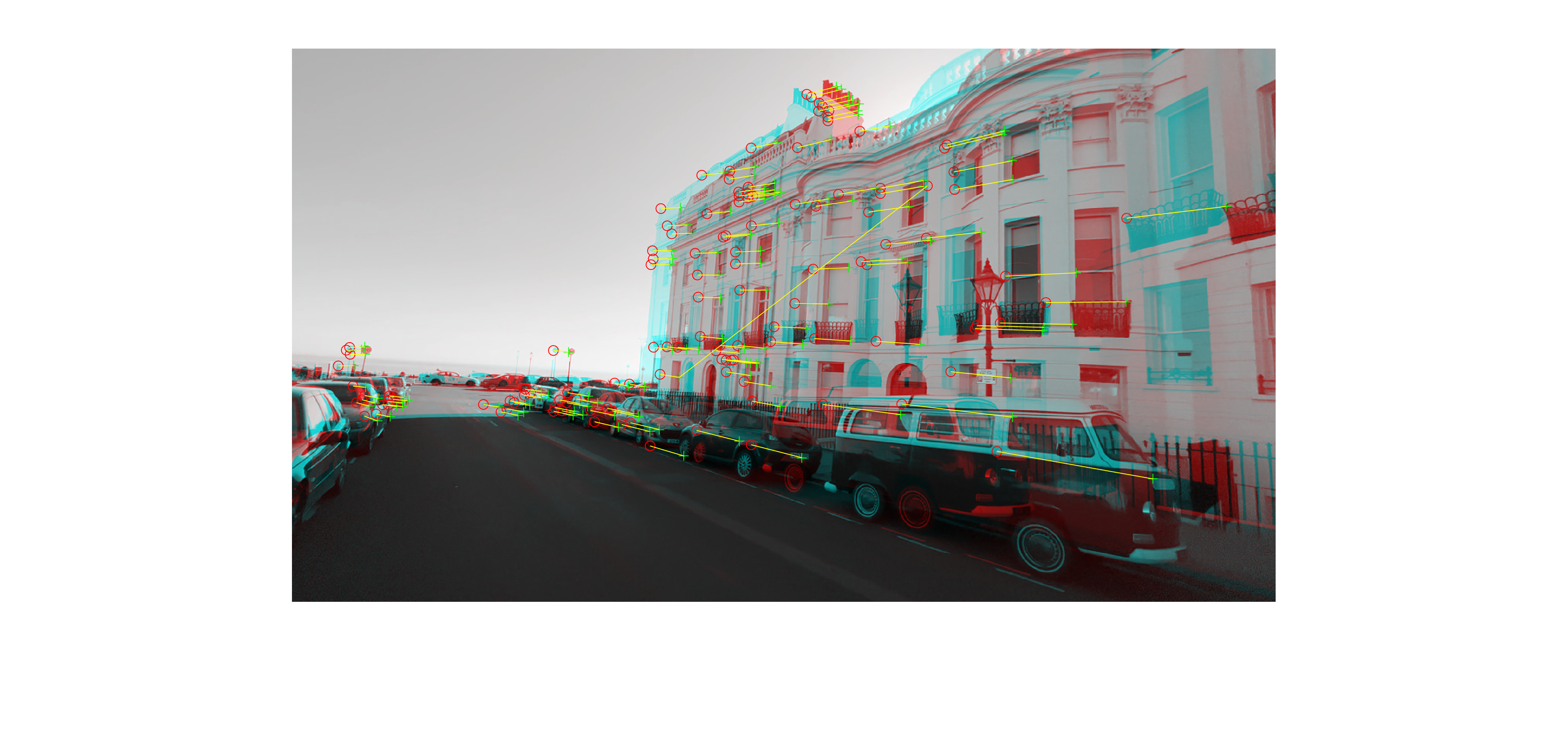}
    \caption{Harris Corner Detection Example on Brighton Data-set}
    \label{fig:HarrisCorner}
\end{figure}

Feature descriptors act as an algorithm that takes images as an input, and outputs feature vectors, acting as a 'fingerprint' to differentiate one feature from another. There are many different algorithms for feature identification such as; Harris corner detection \cite{Derpanis2004TheDetector}, SURF (Speeded Up Robust Feature) \cite{Bay2007Speeded-UpSURF}, FAST (Features from Accelerated Segment Test) \cite{DeepakGeethaViswanathan2011FeaturesFAST}, ORB (Oriented FAST and Rotated BRIEF) \cite{Rublee2011ORB:SURF}, and SIFT (Scale Invariant Feature Transform) \cite{Lowe2004DistinctiveKeypoints}; to name a few. An example of
Harris corner detection is shown in Fig. \ref{fig:HarrisCorner}. 
Extracting a meaningful descriptor from 3D point cloud is a particularly challenging research area, especially in an environment with occlusion, and is worth being investigated extensively.
There are three main categories of descriptors, global-based, local-based and hybrid-based. Global-based generally estimate a single feature descriptor vector that encodes the entire structures geometry, however relies on observation of the full point cloud of the object \cite{Han20183DState-of-the-Art}. This report will focus on local-based feature descriptors, which construct features related to the geometric information of the local neighbourhood of each chosen keypoint.
The SfM pipeline COLMAP uses SIFT feature descriptors, as it has good invariance to scale, rotation and illumination
\cite{Daixian2010SIFTOptimization}. 

\begin{figure}[H]
    \centering
    \includegraphics[width = \textwidth]{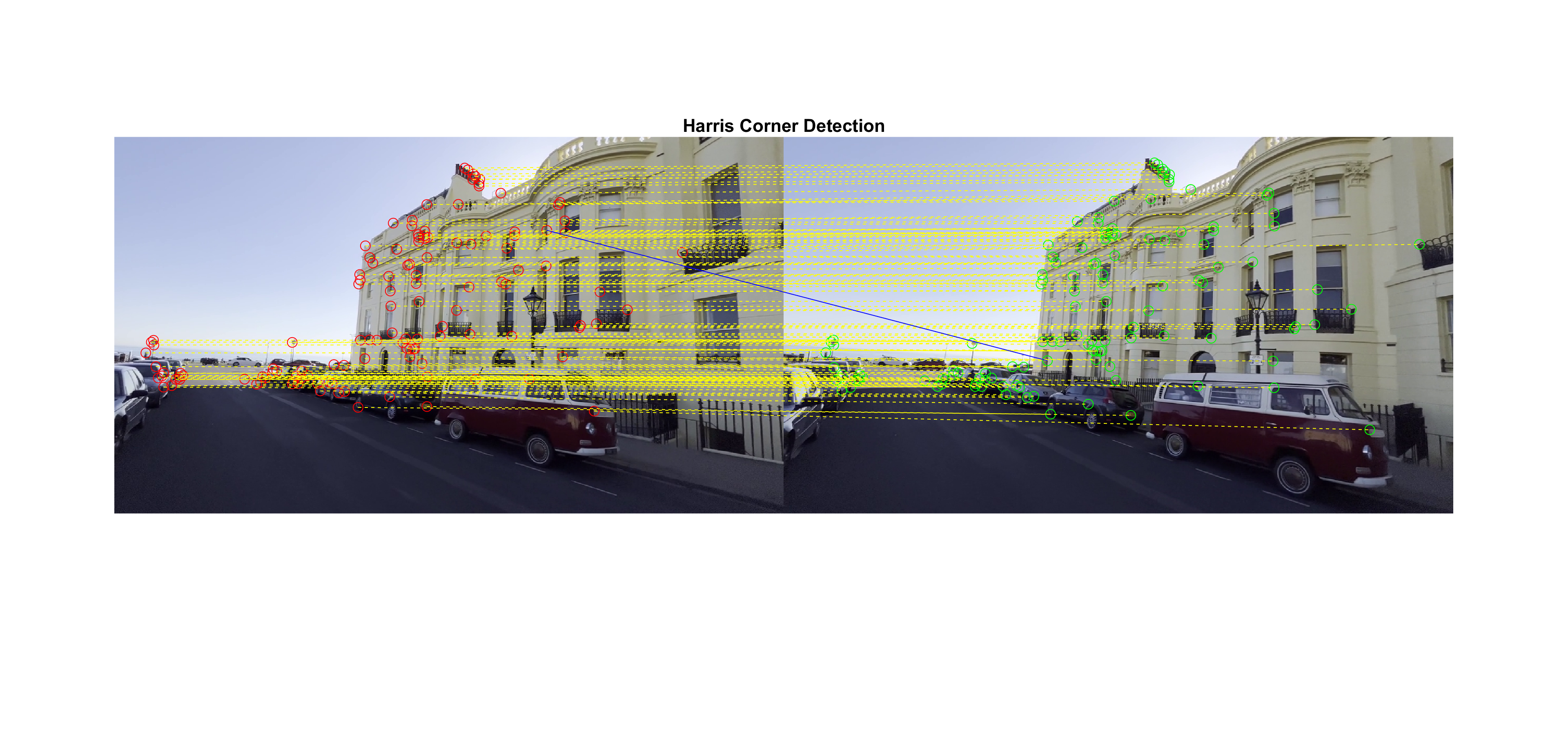}
    \caption{Harris corner detection example on Brighton data-set, inliers plotted in yellow, outliers plotted in blue}
    \label{fig:HarrisCorner2}
\end{figure}

\begin{figure}[H]
    \centering
    \includegraphics[width = 0.8\textwidth]{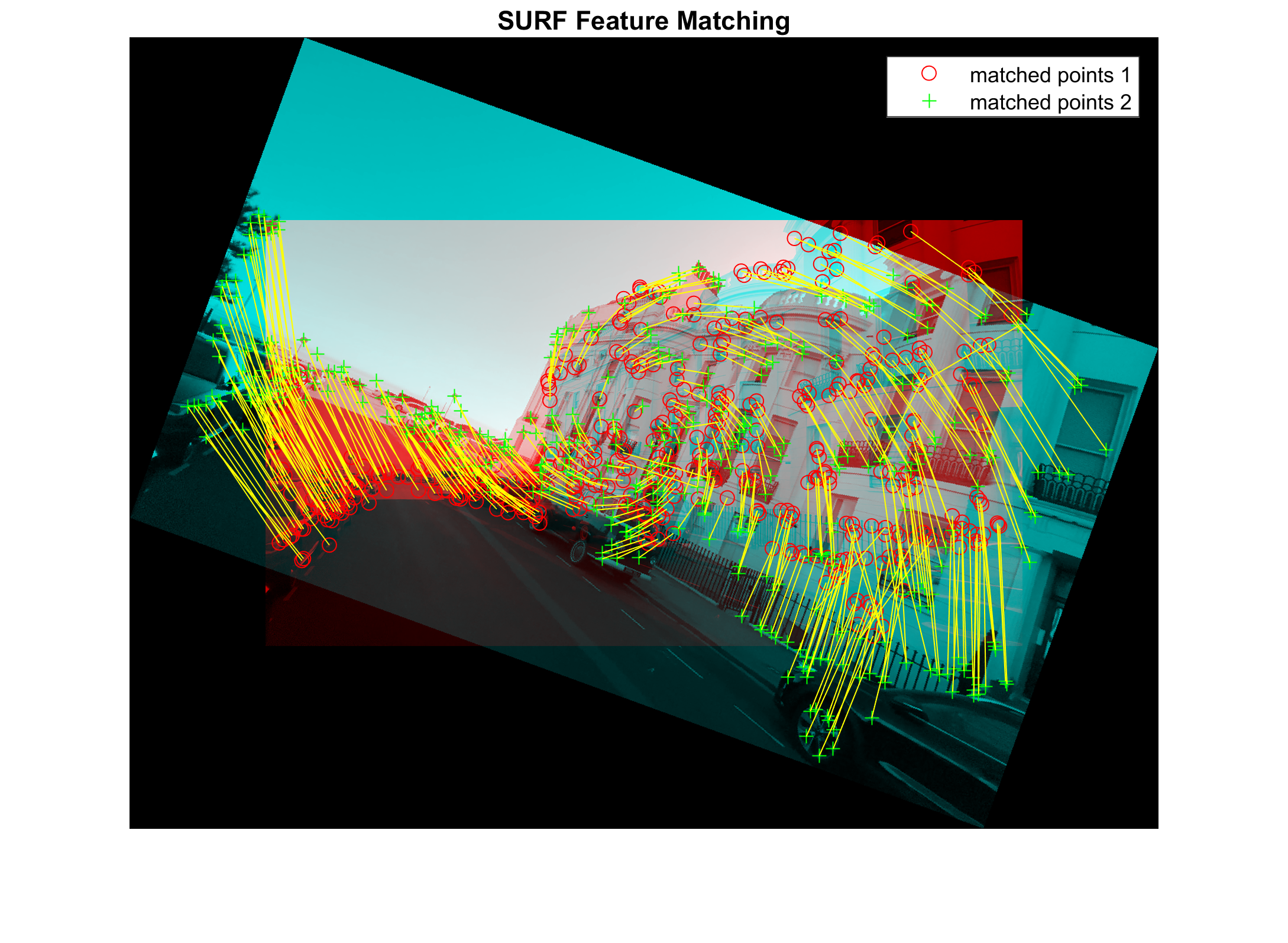}
    \caption{SURF Feature Matching Example on Brighton Data-set, showing Rotation Invariance}
    \label{fig:SURFMatching}
\end{figure}
SIFT is a feature descriptor where the descriptor of a keypoint is a 3D spatial histogram of image gradients, offering sub pixel accuracy \cite{Kupfer2015AnImages}, with rotation and translation invariance. 
There are four steps to extract SIFT features \cite{Lowe2004DistinctiveKeypoints}:
\begin{enumerate}
    \item Scale-Space Generation
    \item Detection of Scale-space Extrema
    \item Keypoint Optimization
    \item SIFT Descriptor Generation
\end{enumerate}

\emph{Scale-Space Generation} \newline 
Scale-Space Generation generates different scale Gaussian filters $G(x,y,\sigma)$, see Eq. \ref{eq:gaussianPyramid}, with a scale factor K of the original image, and $(x,y)$ represent Cartesian position in the image. The filtered output $L(x,y,K\sigma)$, see Eq.  \ref{eq:SIFToutput}, is a Gaussian pyramid image \cite{Daixian2010SIFTOptimization} \cite{Lowe2004DistinctiveKeypoints}.
\begin{equation}\label{eq:gaussianPyramid}
G(x,y,\sigma) = \frac{1}{2 \pi \sigma^2}e^{-\frac{(x^2+y^2)}{2\sigma^2}}
\end{equation}

\begin{equation}\label{eq:SIFToutput}
L(x,y,K\sigma) = G(x,y,K\sigma)*  I(x,y)
\end{equation}
\emph{Detection of Scale-Space Extrema} \newline
This stage of filtering attempts to identify the locations and scales that are identifiable from different views of the same object \cite{SIFTFeatures}. This can be achieved efficiently using a scale space function. Detection of Scale-Space Extrema is implemented using a Difference of Gaussian (DOG) function. This algorithm can identify potential points of interest, which are scale and orientation invariant, hence the \emph{Scale Invariant} Feature Transform. The points are identified in the scale space with the image $D(x,y,\sigma)$, computed by the difference between the sample point and its eight neighbours in the current image, and nine neighbours in the images above and below, Eq. \ref{eq:DOG} \cite{SIFTFeatures}.
\begin{equation} \label{eq:DOG}
    D(x,y,\sigma) = L(x,y,K\sigma) - L(x,y,\sigma)
\end{equation}

\emph{Keypoint Optimisation} \newline
Once a keypoint has been identified by comparison with its pixel neighbours, the subsequent step is to perform a fit to nearby data for location, scale, and ratio of principle curvatures. This process rejects points that are low contrast, and thus more sensitive to noise, or points that are localized poorly along an edge. Brown et al \cite{BrownInvariantGroups} used Taylor expansion to fit a quadratic function to local sample points to determine interpolated location of the maxima, Eq. \ref{eq:taylorexpD}, where $D$ and its derivatives are evaluated at the sample point, $\mathbf{x}$ is the offset from this point $\mathbf{x} = (x,y,\sigma)^\top$.
\begin{equation} \label{eq:taylorexpD}
    D(\mathbf{x}) = D + \frac{\partial D^\top}{\partial \mathbf{x}} \mathbf{x} + \frac{1}{2} x^\top \frac{\partial^2D}{\partial \mathbf{x}^2}\mathbf{x} 
\end{equation}
The location of extrema $\mathbf{\hat{x}}$ is thus calculated by taking the derivative of the function w.r.t $\mathbf{x}$, and setting it to zero, Eq. \ref{eq:derivativeofExtrema}.
\begin{equation}\label{eq:derivativeofExtrema}
    \mathbf{\hat{x}}= -\frac{\partial^2D}{\partial \mathbf{x}^2}^{-1}\frac{\partial D}{\partial \mathbf{x}}
\end{equation}
Extrema can then be evaluated with a threshold on $D(\mathbf{\hat{x}})$, and this rejects unstable extrema with low contrast, achieved by substituting Eq. \ref{eq:derivativeofExtrema} into Eq. \ref{eq:taylorexpD}.

\begin{figure}[H]
     \centering
     \begin{subfigure}{0.49\textwidth}
         \centering
         \includegraphics[width=\textwidth]{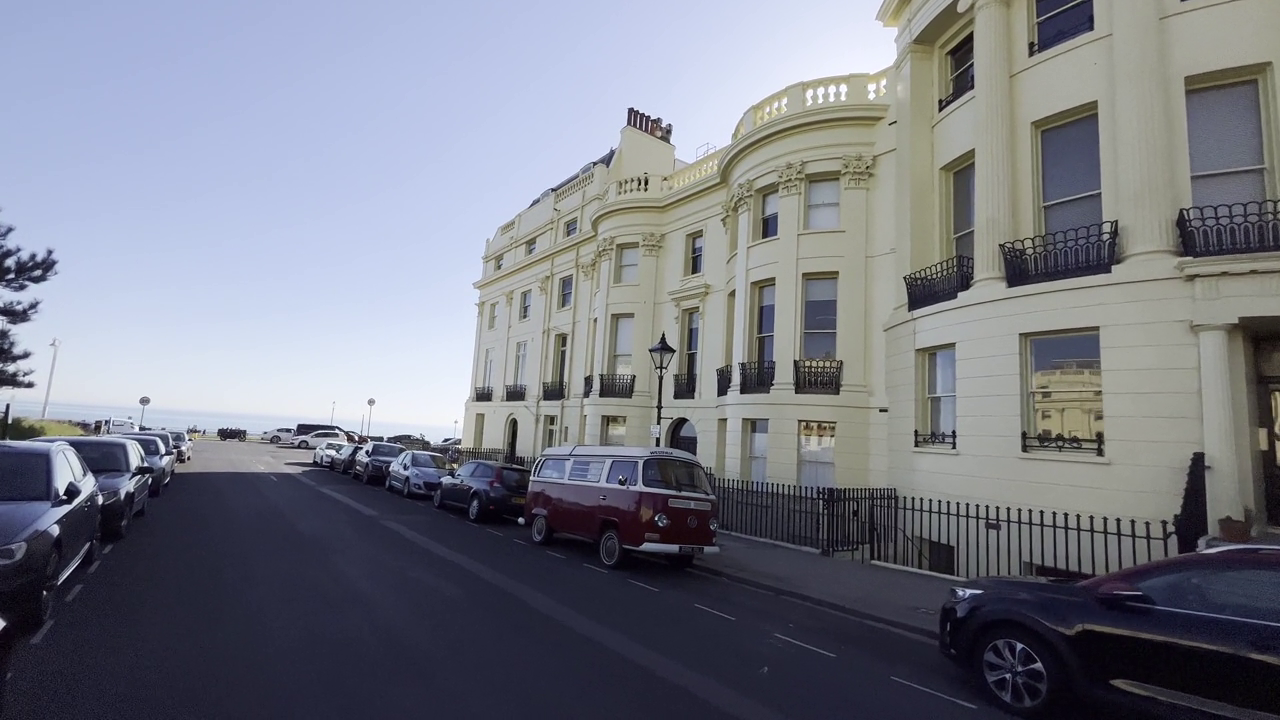}
         \caption{\centering Original image frame from Brighton data-set}
         \label{fig:out90}
     \end{subfigure}
     \hfill
     \begin{subfigure}{0.49\textwidth}
         \centering
         \includegraphics[width=\textwidth]{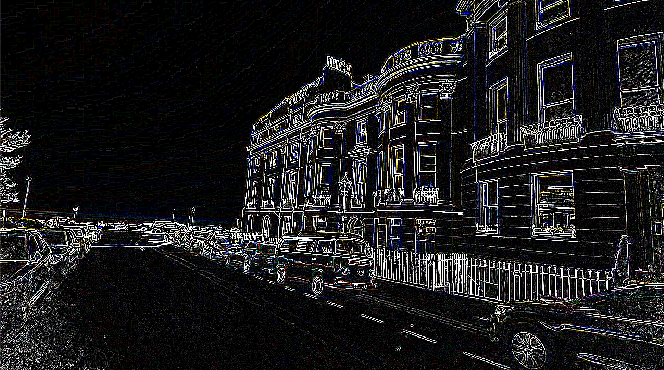}
         \caption{$\sigma=0.25$}
         \label{fig:out90sigma25}
     \end{subfigure}
     \hfill
     \begin{subfigure}{0.49\textwidth}
         \centering
         \includegraphics[width=\textwidth]{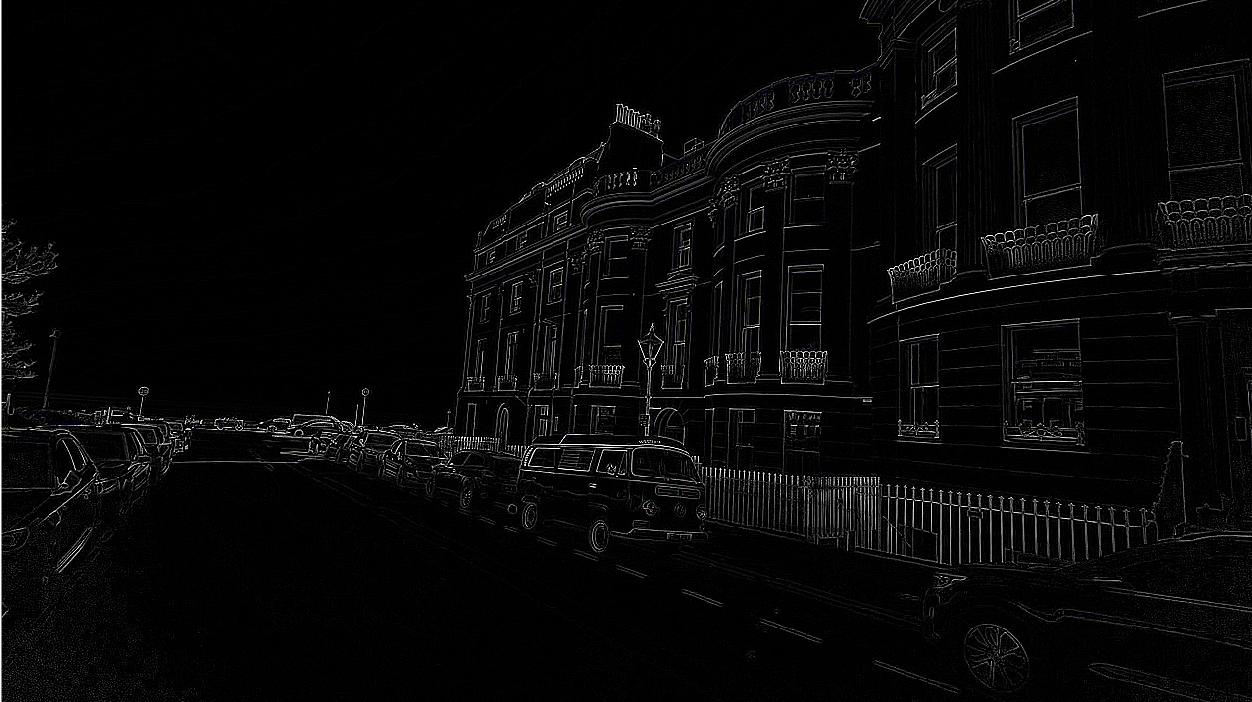}
         \caption{$\sigma = 0.5$}
         \label{fig:out90sigma50}
     \end{subfigure}
     \hfill
     \begin{subfigure}{0.49\textwidth}
         \centering
         \includegraphics[width=\textwidth]{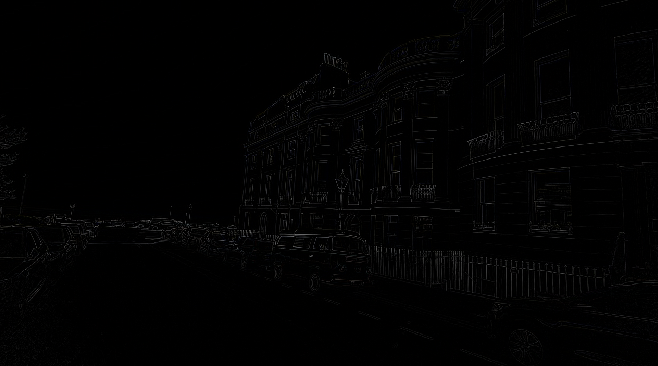}
         \caption{$\sigma = 0.75$}
         \label{fig:out90sigma75}
     \end{subfigure}
    \centering
    \begin{subfigure}[H]{0.49\textwidth}
        \centering
        \includegraphics[width=\textwidth]{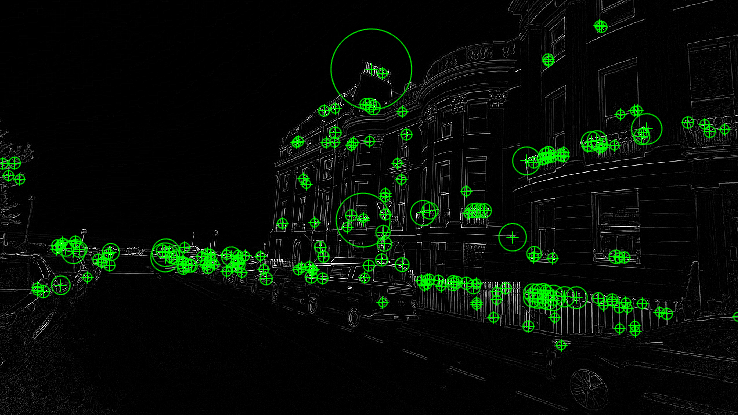}
        \caption[Two hundred most salient SIFT features over-layed on image masks.]{\centering Two hundred most salient SIFT features over-layed on image mask, $\sigma = 0.5$}
        \label{fig:out90siftfeatures}
    \end{subfigure}
    \hfill
    \begin{subfigure}[H]{0.49\textwidth}
        \centering
        \includegraphics[width=\textwidth]{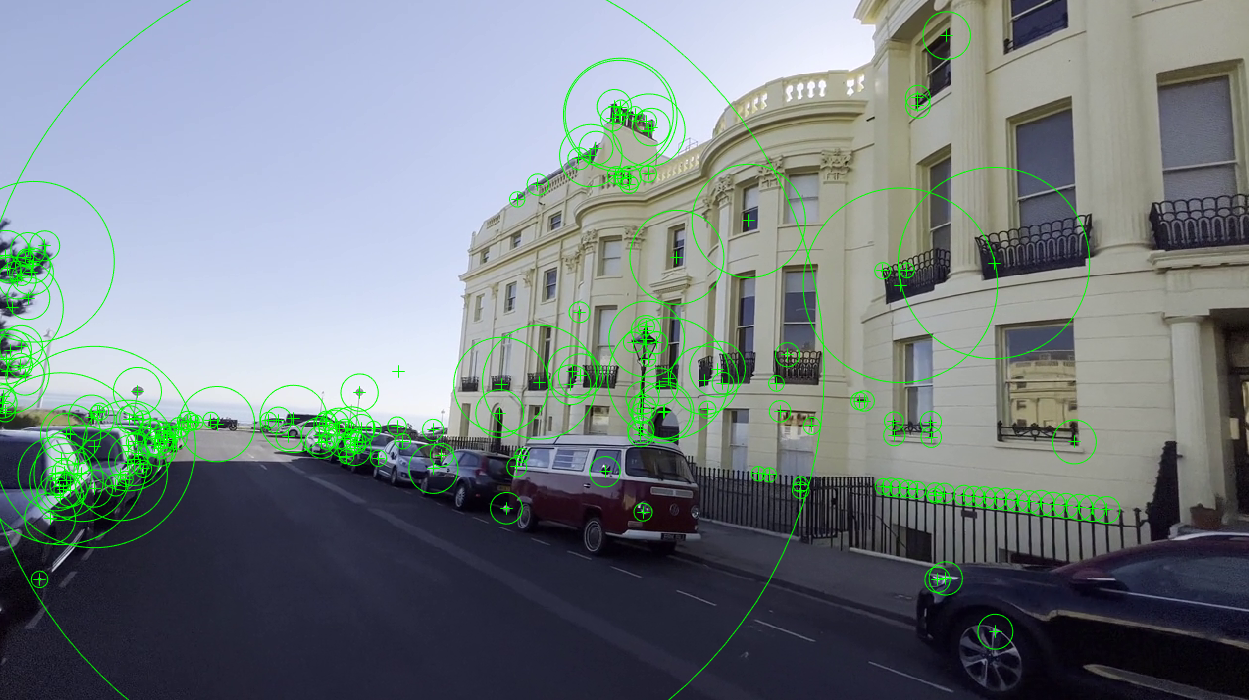}
        \caption[Two hundred most salient SIFT features over-layed on original image]{\centering Two hundred most salient SIFT features over-layed on original image, $\sigma = 1.60$}
        \label{fig:out90siftfeatures2}
    \end{subfigure}
    \caption[Image masks of rotationally symmetric Laplacian of Gaussian filter]{Image masks of rotationally symmetric Laplacian of Gaussian filter applied to Brighton data-set image to illustrate scale-space generation.}
        \label{fig:GaussianFilteredout90}
\end{figure}

\emph{SIFT Descriptor Generation} \newline
SIFT descriptors are generated by assigning a consistent orientation to each keypoint based on the local properties in the image, represented relative to its orientation, to achieve rotation invariance. Each keypoint has an orientation histogram computed from the gradient orientations of the sample points within a mask e.g. $3 \times 3$.  Points within $80\%$ of the highest peak are used to create a keypoint with the dominant orientation.
Image masks of the Gaussian filters with varying standard deviations $\sigma$ are shown in Fig. \ref{fig:GaussianFilteredout90}, the sigma value $\sigma$ should be lowered if the image is blurry. \footnote{\label{note:code}MatLab Code for the feature descriptor diagrams can be found in the git repository detailed in \ref{System Requirements}}
\subsubsection{Feature Matching, Fundamental Matrix and RANSAC}
\begin{figure}[H]
    \centering
    \includegraphics[width =0.75\textwidth]{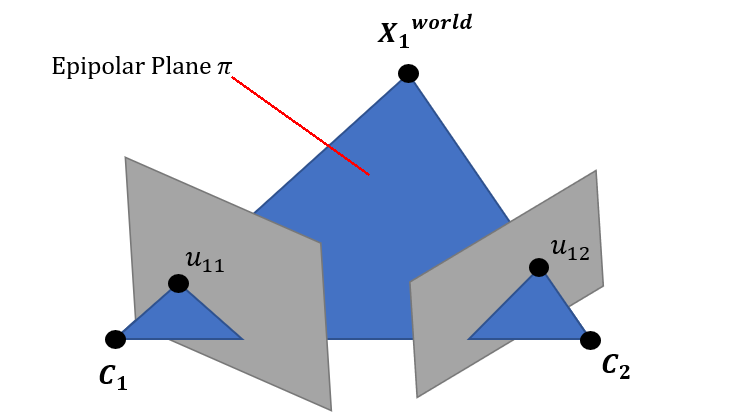}
    \caption[Epipolar Plane Diagram]{Epipolar Plane Diagram. Two camera views of a scene are shown. An observation $u_{11}$ in the first frame generates a ray in $\mathbb{R}^3$, intersecting camera $C_1$,$u_{11}$, and world point $\mathbf{X_1}^{world}$. Similarly with points $C_2$, $u_{12}$ and $\mathbf{X_1}^{world}$. Both rays intersect $\mathbf{X_1}^{world}$, and together with the baseline between $C_1$, $C_2$, define the epipolar plane $\pi$.}
    \label{fig:epipolarPlane}
\end{figure}

Feature matching plays a crucial role in the process of SfM, and can be the most time consuming step. The efficacy of the feature matching algorithm directly affects the accuracy of the output model. For example, the failure of 3D reconstruction in scenes with repetitive features is down to the feature matching step in the process. Exhaustive feature matching between frames can be very computationally time consuming, as each block of frames is matched with every other block, e.g. $50$ blocks is $1275$ combinations of $2$, as per the Eq. \ref{eq:combinations}, where $C(n,r)$ is the number of combinations, $n$ is the total number of elements in the set, and $r$ is the number of elements chosen from the set. 
Computation time can be minimised if sequential feature matching is performed in lieu of exhaustive feature matching.
However, the reduction in computation time comes at the cost of losing loop closure, as exhaustive feature matching (matching between all frames) can allow for matching between frames that are not consecutive, and thus allow for loop closure and model correction as multiple two view geometries are given for a single point. Where loop closure is the task of deciding if the camera has returned to a previously visited area, to correct drift error.
The recommended matching mode for large image collections is vocabulary tree matching, in which every image is matched against its visual nearest neighbors using a pre-trained vocabulary tree with spatial re-ranking \cite{Stathopoulou2019OPEN-SOURCEEVALUATION}.
Spatial matching matches every image against its spatial nearest neighbours \cite{Philbin2007ObjectMatching}, where spatial locations can be extracted via GPS in COLMAP software to use for spatial nearest neighbour search.
Transitive matching functions via the transitive relations of pre-existing feature matches \cite{Lin2018Copy-moveMatching}, i.e. if Image 1 matches with Image 2, and 2 matched with 3, then the matching function will attempt to match 1 to 3 directly, circumventing the requirement for exhaustive matching.

\begin{equation}\label{eq:combinations}
    C(n,r) = \frac{n!}{(r!(n-r)!)}
\end{equation}
An output of SfM is the epipolar geometry, the intrinsic projective geometry between the two views of the input images. The epipolar geometry describes the camera's internal parameters and the relative pose of the camera between two overlapping camera views, thus it is independent of the structure in the scene. The term camera is associated with the physical object of a camera using the same zoom-factor and lens, defining the intrinsic projection model \cite{JohannesL.Schoenberger2022COLMAPDocumentation}.
Point $\mathbf{X_1^{world}}$ in the world space, viewed in two overlapping images, identified and described using SIFT, points captured as in $\mathbf{u_{11}}$ in the first image and  $\mathbf{u_{12}}$ in the second; the points identified are coplanar with the respective camera centres $\mathbf{C_1},\mathbf{C_2}$, see Eq. \ref{eq:coplanar}.
\begin{equation}\label{eq:coplanar}
    \vv{\mathbf{C_1\mathbf{x_1^{world}}}} \cdot (\vv{\mathbf{C_1}\mathbf{C_2}} \times \vv{\mathbf{C_2}\mathbf{x_2^{world}}}) = 0
\end{equation}
An epipolar plane is a plane containing the intersection line joining the camera centres with the image plane.
The \Gls{epipolar plane} formed by these points is denoted by $\pi$. As the points are coplanar, the rays projected back from $u_11, u_12$ intersect at point $\mathbf{X_1^{world}}$, and a feature correspondence is found. This is illustrated in Fig. \ref{fig:epipolarPlane}. 

\begin{figure}[H]
    \centering
    \includegraphics[width =\textwidth]{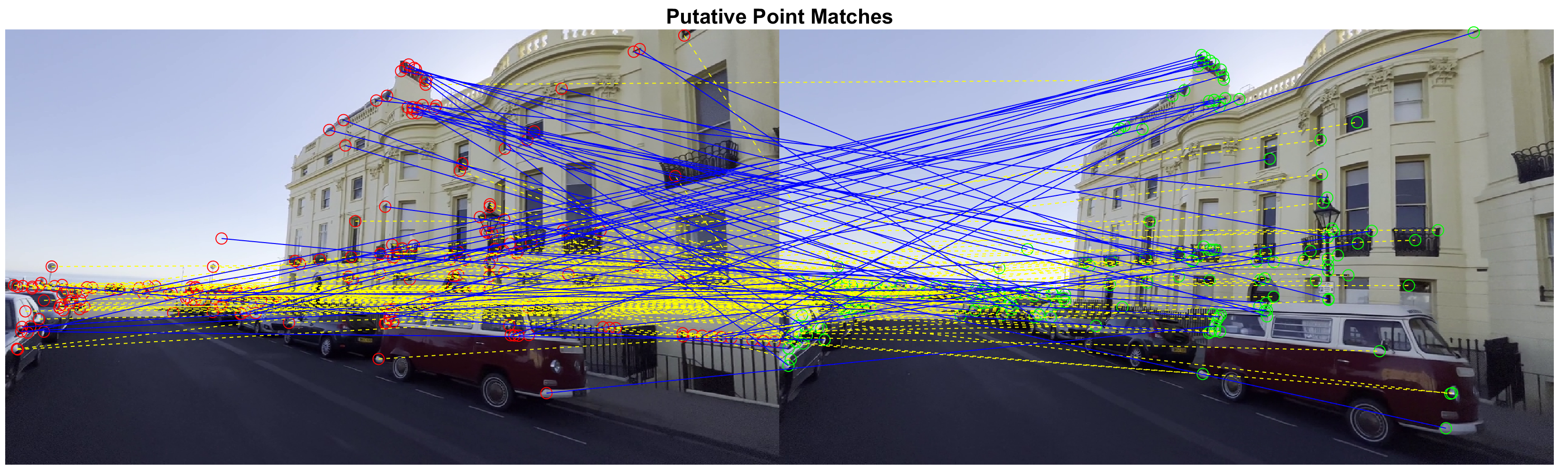}
    \caption[Brighton data-set feature points matched using RANSAC erroneous feature matching.]{Brighton data-set feature points matched using RANSAC erroneous feature matching. Gradients between features $\geq \bar{m} \pm (0.5 \times \sigma_m)$ labelled outliers.}
    \label{fig:featurematcherroneous}
\end{figure}

An example of feature matching between two frames using SIFT feature descriptors is shown in Fig. \ref{fig:featurematcherroneous}, this is performed by estimating the fundamental matrix from corresponding points in stereo images. Outliers are excluded using robust estimation techniques such as \Gls{RANdom Sample Consensus (RANSAC)}. RANSAC is an iterative method to estimate parameters of a mathematical model from a set of observed data which contains outliers. When RANSAC is used, results may not be identical between runs because of the randomized nature of the algorithm. This is also a perfect example of incorrect feature matching with repetitive architecture, as some of the feature matching lines shown do not match up completely in Fig. \ref{fig:featurematcherroneous}.
To illustrate the feature matching errors, as these two images are roughly similar in that there is little rotation, a constraint was applied that showed feature matching with a gradient between features exceeding the mean of the gradients $\pm (0.5 \times \sigma_m)$, shown in blue. Sigma $\sigma$ is the standard deviation of gradients. Of course, the two results are not exactly the same due to the randomized nature of RANSAC in the feature matching process.

\begin{figure}
    \centering
    \includegraphics[width = 0.7\textwidth]{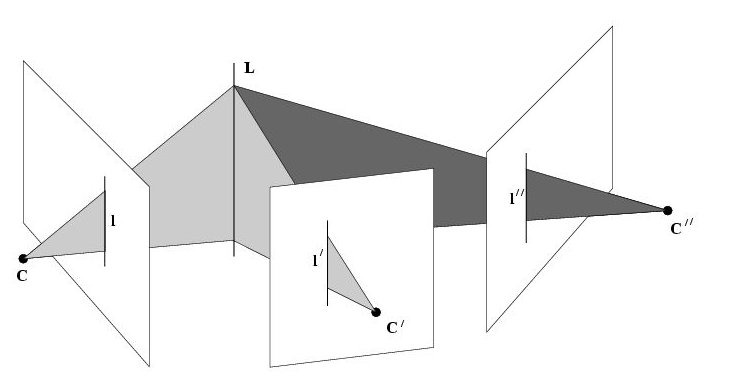}
    \caption[Trifocal Tensor Diagram]{Trifocal Tensor Diagram \cite{Hartley1997LinesTensor}. A line $L$ in $\R^3$, corresponding views $l$, $l'$, $l''$ indicated by their centres, $C$, $C'$ and $C''$ respectively. All projecting lines back from them intersect in a single $3D$ line in space $L$.}
    \label{fig:trifocalTensor}
\end{figure}

The fundamental matrix $\mathbf{F}$ is a $3 \times 3$ algebraic representation, of rank $2$, of the epipolar geometry with 7 Degrees of Freedom (DOF), the epipolar constraint is as follows in the Longuet-Higgins Eq. \ref{eq:LonguetHiggins}. The fundamental matrix stipulates that if a point in the world $\mathbf{X}$ is observed by both cameras as $\mathbf{u_{11}}$ and $\mathbf{u_{12}}$. The image points must satisfy the Longuet Higgins relation, this is illustrated in Fig. \ref{fig:epipolarPlane}.
\begin{equation}\label{eq:LonguetHiggins}
    \mathbf{x}_i^{'\top} \mathbf{F}\mathbf{x}_i=0, i \in (1,m)
\end{equation}
 The Eq. \ref{eq:LonguetHiggins} \cite{Hartley2004Multiple11343} can be solved via Singular Value Decomposition (SVD).
By enforcing the epipolar constraint, the pose from two views can be constructed by either using the 8-point \cite{Osterman1981AProjections} or 5-point \cite{Nister2004AnProblem} algorithm. The trifocal tensor offers solution for if three views are available \cite{Risqi201837Survey}, encapsulating all of the projective geometric relations between the three views that are independent of the scene structure \cite{TheTensor}. The geometric explanation of the trifocal tensor is shown in Fig. \ref{fig:trifocalTensor}. The SfM pipeline used, COLMAP, does not utilise trifocal tensors.
As the point correspondences computed by SIFT can be noisy, RANSAC is used to obtain a better estimate of the fundamental matrix, so to maximise the number of inliers, as shown in Alg. \ref{alg:RANSAC}.
RANSAC is an algorithm for robust model fitting given data which may contain outliers \cite{Fischler1981RandomConsensus}. It is an iterative algorithm for robust model fitting given noisy data that may contain outliers, through generating a set of candidate models using a minimum number of randomly sampled points for each model. Each candidate model in the RANSAC process is scored such that the total number of data points in agreement, the chosen model is the one with the greatest number of inliers. Thus, RANSAC performs outlier rejection, as well as model fitting. Increasing the number of trials increases the probability of generating a high scoring candidate model. To ensure a probability $z$ of randomly sampling from only the inlier set, the number of trials is given as $k = \frac{log(1-z)}{log(1-w^n}$, where $w$ is the probability of selecting an inlier from a set of data points. The set of data points in the context of structure from motion is the correspondences between 2D observations in the image and 3D landmark features in the map.
SfM uses pure appearance-based correspondences between keypoints/descriptors, which are are defined by matches, and the inlier matches are geometrically verified and used in reconstructions in COLMAP.
\begin{algorithm}[H]
\caption{RANSAC for Fundamental Matrix Algorithm}\label{alg:RANSAC}
\hspace*{\algorithmicindent} \textbf{Input: 8 Random Image Correspondence $\hat{x_1},\hat{x_2}$}  \\
\hspace*{\algorithmicindent} \textbf{Output: Determined Inliers}
\begin{algorithmic}

\For{i = 1:M} 
\State F = EstimateFundamentalMatrix$(\hat{x_1},\hat{x_2})$ 
\State $\mathcal{S}=0$ 
    \For{j = 1:N} 
        \If{$|x^\top_{2j}Fx_{1j}| \leq \epsilon$} 
            \State $\mathcal{S} = \mathcal{S} \cup \{j\}$
        \EndIf
    \EndFor
    \If{$n \leq|\mathcal{S}|$}
        \State $n = | \mathcal{S}|$
        \State $\mathcal{S}_in = \mathcal{S}$
    \EndIf
\EndFor

\end{algorithmic}
\end{algorithm}

\subsubsection{Estimate Essential Matrix from Fundamental Matrix}
The Essential Matrix $E$ allows for computation of the relative camera poses between the two images being compared, it is a $3 \times 3$ matrix with 5 DOF, and obeys the pinhole model Eq. \ref{eq:pinholeModel}, where $\mathbf{K}$ is camera intrinsic matrix, and $\mathbf{F}$ is the Fundamental Matrix. $F$ is defined in world space, and $\mathbf{E}$ is defined in normalized image space.
\begin{equation}\label{eq:pinholeModel}
    \mathbf{E = K^\top FK}
\end{equation}
\subsubsection{Estimate Camera Pose from Essential Matrix} 
The Camera Pose $P$ consists of 6 (DOF), rotation: roll, pitch, yaw; and translation that consists of: X,Y,Z.
\begin{equation}\label{eq:cameraPose}
    P = KR[\mathcal{I}_{3 \times 3} - C]
\end{equation}
\subsubsection{Check for Chirality Condition using Triangulation}
The Chirality Condition is a constraint to correct the unique camera poses, ensuring that the reconstructed points are in front of the cameras using \Gls{linear least squares} \cite{Szeliski2011ComputerApplication, Hartley2007Parameter-freeEstimation}. Linear least squares is a least squares optimisation of linear functions to data, by minimising residuals $E_{LS} = \sum_i{||\hat{x'_i}- \tilde{x'_i}||}$.
A 3D point $X$ is in front of the camera if $r_3(\mathbf{X-C}) >0$, where $r_3$ is the z axis of the camera. The best camera configuration is considered to be the camera configuration that maximises the number of points satisfying the Chirality Condition.

\subsubsection{Bundle Adjustment} \label{subsubsec: Bundle Adjustment}

\begin{figure}[h]
    \centering
    \includegraphics{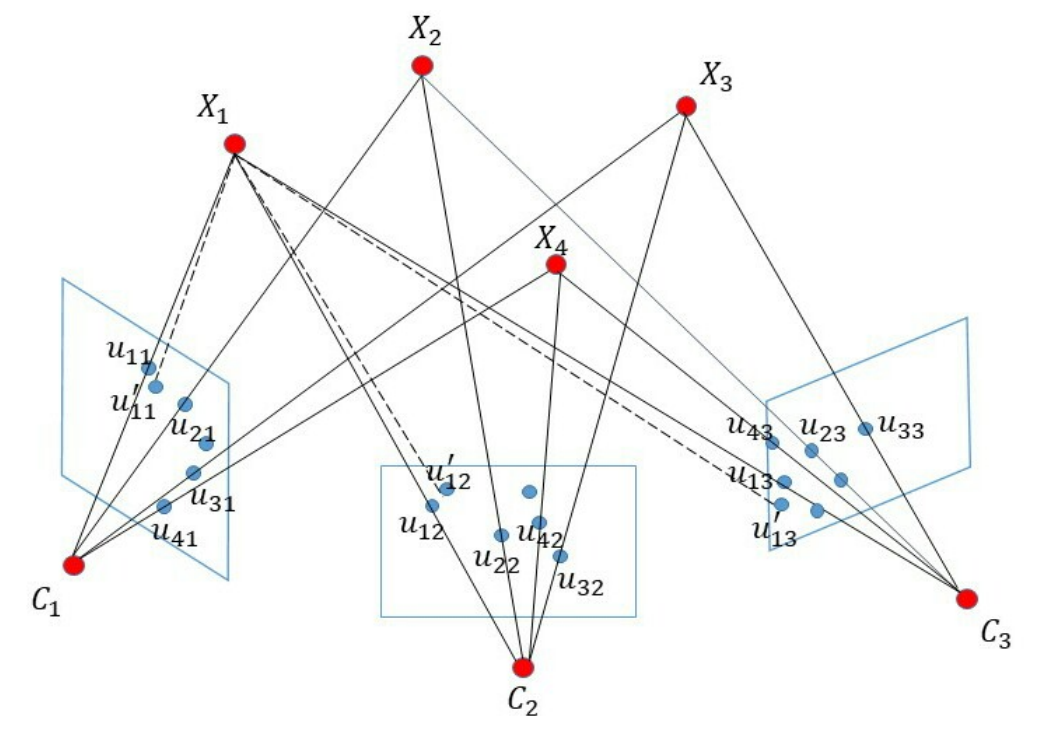}
    \caption[Bundle Adjustment Diagram.]{Bundle Adjustment: $u_{ij}$ are the observations, $u'_{ij}$ are the reprojected 2D points, and $X_i$ are the 3D points in the world frame. The solid line and dotted line represent the projection, and reprojection respectively. \cite{BundleDiagram}}
    \label{fig:BA_diagram}
\end{figure}

Bundle Adjustment (BA) is a non-linear optimisation technique \cite{Cornelis2004DriftAlgorithms} that plays an important role in 3D reconstructions such as SfM, or SLAM. BA optimizes camera parameters and 3D points as a final refinement step of visual reconstructions in the SfM pipeline \cite{Chen2019BundleRevisited}, as shown in Fig. \ref{fig:BA_diagram}. This is performed by minimizing reprojection error as shown in Eq. \ref{eq:minimiseReprojectionError}. The reprojection error is the difference between the projected and re-projected bundles of light.
It was initially invented in the late 1950's and 1960's for the application of aerial photogrammettry, where by the 1970's already offered capability to reconstruct thousands of images into structures \cite{Triggs2000BookVisionAlgorithmsTheoryAndPract}.
\begin{equation}\label{eq:minimiseReprojectionError}
    E(\mathbf{P,X}) = \sum_{i=1}^{m}\sum_{j=1}^{n}D(x_{ij},P_iX_j)^2
\end{equation}

The BA cost function can be minimized by the Levenberg-Marquardt (LM) algorithm \cite{more1978levenberg}, this is the most successful algorithm for solving this problem, as it is simple to implement, robust to initialization and able to take advantage of multi view geometrical forms of sparsity. The LM algorithm combines \gls{Gauss-Newton} (a second order algorithm) and gradient descent (a first order algorithm).
Gauss-Newton is a Newton-like method for solving a non-linear least squares problem, in which the Hessian $H$ is approximated by $H \approx  J^\top WJ$. Hessian is a matrix of second partial differentials of the cost function $H = \frac{d^2f}{dx^2}$. Where $J$ is the design matrix and $W$ is the weights. The normal equations are the resulting prediction equations given as $(J^\top WJ) \delta x = - (JW \Delta z)$.
Gradient Descent is a naive optimization method which consists of steepest descent down the gradient of the given cost function.
Each step of the LM algorithm produces an improved estimate of the camera parameters, and the resulting series of iterates minimizes the objective function at hand \cite{Wright2006Nocedal-Wright2006_Book_NumericalOptimization}.
\begin{equation}\label{eq:LM}
    \min \frac{1}{2}||J_kp +r_k||^2 \textrm{, subject to } ||p|| \leq \Delta_k 
\end{equation}
Where $\Delta_k >0$ is the spherical trust region radius, the use of a trust region avoids one of the weaknesses of Gauss-Newton minimization, namely, its behaviour when the \gls{Jacobian} is rank deficient. Thus the model function is shown in Eq. \ref{eq:LM2}, and detailed step by step in Alg. \ref{alg:LM}. The Jacobian is a matrix of partial differentials of the cost function $\mathbf{J} = \frac{df}{dx}$.
\begin{equation} \label{eq:LM2}
    m_k(p) = \frac{1}{2}||r_k||^2 + p^\top \mathbf{J_k}^\top + \frac{1}{2}p^\top \mathbf{J_k}^\top \mathbf{J_k} p
\end{equation}

\begin{algorithm}[H]
\caption{Levenberg-Marquardt Algorithm}\label{alg:LM}
\hspace*{\algorithmicindent} \textbf{Input:} A training input vector \\
\hspace*{\algorithmicindent} \textbf{Output:} Vector of modified weights and biases to give minimised reprojection error
\begin{algorithmic}
    \Require Reprojection Error $E_k$ \Comment{Using Eq. \ref{eq:minimiseReprojectionError}}
    \While{(not stop-criterion)} 
    \State Compute weight update $\Delta W = (\mathbf{J}^\top \mathbf{J} + \lambda \I)^{-1}\mathbf{J}^\top p$
    \State Update the network weights $(W)$ using $W = W+ \Delta W$
    \State Recalculate $E_k$ \Comment{Using Eq. \ref{eq:minimiseReprojectionError}}
\If{$E_k$ decreased}
    \State $\lambda = 0.1 \cdot \lambda$
\ElsIf{$E_k$ increased}
    \State $\lambda = 10 \cdot  \lambda$
\EndIf
\EndWhile
\end{algorithmic}
\end{algorithm}

\subsubsection{3D Model Representations}
There are many 3D model representations that vary in structure and in properties, point clouds, voxels, RGBD, 3D Meshes and octomaps, to name a few. The choice of representation is important, as a powerful and discriminative feature descriptor in SfM must be able to capture the geometric structure and be invariant to translation, scaling and rotation, thus the data type must cooperate.
3D Point clouds \cite{Han20183DState-of-the-Art} are a set of data points in 3D space, which typically have 7 dimensions in SfM algorithms, $(XYZRGBA)$, which represent a position in 3D space $(XYZ)$, pixel colour $(RGB)$, and pixel transparency alpha $A$. Point clouds are typically used for classification \cite{Lhuillier2005AImages, Qi2017PointNet:Material}, as there are sensors available that can directly capture point cloud data \cite{Tamura2020TowardsDeformation}. 
Some advantages to using point clouds are that its possible to multiply the discrete points with linear transformation matrices, objects can be combined by merging point clouds together, and an exact representation can be achieved. However, point clouds can model neither unknown areas, nor free space, and the sensor noise and dynamic objects cannot be dealt with directly. Thus, point clouds main application is for high precision sensors in a static environment without dynamic parties, and when the occluded areas do not need to be represented. Furthermore, the memory consumption of point cloud representations increases with the number of measurements, which can be problematic as there is no upper bound.
Octomaps \cite{Hornung2013OctoMap:Octrees}, based on octrees offer a memory efficient way to visualise and process 3D structure data, providing a volumetric representation of space with variable degrees of precision. The mapping process in octomaps uses probabilistic occupancy estimation to present 3D points, and octree map compression keeps the models compact. Probabilistic occupancy is derived from the uncertainty in 3D range measurements, where multiple uncertainty measurements can be fused into a robust estimation of the true state of the environment. The volumetric representation of the environment is a grid of cubic volumes of equal size, called voxels, discretizing the mapped area, decreasing the size of the voxels thus increases precision in the model. However, voxels are unsuitable for representing detailed 3D shapes owing to a trade-off between resolution and memory consumption \cite{Tamura2020TowardsDeformation}.
RGBD is often the format used for acquiring spatial data, offering $RGB$ and depth $D$.
A sparse point cloud is used to find out which image pairs are overlapping i.e. which images have sufficient number of common SIFT points, and dense clouds generate the depth maps for such overlapping pairs.

\subsection{COLMAP}
Several conventional Multi-view Stereo (MVS) and SfM algorithms were proposed, including PMVS \cite{Wang2021AnCharacteristics}, OpenMVG \cite{Moulon2017OpenMVG:Geometry}, VisualSfM \cite{Morgan2016HowVisualSFM} and COLMAP \cite{Schoenberger2022StructureCOLMAP}. COLMAP was chosen as it performs best on the ETH3D data-set \cite{Schops2022AVideos}. Generally, SfM algorithms can be categorized into incremental, global and hybrid methods. Incremental SfM reconstructs a 3D map from two initial images and incrementally adds new images to update the map. Global SfM simultaneously estimates all the camera poses, while the hybrid method, such as COLMAP offers advantages of both.
\par
\begin{figure}
    \centering
    \includegraphics[width =0.8\textwidth]{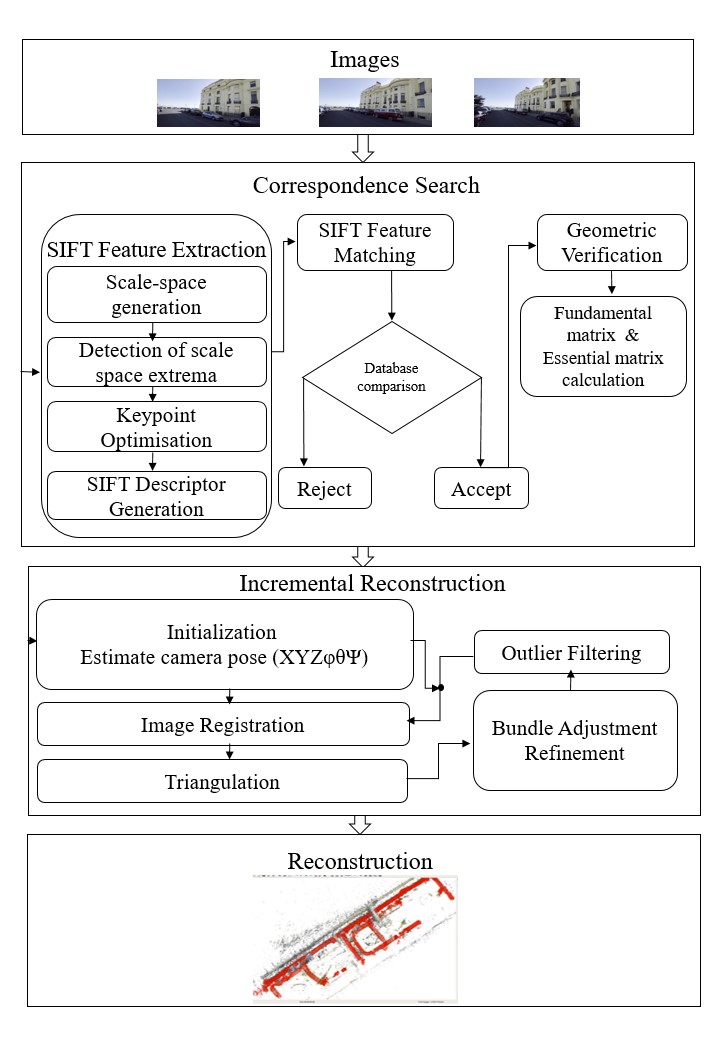}
    \caption[COLMAP incremental SfM pipeline.]{COLMAP incremental SfM pipeline flowchart adapted from \cite{JohannesL.Schoenberger2022COLMAPDocumentation}}
    \label{fig:COLMAP_flowchart}
\end{figure}

COLMAP is an open source SfM pipeline with a Graphical User Interface (GUI), Command Line Interface (CLI), and the ability to reconstruct 3D objects automatically from either monocular or stereo camera setups. Created for research, COLMAP offers advanced functionality in that the software can extensively define the camera intrinsic and extrinsics that was used to capture the best structure. COLMAP dense reconstructions require CUDA, however sparse reconstructions can be performed on CPU. SfM recovers coarse point clouds in areas where the keypoints match, whereas MVS reconstructs dense point clouds using matched areas among images. Although COLMAP achieves state of the art performance in many benchmarks, it requires a significant amount of computation time to reconstruct 3D models. 
Consider the photographs in Fig. \ref{fig:COLMAP_flowchart} input images; these show the same environment observed from a few different viewpoints. Even with this being the first time you (the observer), has seen this environment, it is not difficult to infer the inherent spatial structure of the scene and the object organisation in physical space. Furthermore, upon human observation, the semantic content of the scene and distinguishing it's individual object components is clear. 
In the flowchart, correspondence search refers to an algorithm such as Scale Invariant Feature Transform (SIFT), in which the features are detected, described, and matched with local features in the images. Feature matching is finding feature scene points that are observed in multiple frames. The initialisation refers to the robust estimation of camera poses and/or scene points, and the bundle adjustment refers to the refinement of camera poses and scene structure points, see Sec. \ref{subsubsec: Bundle Adjustment}.
\par
SfM recovers the camera parameters and 3D locations of the identified features given observed 2D image feature points, often represented in a point cloud, points with position and colour. Bao et al \cite{Bao2011SemanticMotion} propose a pipeline for semantic structure from motion (SSFM). By combining geometric constraints from SfM, SSFM addresses the problem of recovering camera parameters through semantic and geometric properties of the associated identified objects and features within the scene.
SSFM differs from vanilla SfM as the images are input to the algorithm, feature detection is performed frame by frame, feature matching is performed, as in SfM. However, in addition, an object detection and recognition algorithm is deployed and identifies the objects in 2D without reasoning the geometry. In SfM, 3D point clouds are returned with no semantic information attached, and the SSFM jointly recognizes the observed objects and reconstructs the underlying 3D geometry of the scene (cameras, points, objects) given the object recognition priors also.
The unique aspect of Yingze et al's framework is that it can estimate camera poses given only object detections, in turn, this can enhance the estimated camera pose compared to vanilla feature point based SfM. Furthermore, this pipeline can improve object detections given a series of uncalibrated images compared to object detection in single frames and single point of view. 
This paper fills the void between object recognition and SfM, offering 2D location and scale, 3D spatial structure, and semantic content of the components.

\subsection{Semantic Segmentation}
\begin{figure}
    \centering
    \includegraphics[width = 0.8\textwidth]{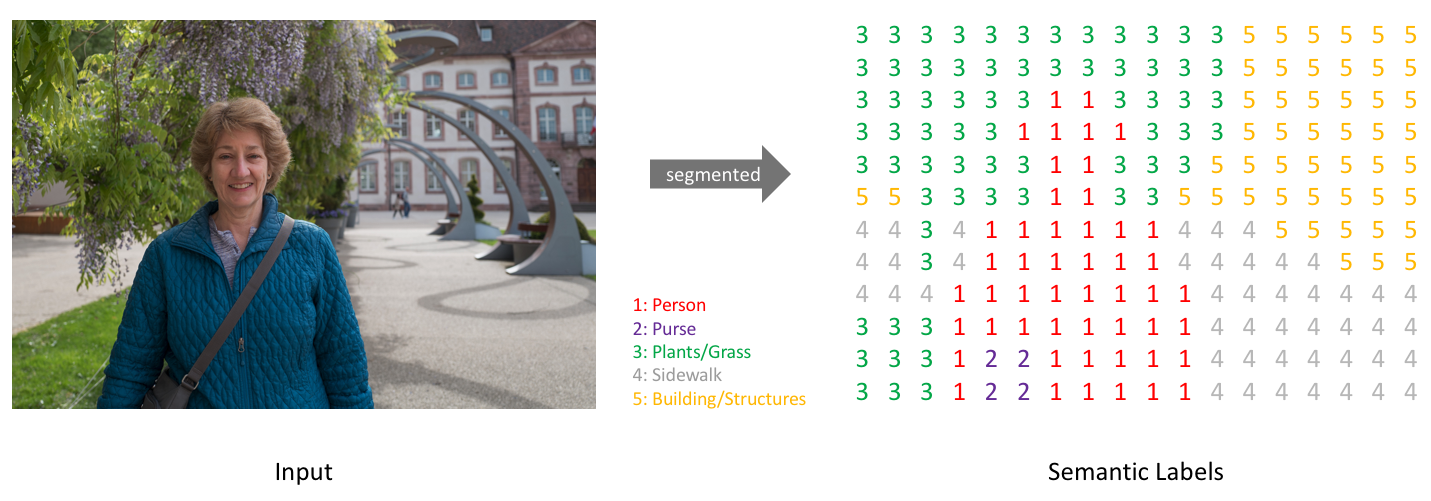}
    \caption{Semantic segmentation prediction map example \cite{JordanAnSegmentation.}
    \label{fig:semsegexample}}
\end{figure}

The goal of semantic segmentation in images is to label each pixel with a corresponding class of what is being represented, allowing for a rich understanding of the scene shown. The objective in semantic segmentation is to take an RGB color image, or grayscale image, and output a segmentation map where each pixel contains a class label represented as an integer. An example of a low resolution prediction map is shown in Fig. \ref{fig:semsegexample}.
\subsubsection{DeepLab Deep Learning Model for Semantic Segmentation}
\begin{figure}[H]
    \centering
    \includegraphics[width =0.75\textwidth]{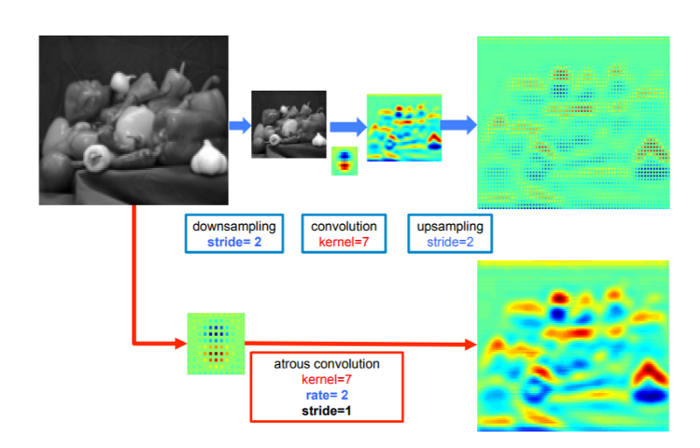}
    \caption[DeepLab atrous convolution flowchart.]{Convolution followed by up-sampling (top). Atrous convolution resulting in denser feature map output (bottom) \cite{Chen2016DeepLab:CRFs}}
    \label{fig:deeplabAtrous}
\end{figure}

\begin{figure}[H]
    \centering
    \includegraphics[width =0.75\textwidth]{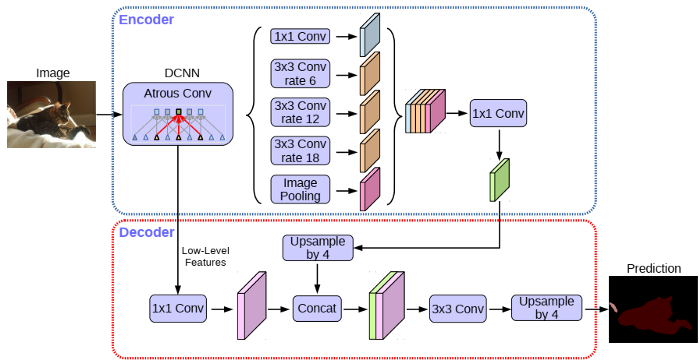}
    \caption{DeepLabv3+ Model Architecture \cite{Chen2017RethinkingSegmentation}}
    \label{fig:deepLabv3ModelArchitecture}
\end{figure}

DeepLab is a state-of-the-art semantic segmentation model designed and distributed by Google. DeepLab can achieve dense prediction, generating an output map of the same size as the input image, through up-sampling the output of the final layer of the Fully Convolutional Network (FCN) via atrous convolution \cite{Chen2016DeepLab:CRFs}. Atrous convolution $y[i]$, for a single dimension signal $x[i]$, with filter $w[k]$ or length $K$ and stride rate $r$ can be defined by Eq. \ref{eq:atrousConvolution}.
\begin{equation}\label{eq:atrousConvolution}
    y[i] = \sum_{k=1}^{K}x[i+r\cdot k]w[k]
\end{equation}
DeepLab excels as it uses atrous convolution \cite{chen2017deeplab}, that allows enlargement of the field of view of filters to incorporate larger context.
In traditional Deep Convolutional Neural Networks (DCNN), convolutional layers systematically apply learned filters to input images to create feature maps that summarize the presence of those features in the input. Stacking of these convolutional layers allows for layers close to the input to learn low level features, and deeper layers to learn higher-order, more abstract features. A pooling layer can make the model more robust to slight changes in feature position/orientation. Max-pooling calculates the maximum value for each patch of the feature map, and often is of size $2 \times 2$ pixels applied with a stride of $2$ pixels, thus reducing feature map size by $2 \times$.
Thus, repeated max-pooling and striding in consecutive layers will significantly reduce the spatial resolution of resulting feature maps. DeepLab's approach to DCNN utilises atrous convolution, resulting in a denser feature map than traditional DCNN approach of up-sampling the output feature map, as shown in Fig. \ref{fig:deeplabAtrous}.
Due to multiple pooling and down-sampling stride in traditional DCNN, a reduction in spatial resolution occurs. DeepLab's model removes the down-sampling operation from the max pooling layers, and instead up-sample the filters through atrous convolution in the subsequent convolutional layers, this results in denser feature maps computed at a higher sampling rate. In DeepLabv3+, the encoder is has an output stride of 16, thus the input image is down-sampled by a factor of 16. In decoding, instead of using bi-linear up-sampling of factor $16\times$, meaning the encoded features are first up-sampled by a factor of 4, and concatenated with corresponding low level encoded features with the same spatial dimensions that have had $1\times1$ convolutions applied to reduce the number of channels. After concatenation, $3 \times 3$ convolutions are applied and the features are up-sampled by a factor of 4, giving a dense feature map output the same size as the input image, as shown in Fig. \ref{fig:deepLabv3ModelArchitecture}.


\subsection{Planar Reconstruction from Sparse Views and Images}
In general, surface reconstruction is an ill-posed problem, as there are several triangulated surfaces that can fulfill the criteria of a surface. Sparsity, redundancy, noisiness of the point cloud, and boundaries of the surface are all challenges faced in surface reconstruction.
The SfM and MVS pipeline COLMAP offers two techniques for meshing of dense point clouds, Poisson meshing and triangulation.
A Poisson surface reconstruction generates a mesh by estimating the surfaces using the surface normal of input points, thus accurate surface normal is required, and outliers can cause bad surfaces, or holes in the model. 
Poisson surface reconstruction has three main steps \cite{Kim2018DigitalModels}: (a) transforming the oriented point cloud into a continuous vector field in three dimensions, (b) determining a scalar function that best matches the continuous vector field, and (c) extracting the appropriate iso-surface through the normals of the scalar functions. Thus, a mesh is generated from the oriented point cloud.
Delaunay triangulation is more robust to outliers than Poisson, and operates by establishing neighbourhood connections between point cloud points, in all relevant directions in a way that accommodates non uniform samples \cite{Cazals2006DelaunayReconstruction} to form polygonal meshes.
In literature, deep learning based three dimensional shape understanding has been applied in different tasks with promising results \cite{Han2020DRWR:Images, Han2020ShapeCaptioner:Sentences, Han2020SeqXY2SeqZ:Coordinates}. Deep learning models function through learning priors from pre-labelled data-sets. 
\par
Planar surface reconstruction is an important aspect in SfM reconstructing from point clouds. Current methods in the space of planar surface reconstruction in SfM require that the point cloud be a dense reconstruction, where surfaces are learned by Signed Distance Functions (SDFs) \cite{Osher2003SignedFunctions} from single point clouds without ground truth signed distances or point normals \cite{Sharma2017SpecialCross-sections}. SDFs gives the distance of a 3D point $X$ from the boundary of a surface, and this can be used to determine if a point lies inside or outside the boundary of a surface. Eq. \ref{eq:SDF1}, \ref{eq:SDF2} illustrates an SDF example in 2D Euclidean space, where $f(x,y)$ is the implicit function of a circle centred at the origin, with unit radius, and $\psi(x,y)$ is the function's SDF.
\begin{equation}\label{eq:SDF1}
    f(x,y) = x^2 + y^2 -1
\end{equation}
\begin{equation}\label{eq:SDF2}
    \psi = \sqrt{x^2 + y^2}-1
\end{equation}

The requirement for dense point cloud input limits the utility in real applications. There exists a method of planar surface reconstruction in which a surfaces can be reconstructed accurately from sparse point clouds and an on-surface prior. \cite{Ma2022ReconstructingPriors} trains a neural network to learn SDFs through projecting queries onto the surface represented by the sparse point cloud.

Learning SDFs $f_{\theta}$ from sparse point cloud $G$ without requiring ground truth signed distances and normals on points in $G$.
The SDFs can be used to predict signed distances $s = f_{\theta} (q,c)$ where $q$ is an arbitrary query sampled in $G$, and c is the condition identifying $G$. After learning $f_{\theta}$, the surface can be reconstructed using a marching cubes algorithm \cite{Lorensen1987MarchingAlgorithm}. Marching cubes algorithm calculates triangle vertices using linear interpolation and finds the gradient of the original data, normalizes it, and uses it as a basis for shading the 3D reconstructed planar models. The algorithm takes eight neighbor locations at a time, to form an imaginary cube, and thus determines the polygons required to represent the part of the isosurface that passes through the cube. The individual polygons can then be fused together to form the desired surface.
\par 
PlanaRCNN \cite{Liu2018PlaneRCNN:Imageb} is a CNN for 3D plane detection and reconstruction from single RGB image views. PlanaRCNN is based on three main steps; a plane detection network, a segmentation refinement network, and a warping loss module. PlanaRCNN employs R-CNNs to detect planes and segmentation masks, and to then jointly refine segmentation masks by enforcing a consistency with a nearby view during training. This allows for robust plane extraction in indoor scenes. The model PlanaRCNN is trained on ScanNet \cite{Dai2017ScanNet:Scenesb} indoor data-set and is attempted to generalise to an outdoor data-set in this project. 

\subsection{Consistency Grammar}
\begin{table}[ht]
  \caption{\centering List of Different Semantic Incorporation in Structure from Motion Systems}\label{tab:SSfMList}
  \small
  \centering
  \begin{tabular}{lcccccr}
  \toprule[\heavyrulewidth]\toprule[\heavyrulewidth]
  \textbf{Year} & \textbf{Name} & \textbf{Method} & \textbf{Reference}\\ 
  \midrule
  2011 & Semantic Structure from Motion &\makecell{Geometry estimation  \\ with object detection} &  \cite{Bao2011SemanticMotion}\\
  \hline
  2011 & \makecell{Semantic Structure from Motion with \\ Object and Point Interactions} & \makecell{High level  \\semantic correlations}& \cite{Bao2011SemanticInteractions} \\
  \hline
  2012 & \makecell{Semantic Structure from Motion  \\ with Points, Regions, and Objects} & \makecell{Point-object interaction  \\
  constraints} &\cite{Bao2012SemanticObjects}\\ 
  \hline
  2012 & \makecell{CityGML - Interoperable Semantic \\ 3D City Models} &\makecell{Geometric, topologic \\ and semantic definitions}&\cite{Groger2012CityGMLModels} \\
  \hline
  2018 & \makecell{Semantic Match Consistency  \\for Long-Term Visual Localisation} &\makecell{Semantic feature \\ matching consistency}& \cite{Toft2018SemanticLocalization} \\
  \hline
  2018 & \makecell{Deep Functional Dictionaries: \\ Learning Consistent Semantic  \\Structure on 3D Models from Functions} &\makecell{Learned geometric  \\ features}& \cite{Sung2018DeepFunctions} \\
  \hline
  2021 & \makecell{Semantic Structure from Motion  \\for Railroad Bridges Using Deep Learning} & DCNN& \cite{Park2021SemanticLearning} \\
  \hline
  2021 & \makecell{Semantic Consistency Networks  \\for 3D Object Detection} & Semantic consistency  &  \cite{Wei2021SemanticDetection}\\
  \hline
  2021 & \makecell{An Improved Method for Stable Feature Points \\ Selection  in Structure from Motion considering \\ Image Semantic  and Structural Characteristics}& \makecell{Semantic and geometric \\ consistency} & \cite{Wang2021AnCharacteristics}\\\hdashline
  
  \bottomrule[\heavyrulewidth] 
  \end{tabular}
\end{table}

In recent years, there has been vast progress in indoor data collection \cite{Nikoohemat2021ConsistencyChecking}, partly due to the advancements in remote sensing devices such as RGB-D cameras. Such sensing systems as these provide high quality rasters, point cloud information, and depth information. The abundance of information available enables reconstructions of indoor 3D models. Where rasters are images that are compiled using pixels, or tiny dots, containing unique color and tonal information that comes together to form an image.
Indoor 3D models in question can be constructed as 3D digital representations of scanned data from such like RGBD cameras (digital depth cameras), laser scanners, or CAD drawings. As a consequence to the data acquisition methods, this data is noisy and can cause erroneous loop closure, rendering the model useless. Nikoohemat et al \cite{Nikoohemat2021ConsistencyChecking} proposed a conceptual framework for checking semantic, geometric, and topologic consistency in reconstructed structure models.

The three steps taken to validate the 3D point cloud model in the paper \cite{Nikoohemat2021ConsistencyChecking} are as follows:
\begin{enumerate}
    \item Correctness checking of individual components
    \item Consistency verification of instances interactions
    \item Model consistency check for targeted applications 
\end{enumerate}
The proposed pipeline steps are performed by storing the model in a context free grammar structure, as a class which is either an abstract (e.g. space) or real component (e.g. furniture). Whereas spaces are represented by their surroundings, real components are represented by geometric objects such as solids, surfaces or boundaries. Furthermore, an instance refers to an object of this class (e.g. each room is an instance of the space class, or each piece of furniture is an instance of the furniture class). The 3D models are then represented as a vector geometric model or \gls{raster}.
Constraints are also an aspect of the consistency grammar system, in which they act as rules to (in)validate the model, e.g. a brick wall is not transparent.
The classes used in the model are as follows:
\begin{enumerate}
    \item Permanent structure classes including walls, floors and ceilings.
    \item Openings 
    \subitem Doors
    \subitem Windows
    \item Furniture, obstacles and other objects 
    \item Spaces are an abstract class represented by surroundings.
    \item Stairs or staircases 
\end{enumerate}
A constraint could be orthogonality and parallelism in walls, as well as planarity, however this may not be correct with older buildings.
To determine the constraints to use for each material, the International Organisation for Standardization (ISO) standards are used, for example, a door can be represented by the ISO19107 \cite{ISOStandard}. Constraints and rules for such objects are then extracted from the standard, with preconditions if necessary. 
Additional optional constraints could be:
\begin{itemize}
    \item Angular arrangements e.g. all corridors have the same turning angle
    \item Wall thickness operations e.g. all walls have the same thickness
    \item Aesthetic operations e.g. all doors are the same type and properties such as ISO 970102
\end{itemize}
The semantic validation algorithm using consistency grammar is thus shown in Alg. \ref{alg:ConsistencyGrammarSemVal Alg}.
\begin{algorithm}[H]
\caption{Consistency Grammar Semantic Validation Algorithm \cite{Nikoohemat2021ConsistencyChecking}}\label{alg:ConsistencyGrammarSemVal Alg}
\hspace*{\algorithmicindent} \textbf{Input:} Point Cloud e.g. XYZRGBA PCL point cloud \\
\hspace*{\algorithmicindent} \textbf{Output:} Validated 3D Model 
\begin{algorithmic}
\Require Consistency Knowledge, ISO Standards
\If{Consistency Grammar == Reject}
\While{(not stop criterion)}
  \State Geometry Derivation
  \State Topological Derivation
  \State Primary Model 
  \State Add Semantics \Comment{Input images, floor plans, sensor outputs}
  
\EndWhile
\State{Apply Consistency Grammar}
\ElsIf{Consistency Grammar == Succeed}
  \State{Validate 3D Model}
\EndIf
\end{algorithmic}
\end{algorithm}

The quantitative method \cite{Nikoohemat2021ConsistencyChecking} used to create a loss function when determining if the model is consistent with the grammar dictionary was as follows:
\begin{equation} \label{eq: ConsistencyGrammarLossFunction}
    E_i = E_{self}(i) + X_{j \in N(i)}E_{neighbour}(i,j) + X_{k \in V} E_{missing}(i,k)
\end{equation}

Where $E_{self}(i)$ is the self energy describing the instance fit for an object $i$, $E_{neighbour}(i,j)$ is the pair energy describing inappropriateness of connection between objects $i$ and neighbourhood objects $j$, $E_{missing}(i,k)$ is the pair energy describing possible missing connections of object i, classes $V = {wall, floor, ceiling, door, window, stair, room, furniture}$, and $E_{total} = \sum E_i$.
The Eq. \ref{eq: ConsistencyGrammarLossFunction} thus quantifies the greater energy meaning greater inconsistencies in the model. 

\chapter{Methodology} \label{Chap3}
\section{Data-sets}

In SfM, the ideal data-set for optimal reconstruction results is as follows:
\begin{itemize}
    \item Images captured should have good texture, and texture-less images should be avoided. For example, an empty desk, or white wall is texture-less. If a scene does not contain enough texture, additional background objects could be placed to improve texture.
    \item Images should be captured in similar illumination conditions, as high dynamic range scenes can cause errors in feature matching \cite{Irie2011ALocalization}. 
    Furthermore, specularities on shiny surfaces should be avoided \cite{Kersten1992InteractionMotion}.
    \item Images captured should have high visual overlap, showing each object in a minimum of three images. As a rule of thumb, the more images the better, although if using video input, down-sample frame rate to avoid a slow reconstruction process. Sample images at a rate where the maximum disparity between views is no more than about 64 pixels.
    \item Images should be captured from different viewpoints. Images should not be taken from a single viewpoint by only rotating the camera. Pioneers of the field of SLAM call the deliberate motion to initialise
     a SLAM algorithm the "SLAM wiggle" \cite{Davison2007MonoSLAM:SLAM}, a similar process is useful in SfM.
    \item COLMAP has the option for different camera models, if intrinsics are not known, the simplest camera model will suffice. However, simple pinhole or pinhole camera models should be used if images are un-distorted apriori, using $1$ and $2$ focal length parameters respectively. A radial camera model should be used if the intrinsics are unknown and every image has a different calibration (e.g. webscraped images), this models radial distortion effects. Fisheye models should be used if other models are incapable of modelling the distortion effects.
\end{itemize}
To achieve an optimal reconstruction, different camera models should be tried, generally, if the reconstruction fails and the estimated focal lengths are grossly wrong, its an indication that the camera model is too complex. Contrarily, if many iterative and global bundle adjustments are performed, its a sign that the model cannot sufficiently model distortion effects.
The Brighton Data-set was collected on an iPhone camera of fixed focal length $1536.00px$, $3024 \times 4032$ $60fps$, simple radial camera model, single camera iPhone 13 mini \cite{Apple2022AppleSpecifications}. The camera has  $f/2.44$ aperture and $120$ degree field of view. Due to hardware limitations, the data-set was down-sampled to $1280 \times 720 p$ and $5fps$ for computational ease.
The Brunswick Square, Brighton Data-set was chosen because it is particularly problematic in SfM, it has many repetitive architectural features and performs consistently poorly on conventional SfM without semantic segmentation integrated, such as COLMAP. The repetitive features causing erroneous matching can be seen in the Fig. \ref{fig:featurematcherroneous}.

\section{Implementation Details}
The following was performed on an Ubuntu 18.04 Linux system.
COLMAP is a general-purpose, open source Structure from Motion (SfM) \cite{Schonberger2016PixelwiseStereo} and Multi View Stereo (MVS) \cite{Schonberger2016Structure-from-MotionRevisited} pipeline with a GUI and CLI \cite{Schoenberger2022StructureCOLMAP}. The COLMAP C++ software offers a wide range of features for reconstruction of ordered and unordered frames of video or images. Alongside the COLMAP software, is the Ceres Solver \cite{AgarwalCeresLibrary}, an open source C++ library for modelling and solving large non linear optimisation problems, such as non linear least squares with bounded constraints, or general unconstrained optimisation problems. The Ceres Solver 2.0 is utilised in conjunction with the COLMAP software, as C++17 is required. Dependencies for the pipeline are detailed in the Appendix \ref{System Requirements}.

An initial sparse reconstruction was performed on a 4K 60Hz data-set acquired in Brighton, down sampled to 720p and 1Hz (1102 frames) for computational ease, using COLMAP. The dense reconstruction was performed, GPU accelerated, in Colab Pro in a Jupyter notebook, and by supervisors\footnote{Code for which can be found in git repository detailed in \ref{System Requirements}}. 
COLMAP offers two reconstruction methods, Poisson and Delaunay, both of these were performed on the data-set.
In the case of dense reconstruction, there is a trade-off between completeness and accuracy. Poisson surface reconstruction can create watertight surface meshes from sparse point cloud data \cite{Kazhdan2013ScreenedReconstruction, Kazhdan2006PoissonReconstruction}, and COLMAP also supports graph-cut based surface extraction from a Delaunay triangulation. Poisson reconstruction will typically require outlier-free input point cloud data and will produce bad surfaces otherwise, and may even leave holes in the surface. Delaunay triangulation based meshing is generally more robust to outliers in input point cloud data, and more scalable to large data-sets than the Poisson reconstruction method. Although the Delaunay triangulation method will produce less smooth surfaces than the Poisson algorithm. To improve the smoothness of the surfaces, a Laplacian smoothing \cite{Liu2017QualityIntersection} was performed in MeshLab \cite{VisualComputingLabMeshLabMeshes}, so to avoid arbitrarily complex geometry. There are shortcomings of both sparse and dense reconstructions, sparse reconstructions can be clustered around a small area of the image \cite{Ng2021UncertaintyNavigation}, or encounter problems with planar degeneracy \cite{Hartley2004Multiple11343}, resulting in a motion biased estimate.

Semantic segmentation was performed with the use of the pre-trained Tensorflow model DeepLabv3+, the model was pre-trained with the semantically labelled CityScapes Data-set \cite{Cordts2016CityscapesScenes}, giving semantic understanding of urban street scenes through transfer learning. An extract of the results are shown in Fig. \ref{fig:brighton_semseg2} \footnote{Code for which can be found in git repository detailed in \ref{System Requirements}}. Where transfer learning is a deep learning technique where a model trained on one task is re-purposed on a second related task, allowing for the improvement of learning in the new task through the transfer of knowledge from a related task that has already been learned \cite{Olivas2009HandbookTechniques}. This process is only viable if the model learned does not overfit, and does not have low bias, high variance. If the model exhibits low bias, high variance, it has overfit to the noisy or otherwise unrepresentative training data and thus cannot generalise to unseen data in transfer learning. However, if the model is too general (underfitting with high bias, low variance), it may overlook key regularities in the data.

\section{Overview}
The methodology to validate structure from motion models with semantic consistency constraints and ray tracing between cameras and points to check for planar occlusions is detailed in the steps below:
\begin{enumerate}
    \item Acquire a video data-set of the desired structure to reconstruct, being sure to get images captured in similar illumination, and captured from different viewpoints.
    \item Decompose the video into images using \texttt{ffmpeg} \cite{FFMPEG} and down-sample to a frame rate that still has a high overlap between images, however a down-sampled frame rate to avoid a slow reconstruction process.
    \item Run GPU accelerated DeepLab semantic segmentation model pre-trained on CityScapes data-set, applying transfer learning to the data-set, so to determine semantic labels in the input images.
    \item Run COLMAP SfM pipeline sparse reconstruction to acquire database.db output, as well as extrinsic and intrinsic parameters.
    \item Run Python script to extract SIFT keypoints and two view geometries from SQLite3 COLMAP output database and determine semantic label for each keypoint, based on DeepLab semantic segmentation output. Find the corresponding location in the segmented image, and determine class of the particular keypoint pixel in 2D. 
    \item Append keypoint class label to point cloud points in COLMAP SfM 3D model.
    \item Each identified keypoint that appears in two or more images, determine the most common semantic label, and discard, 2D keypoints that are inconsistent with the most common keypoint semantic label.
    
    \item If the resulting 3D point has fewer than two observations between frames, the 3D point cloud point is discarded entirely.
    
    \item Perform dense reconstruction using Delaunay triangulation from sparse point cloud.
    
    \item Perform planar surface reconstruction on the mesh from dense point cloud 3D model using Marching Cubes Algorithm and learned Signed Distance Function.
    
    \item Perform ray tracing from calculated camera position to calculated 3D point.
    
    \item If there is a plane of semantic label "wall" or "building" between the camera observation and 3D point, the point is labelled erroneous. As the prior knowledge is that walls are opaque, and no observations can be made behind them. Intersections in the plane between ray tracing of camera and point ($\pm$ some lateral tolerance as the point cloud is not continuous and it is unlikely that there are two features on the same linear trajectory from the camera).  \\
    \emph{Note}: Only invalidate points that rays are intersected by planes with labels that are opaque, for example, foliage or sky may form a plane in the dense reconstruction, but is transparent. Whereas wall or building is opaque, thus observations cannot be made behind them.
   
\end{enumerate}

\subsection{DeepLab Semantic Segmentation}

Semantic segmentation using Google's DeepLab \cite{Chen2016DeepLab:CRFs} was performed using the code in the repository details in Appendix. DeepLab labels each pixel of the input image with a class of object, given that it is trained on CityScapes data-set in our case. The DeepLab DCNN returns a segmentation map with each pixel labelled for each image. The segmentation map can then be used to look up the semantic class label of each identified feature, and corresponding 3D point cloud point, in the COLMAP SfM pipeline. A post processing step can then be performed to determine semantic consistency and opaque occlusions in the SfM model. \par
The integration of semantic segmentation into SfM models also offers the functionality to periodically map static structures; through the removal of dynamic objects and feature points identified within them in the 3D point cloud. The removal of motion acts as a post-processing stage to filter out data that is associated with moving objects. This can allow for robust change detection of the static scene throughout time \cite{Ku2021SHRECScenes}. Usually, the removal of unwanted features must be performed manually, this is labor-intensive and time consuming. So, to address this issue, we have developed a system to remove unwanted features of a particular semantic label. 
Most of existing RGB-D vSLAM or SfM methods assume that the traversed environments are static during the data acquisition process, a convenient assumption when moving objects in dynamic environments can severely degrade the vSLAM  or SfM performance \cite{Sun2017ImprovingApproach}.
The removal of dynamic objects was achieved through looking up the semantic label of the input images that contributed to the views of the 3D point, from this, the label can determine whether dynamic or not, and this acts a criteria for removal, or not. The proposed system of motion removal in SfM using semantics could potentially support the monitoring and inspection of infrastructure construction, as well as maintenance projects \cite{saovana2020development}. Projects such as these have numerous similar components, and would be simple to train a semantic segmentation algorithm and generalise our pipeline to remove other unwanted feature points.

\subsection{COLMAP Structure from Motion}
COLMAP provides a tool for automatic reconstructions that takes a collection of input images and produces sparse and dense reconstructions \cite{JohannesL.Schoenberger2022COLMAPDocumentation}. For the general user, COLMAP only requires a few steps, however to improve reconstructions, parameters can be modified to improve reconstruction quality. Although there is a trade-off between robustness, speed, and reconstruction quality. The quality of reconstruction parameters can be set at low, medium, high, or extreme; and this will determine the density of the point clouds in sparse reconstruction. COLMAP assumes all images are in a single directory, and can support various image types, see \cite{Drolon2022TheProject}.
To quick start run COLMAP GUI, run the pre-built binaries \texttt{COLMAP.bat} or execute \texttt{./src/exe/colmap gui} in the CLI, from the CMake build folder.
\par
COLMAP will output a project configuration file along with the reconstruction, this stored the absolute path information of the database and image folder, as well as specified parameters. The database file can be shared between reconstructions if multiple are being run. 
The first step of a reconstruction is the feature detection/extraction, the working of are detailed in Sec. \ref{sec:sfm}. COLMAP can automatically extract focal length information from embedded exchangeable image file format (EXIF) Information, or the intrinsic parameters can be manually specified. If partial EXIF Information is present in the images, COLMAP has a large database of camera models to find the rest of the camera model. If all images are captured withe the same model camera, and same zoom factor, shared intrinsics can decrease computation time. Although if webscraped images are used, and taken with different cameras, this is not suitable. 
COLMAP identifies SIFT features, although existing feature descriptors can be imported in text file format for each image\footnote{Note: COLMAP image coordinate convention is that $(0,0)$ is upper left corner of an image, and the centre of the upper left most pixel has coordinate $(0.5,0.5)$, decimal pixel coordinates are necessary for SIFT sub-pixel precision.}.


\subsection{Planar Surface Reconstruction}
Given a sparse point cloud $G \in \R^{G \times 3}$, the objective is to reconstruct the surfaces in the model. Once surfaces are reconstructed, we can use the DeepLab semantic labels on input images to label the planes, and determine opacity. From determining the planes and opacity, ray tracing can be performed from camera to observed point, and if there's an opaque planar occlusion, the point is erroneous. The planar identification is achieved by training and implementing the model in \cite{Liu2018PlaneRCNN:Imageb}.
Planar segmentation was attempted in the 2D images, using PlaneRCNN \cite{Liu2018PlaneRCNN:Imageb} CNN, to detect 3D planes from a single input image. This model was pre-trained on an indoor data-set ScanNet \cite{Dai2017ScanNet:Scenes}, although did not generalize well to outdoor data, as can be seen in Chap. \ref{Chap4}.

\subsection{Ray Tracing, Priors, and Camera Projections}
To determine whether there are planar occlusions between the camera and observed points, the camera coordinates and projections must be calculated. COLMAP outputs a reconstructed pose of an image, specified by the projection from world to camera coordinate system, using a quaternion (defined using Hamilton convention) $(QW,QX,QY,QZ)$ and a translation vector $(TX,TY,TZ)$. Thus, the coordinates of the camera centre can be given by $-R^\top \times T$, where $R^\top$ is the transpose of the $3 \times 3$ rotation matrix composed from the quaternion, and $T$ is the translation vector. The local camera coordinates in COLMAP are of a left hand coordinate system. To determine planar occlusions, the semantic label of the plane must be known, as it is only defined as an occlusions if the plane is opaque.
A planar occlusion is defined by a ray from the camera to the point be intersected by an opaque semantically labelled plane, as points behind this are occluded and must be erroneous. An assumption in this model is that there are not observations behind the planes, e.g. the user has not walked around and mapped behind the wall.
For example, planes may form in the dense reconstruction from foliage or sky, which are transparent, and observations may be made behind them. Conversely, if the plane has a semantic label of wall or building, the observations behind the plane can be labelled erroneous. The point can be determined as behind the plane if the ray traced vector from camera to point intersects the plane.
The planar surfaces constructed from dense reconstruction can be represented by a normal vector and a point on the plane $(p - p_0)\cdot n = 0$, where $n$ is the normal (perpendicular) vector to the plane, and $p_0$ is the point on the plane \cite{ComputationalBooks}. The normal can be calculated through the cross product of two vectors in the plane. The intersection of a line and a plane in 3D can be represented by lemma $d = \frac{(p_0 -l_0)\cdot n}{(l \cdot n)} $, where $d$ lies on line and plane. If $l$ is perpendicular to the normal of the plane, $l \cdot n = 0$, and there is no intersection.
\par
The ray tracings between camera and points are exhaustively checked to intersect with every plane present in the model. An interesting future work to increase computational efficiency could be to utilise binary space partitioning (BSP) trees into the planar intersection checking. 
BSP can be implemented for recursively subdividing a space into two convex sets by implementing hyperplanes as partitions, thus reducing the number of planes needed to verify ray intersection or not.

\chapter{Results and Analysis} \label{Chap4}
\section{Structure from Motion}
\begin{figure}[ht]
    \centering
    \includegraphics[width=\textwidth]{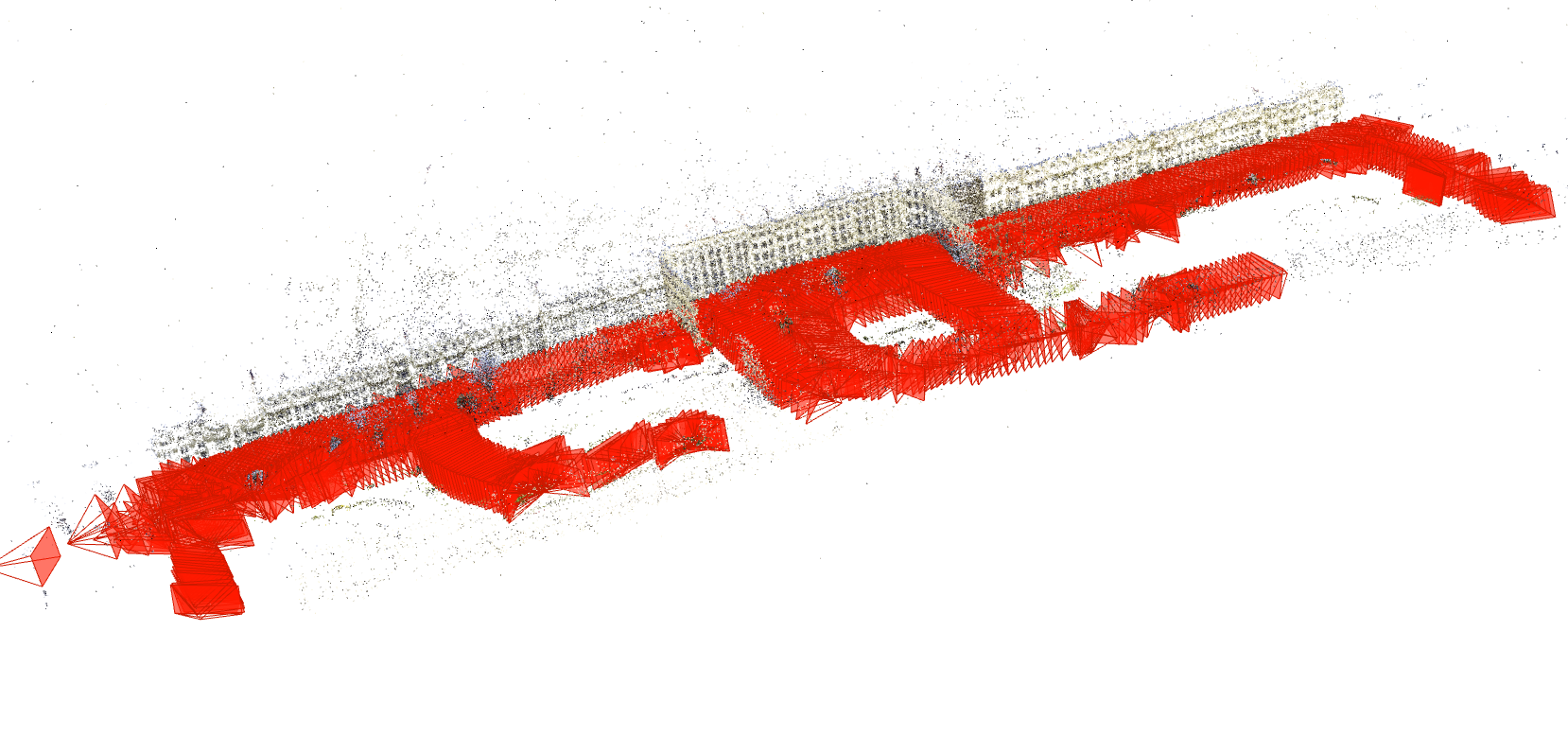}
    \caption{Brighton data-set COLMAP sparse reconstruction}
    \label{fig:brighton_sparse} 
\end{figure}

\begin{figure}[ht]
    \centering
    \includegraphics[width=\textwidth]{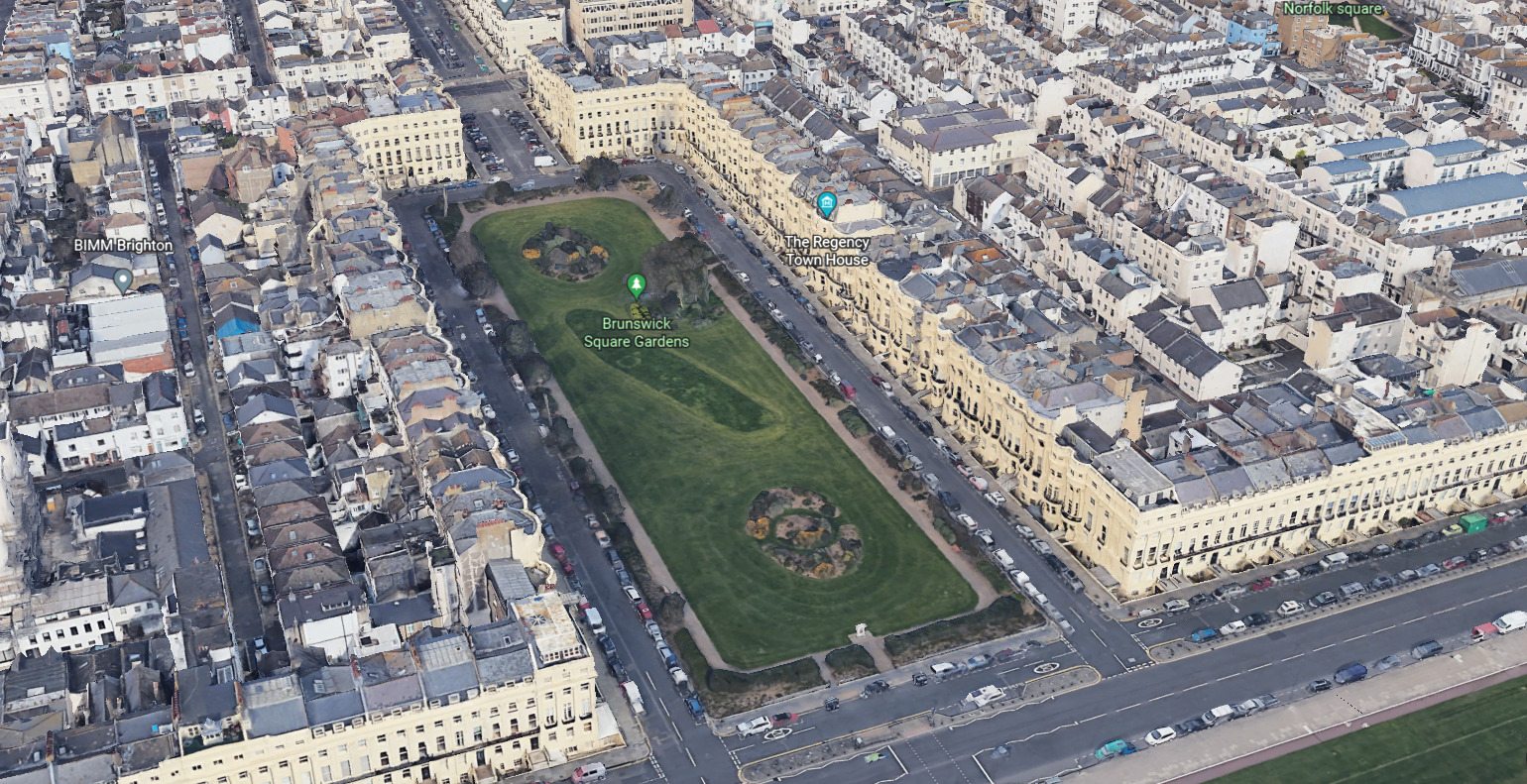}
    \caption{Brunswick Square, Brighton aerial view \cite{GoogleGoogleEarth}}
    \label{fig:brunswickSquareAerialView}
\end{figure}

The initial sparse reconstructions of the Brighton data-set are shown in Fig. \ref{fig:brighton_sparse}, this clearly shows both sides of the square mapped on to one side of the street, an error due to the similar features each building presents. A key indicator of this is that there are observations of walls, behind other walls \footnote{The full results and database of the reconstruction can be found here: \url{https://drive.google.com/drive/folders/11mBhHCv-USc8lXueiiMFWsQG4jC8y6st?usp=sharing}}. The problem that has occurred is in the feature matching step, due to the repetitive similar features, features have been mapped together erroneously, producing an incorrect model in which both sides of the square have been mapped to the same side. As a qualitative ground truth, an aerial view of Brunswick Square is shown in Fig. \ref{fig:brunswickSquareAerialView}, sourced from Google Earth.

\begin{figure}[ht]
    \centering
    \includegraphics[width=\textwidth]{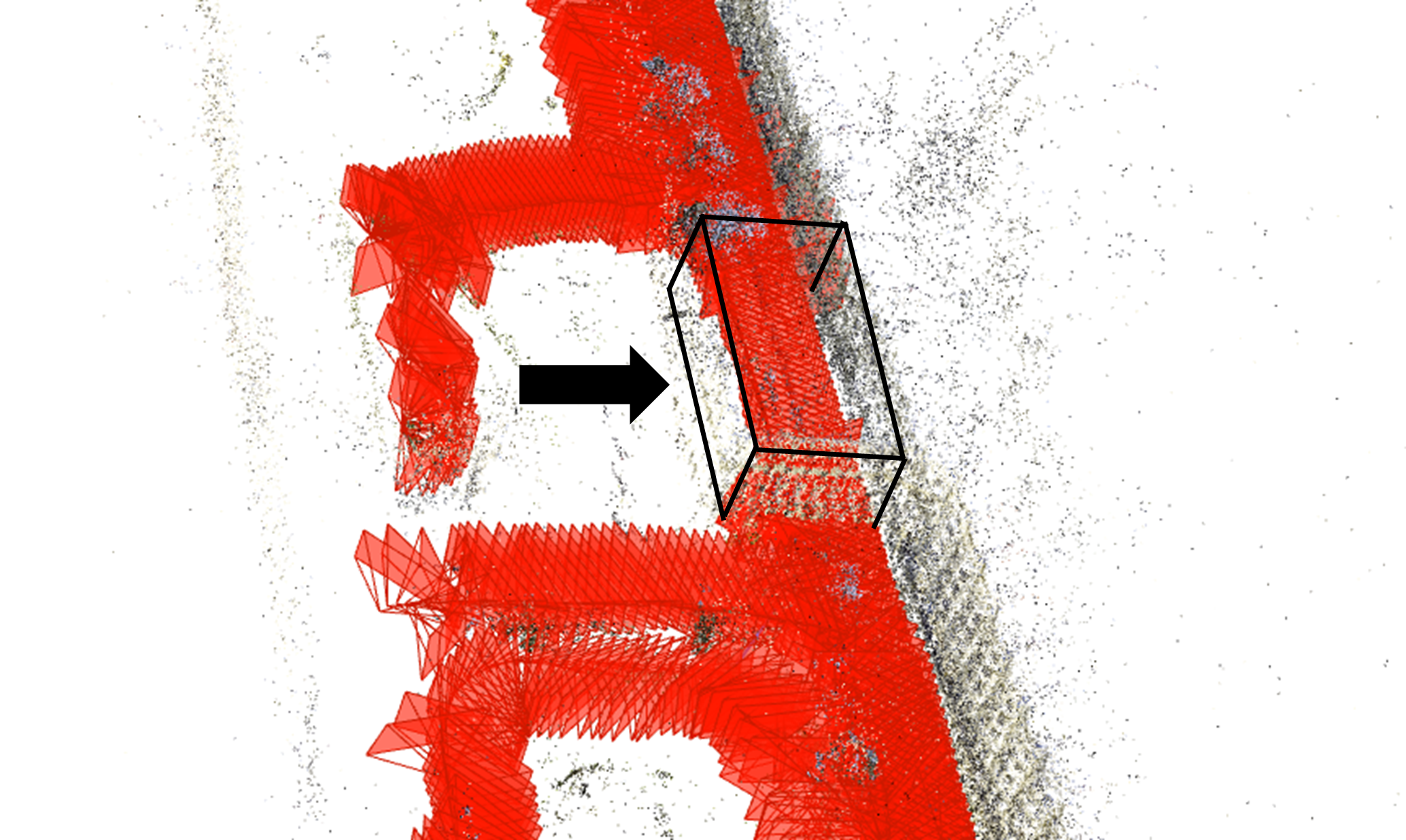}
    \caption{Brighton data-set COLMAP sparse reconstruction close up}
    \label{fig:brighton_sparse_close_up}
\end{figure}

\begin{figure}[H]
     \centering
     \begin{subfigure}{0.49\textwidth}
         \centering
         \includegraphics[width=\textwidth]{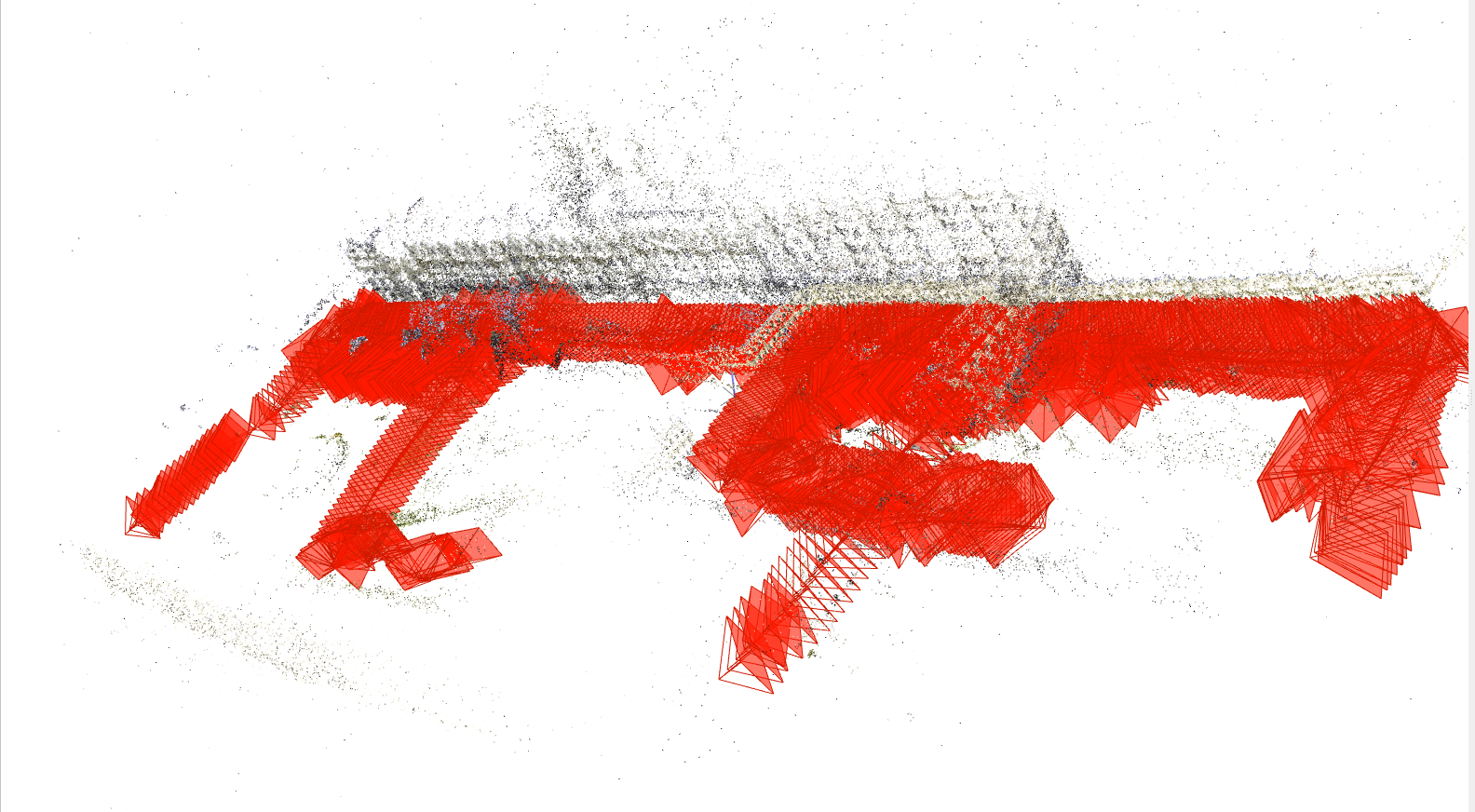}
         \caption{\centering Brunswick Square run 1}
         \label{fig:run1}
     \end{subfigure}
     \hfill
     \begin{subfigure}{0.49\textwidth}
         \centering
         \includegraphics[width=\textwidth]{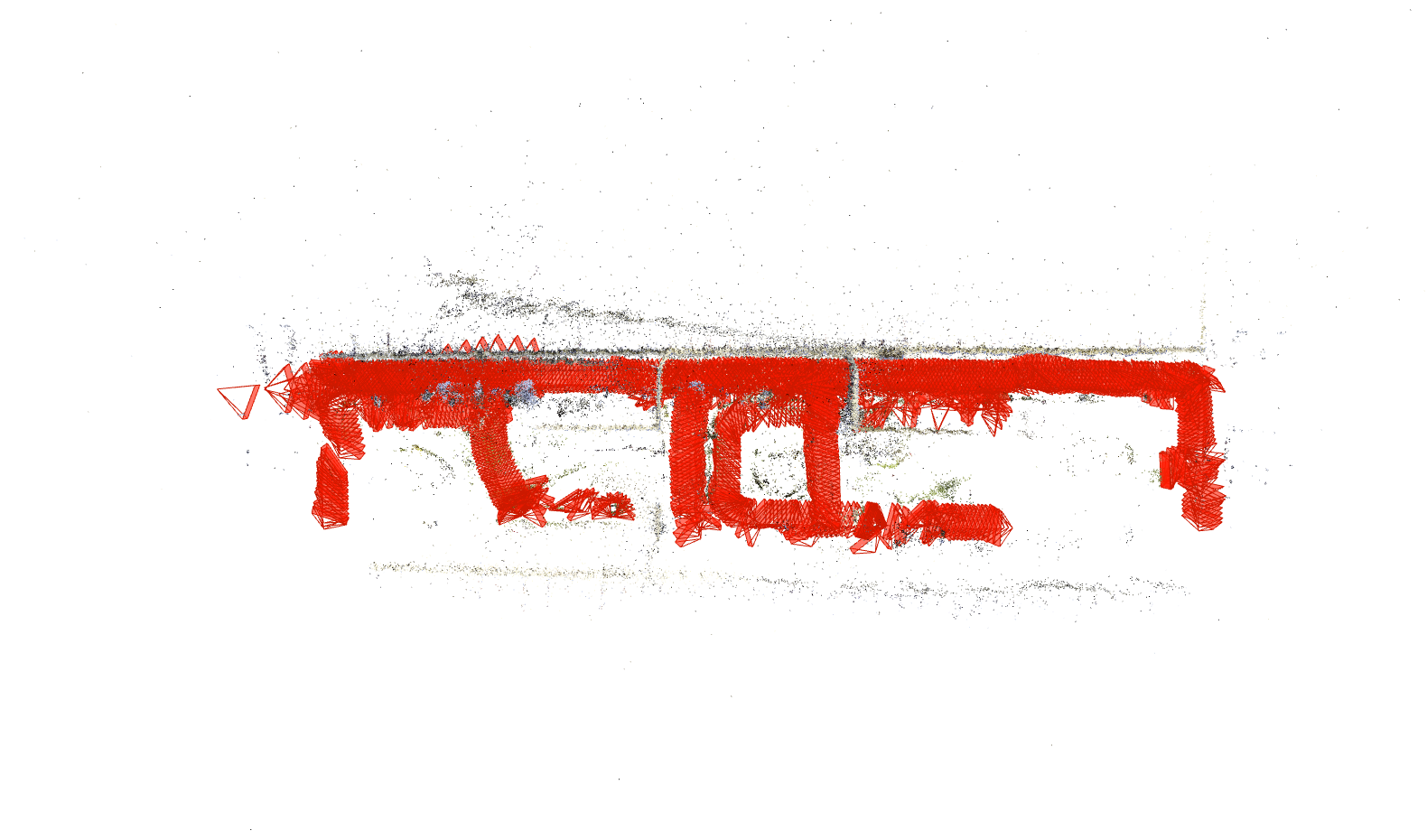}
         \caption{Brunswick Square run 2}
         \label{fig:run2}
     \end{subfigure}
     \hfill
     \begin{subfigure}{0.49\textwidth}
         \centering
         \includegraphics[width=\textwidth]{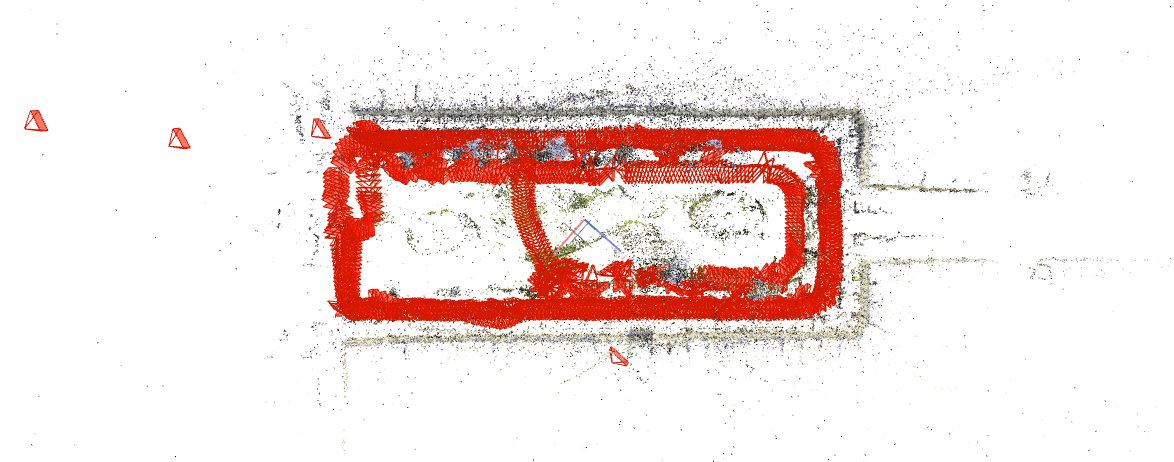}
         \caption{Brunswick Square run 3}
         \label{fig:run3}
     \end{subfigure}
     \hfill
     \begin{subfigure}{0.49\textwidth}
         \centering
         \includegraphics[width=\textwidth]{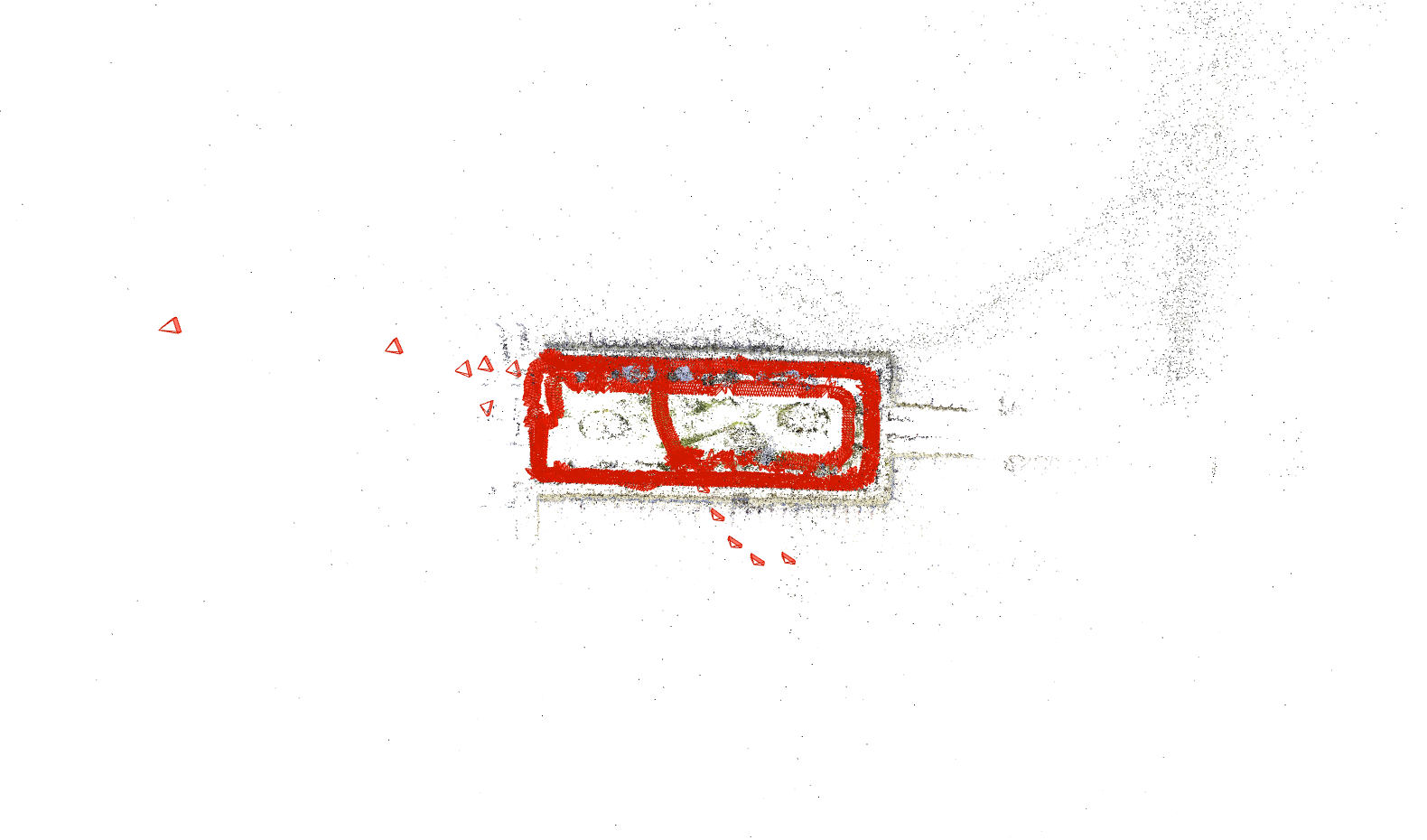}
         \caption{Brunswick Square run 4}
         \label{fig:run4}
     \end{subfigure}
        \hfill
     \begin{subfigure}{1\textwidth}
         \centering
         \includegraphics[width=\textwidth]{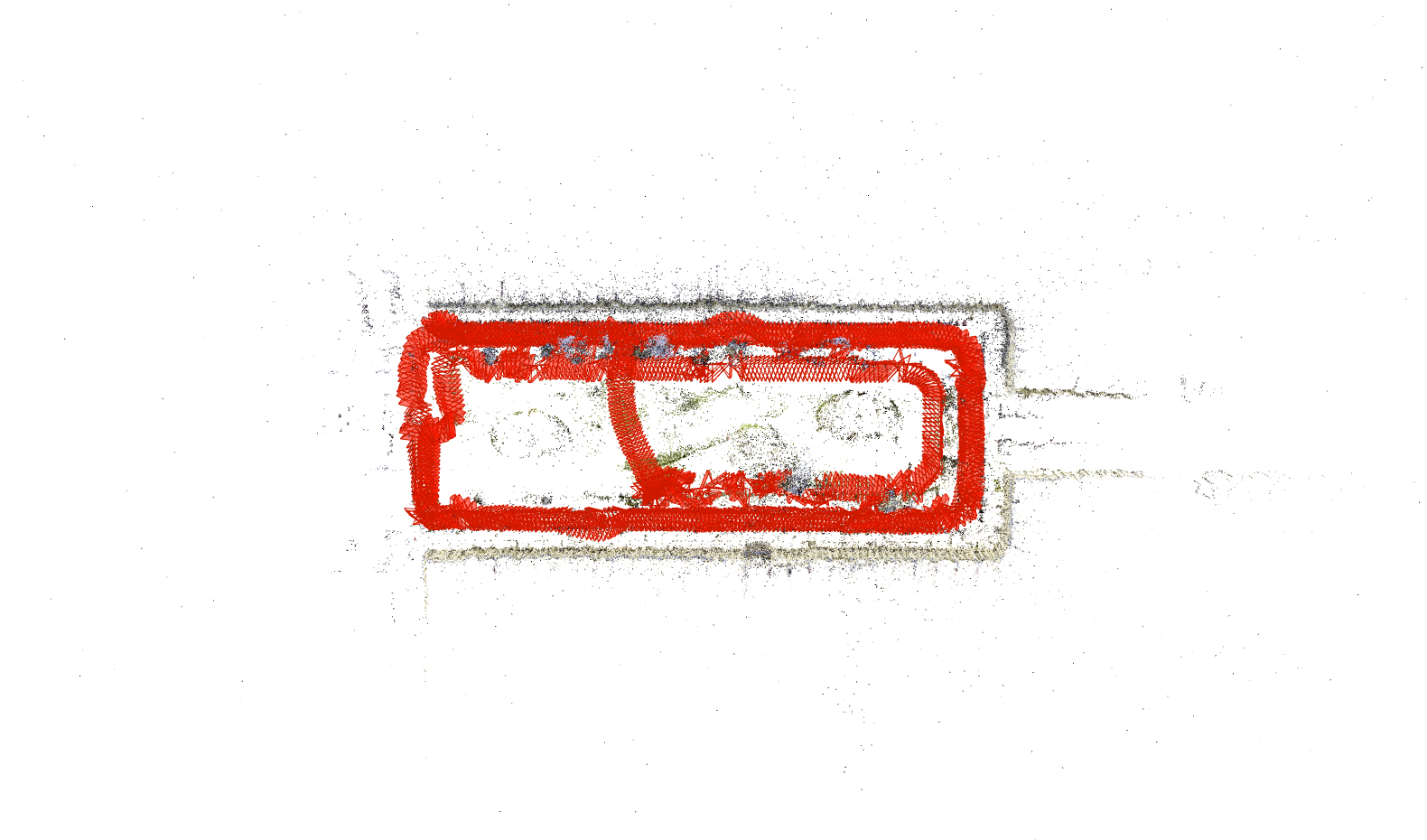}
         \caption{Brunswick Square run 5}
         \label{fig:run5}
     \end{subfigure}
    \caption[Variation between COLMAP 3D sparse reconstructions]{Variation between COLMAP 3D sparse reconstructions of Brunswick Square, Brighton,  given the same input data}
        \label{fig:colmapVariation}
\end{figure}

An alternative angle of the initial sparse reconstruction of the Brighton data-set is shown in Fig. \ref{fig:brighton_sparse_close_up}, showing clearly the incorrect reconstruction, a view of a wall and a view behind that wall. 
There is great variation in the reconstructed model given the same input data, using the same software and parameters. Examples of COLMAP being run several times with different outputs are shown in Fig. \ref{fig:colmapVariation}. The inconsistencies between reconstructions, given the same data, enforces the necessity for  a validation pipeline.

The SfM pipeline COLMAP was run with both sparse and dense reconstruction several times, each with varying degrees of success. The great variation in reconstruction results, given the same input data necessitates a validation algorithm.

The reconstruction statistics from each of the Brunswick Square Brighton data-set are shown in Table \ref{tab:BrightonStat}. Each of the reconstruction files can be found on the project Google Drive.
  

\begin{table}[H]
  \caption{\centering Brighton Data-set Reconstructions Statistics for Raw Sparse Models}\label{tab:BrightonStat}
  \small
  \centering
  \begin{tabular}{lccccr}
  \toprule[\heavyrulewidth]\toprule[\heavyrulewidth]
  \textbf{Statistic} & \textbf{1} & \textbf{2} & \textbf{3} & \textbf{4}  \\
  \midrule
  Cameras &  1102 & 1099 & 1102 & 1102  \\
  \hdashline
  Images  & 1102 & 1099 & 1102 & 1102  \\
  \hdashline
  Registered Images & 1102 & 1099, & 1102 & 1102  \\
  \hdashline
  Points  & 339,029 & 335,422  & 338,652 & 272,017  \\
  \hdashline
  Observations & 1,910,495 & 1,869,222 & 1,929,064 & 1,611,163  \\
  \hdashline
  Mean Track Length &5.6352 & 5.57275& 5.6963 & 5.92302  \\
  \hdashline
  Mean Observations per Image & 1733.66& 1700.84 & 1750.51 & 1462.04  \\
  \hdashline
  Mean Re-projection Error & 0.63351& 0.613023& 0.621533 & 0.633162  \\
  \bottomrule[\heavyrulewidth] 
  \end{tabular}
\end{table}

  

\begin{figure}[H]
     \centering
     \begin{subfigure}{0.33\textwidth}
         \centering
         \includegraphics[width=\textwidth]{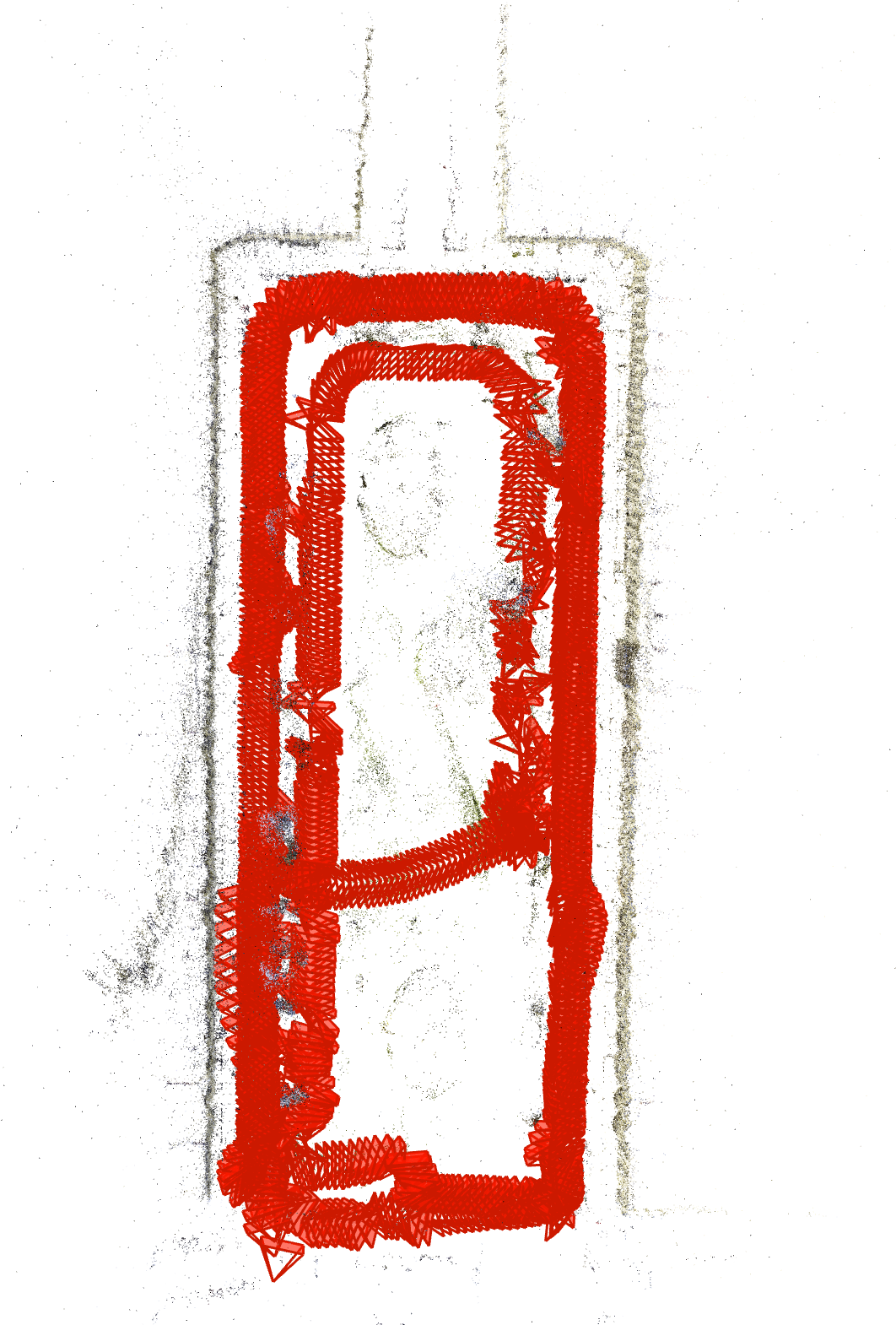}
         \caption{Original: 338,652 points}
         \label{fig:20220828_original_2}
     \end{subfigure}
     \begin{subfigure}{0.33\textwidth}
         \centering
         \includegraphics[width=\textwidth]{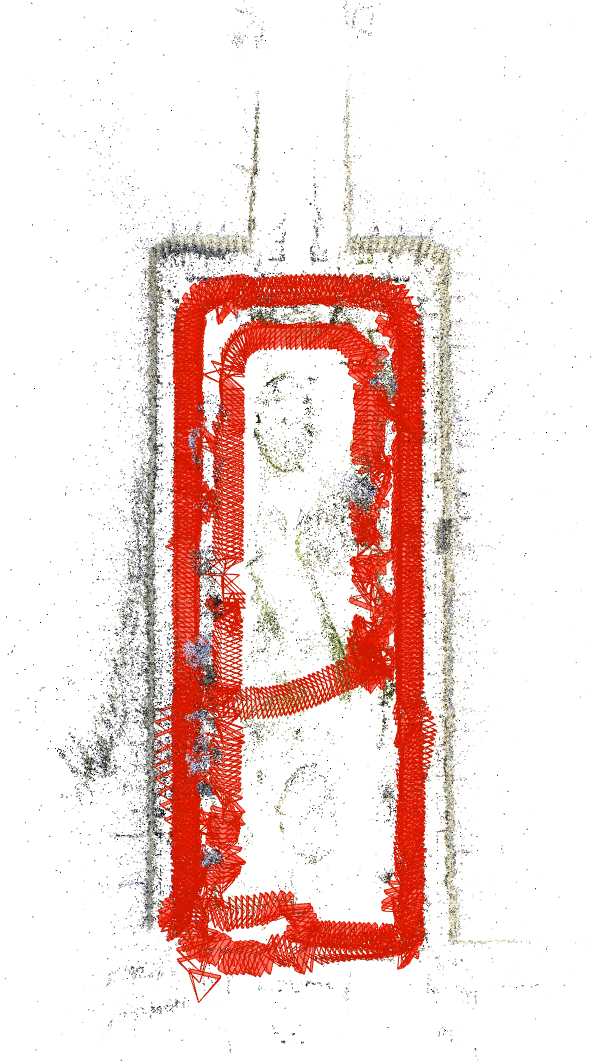}
         \caption{\centering Semantically Consistent: 227,286 points}
         \label{fig:20220828_corrected}
     \end{subfigure}
    \caption[Semantic Consistency Brunswick Square Sparse Models]{Semantic Consistency Brunswick Square Sparse Models}
        \label{fig:corrected20220828}
\end{figure}

\begin{table}[H]
  \caption{\centering Brighton Data-set Reconstructions Statistics for Semantic Consistency Validated Sparse Models}\label{tab:BrightonStatConsistency}
  \small
  \centering
  \begin{tabular}{lccccr}
  \toprule[\heavyrulewidth]\toprule[\heavyrulewidth]
  \textbf{Statistic} & \textbf{1} & \textbf{2} & \textbf{3} & \textbf{4}  \\
  \midrule
  Cameras &  1102 & 1099 & 1102 & 1102  \\
  \hdashline
  Images  & 1102 & 1099 & 1102 & 1102 \\
  \hdashline
  Registered Images & 1102 & 1099 & 1102 & 1102  \\
  \hdashline
  Points  & 226,623 & 223,567  & 227,286 & 189,484  \\
  \hdashline
  Mean Track Length &8.43028 & 8.3609 & 8.48739 & 8.5029  \\
  \hdashline
  Mean Re-projection Error & 0.63152 & 0.606963& 0.620227 & 0.642298\\
  \hdashline
  \makecell{Semantic Consistency Constraint \\ Violation Points} &112,406 & 111,855& 111,366& 82,533\\
  \bottomrule[\heavyrulewidth] 
  \end{tabular}
\end{table}
After performing the semantic consistency check and discarding observations that were semantically inconsistent, and discarding points entirely if there were less than 2 observations in total, the number of 3D points in one of the sparse models was reduced from $272,017$ to $189,484$, with mean track length increasing from $5.923$ to $8.5029$. The total number of points that violated the semantic consistency constraint was $82,533$ in the model 4, Tab.  \ref{tab:BrightonStatConsistency}. Although the semantic consistency constraint removed points that were inconsistent, this did however increase the mean track length, as removing points in a sparse reconstruction only makes it more sparse. Tab. \ref{tab:BrightonStatConsistency} shows the points that violate the semantic consistency constraint and the reduction in number of points in the models. One of the semantically consistent corrected sparse reconstructions is shown in Fig. \ref{fig:corrected20220828}.

\begin{figure}[H]
    \centering
    \includegraphics[width=0.75\textwidth]{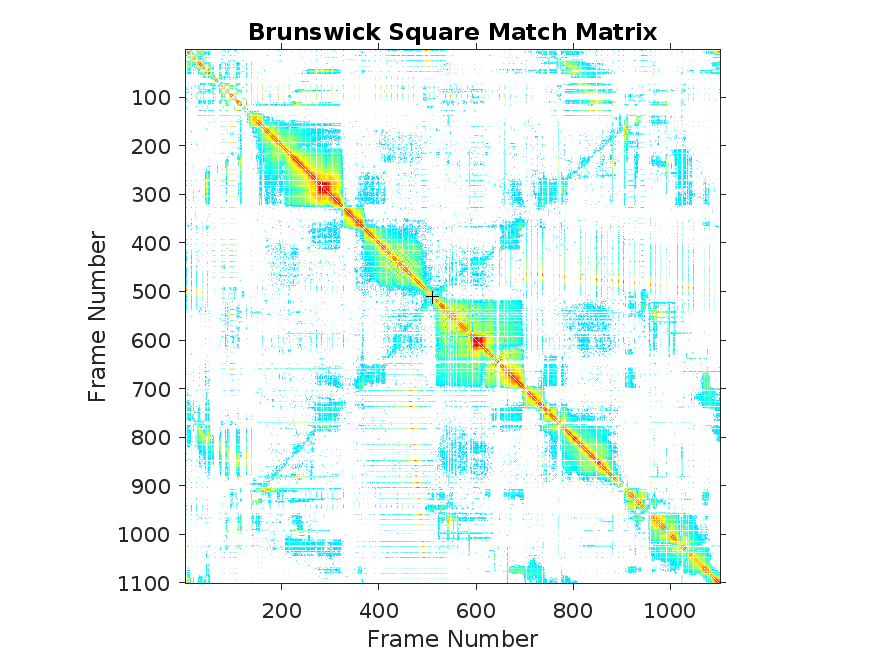}
    \caption{Brighton data-set match matrix}
    \label{fig:brighton_match_matrix}
\end{figure}

The match matrix (adjacency matrix) in Fig. \ref{fig:brighton_match_matrix} illustrates the number of feature matching occurrences between \emph{ordered}\footnote{SfM does not require ordered frames to function, however a match matrix is only visibly meaningful with ordered images.} frames in the Brunswick Square data-set. The colourmap indicated the density of the matches, and the axis is frame number. Some interesting artefacts in the matrix can be observed; for example, at frame $\sim 500$ there is a sudden lack of matching between immediately adjacent frames and an abundance of matches in other frames, this occurs as the point that when the data was acquired, the camera turns to face the opposite side of the square, thus there are no matches with the consecutive frames, and many matches with other observed frames briefly. 

\begin{figure}[H]
    \centering
    \includegraphics[width = 0.7\textwidth]{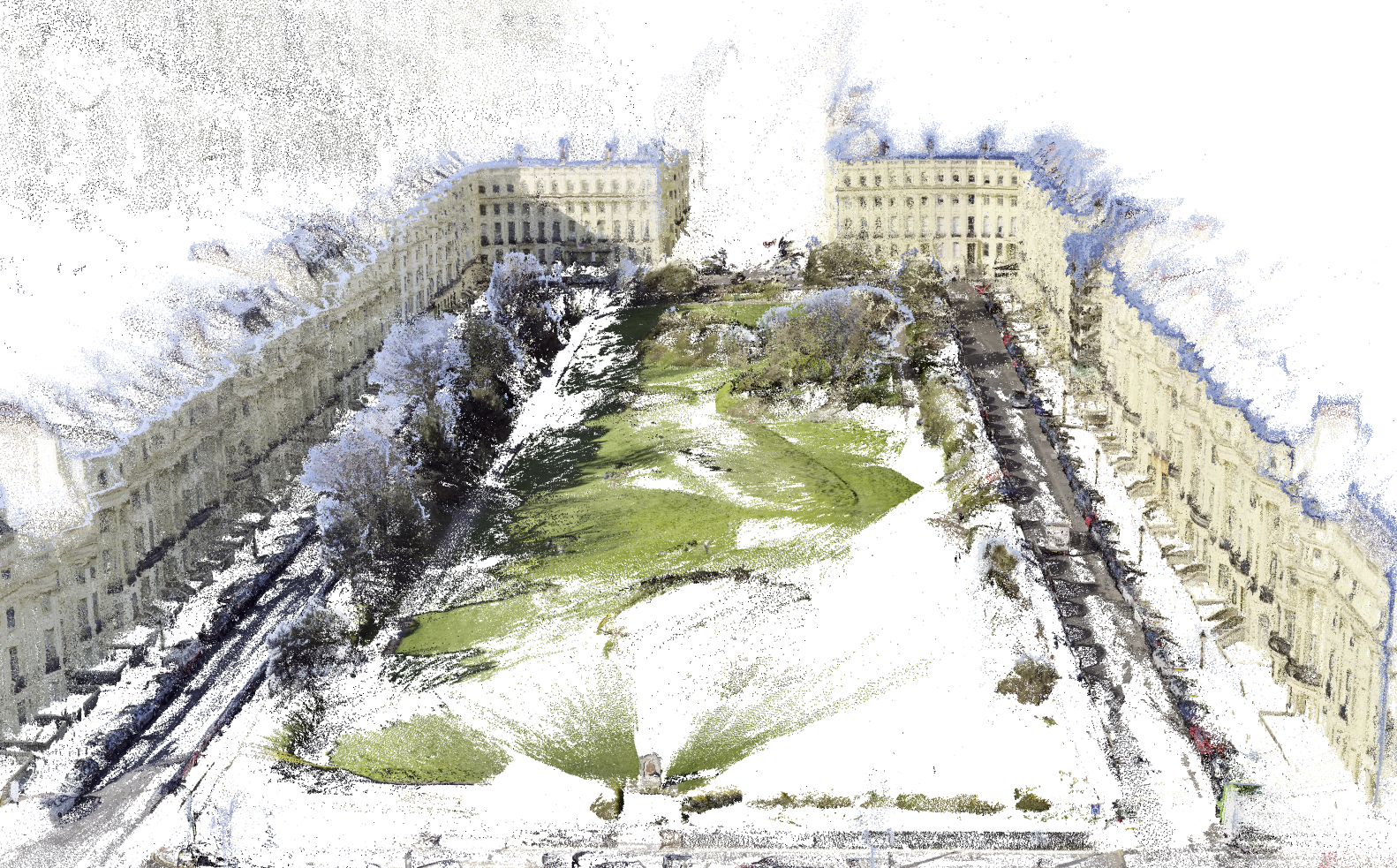}
    \caption{Brunswick Square, Brighton, dense reconstruction view 1}
    \label{fig:dense1}
\end{figure}

\begin{figure}[H]
    \centering
    \includegraphics[width = 0.7\textwidth]{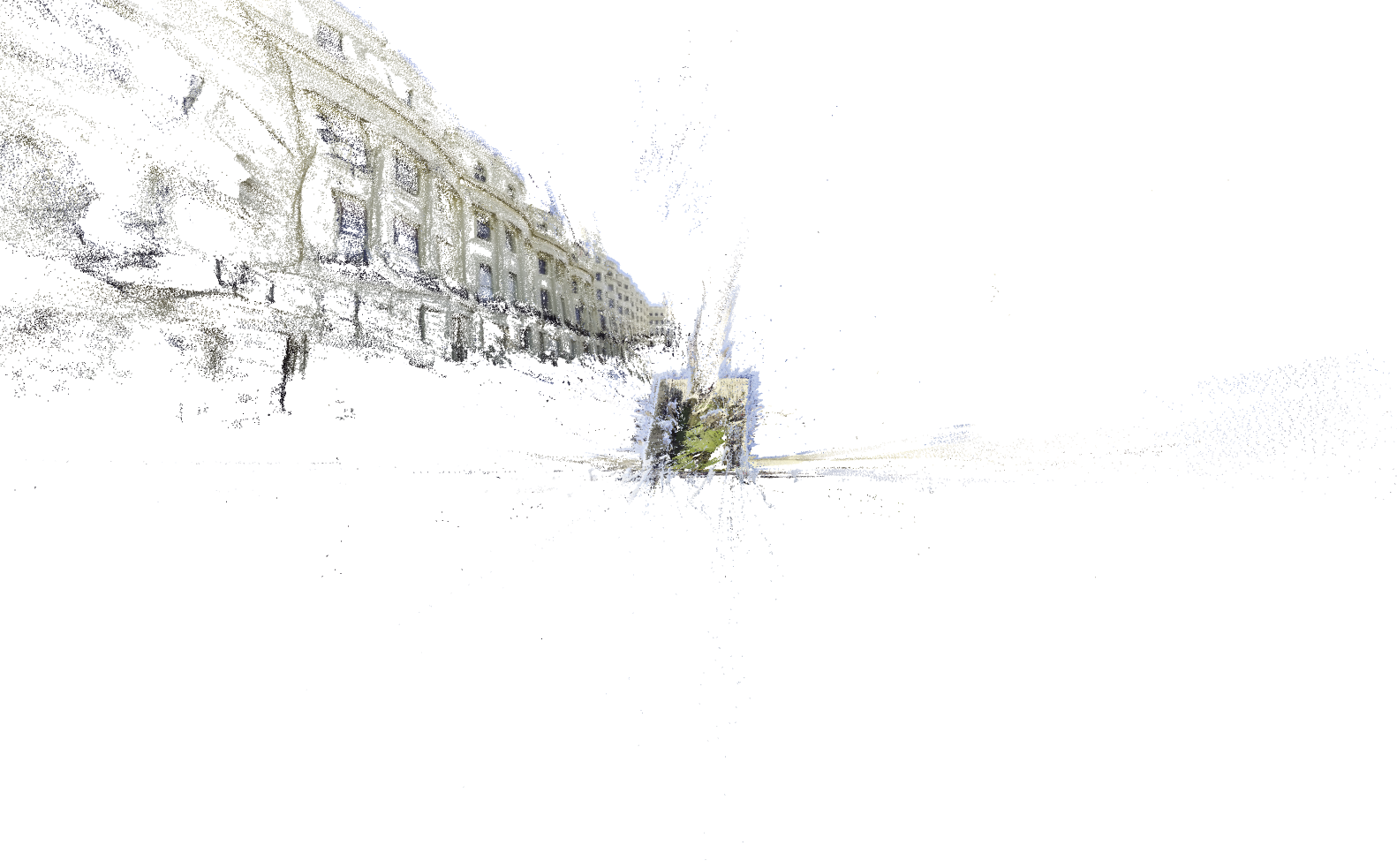}
    \caption{Brunswick Square, Brighton, dense reconstruction view 2, showing erroneous points}
    \label{fig:dense1view2}
\end{figure}

One of the dense reconstructions is shown in Fig. \ref{fig:dense1}, and at first glance, the model seems visibly correct, Fig. \ref{fig:dense1view2} shows the same model zoomed out, showing how COLMAP has built upon an error in feature matching early on in the reconstruction process. Many of the erroneous points are occluded by planes, and this is a prime example of when a validation model is required, to show a constraint violation.

\section{Semantic Segmentation}
\begin{figure}[ht]
     \centering
     \begin{subfigure}{\textwidth}
         \centering
         \includegraphics[width=\textwidth]{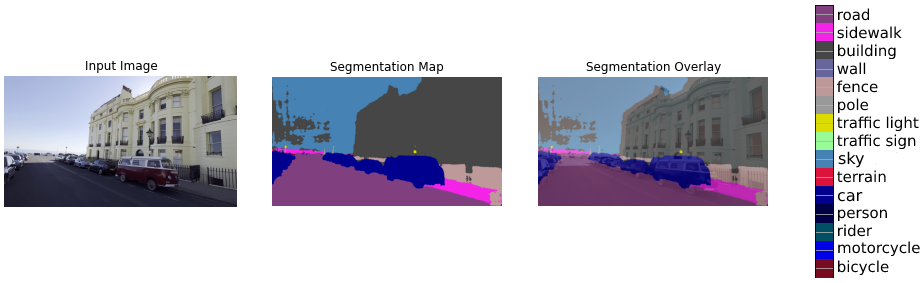}
         \caption{\centering Example frame 1}
         \label{fig:brighton_semseg2}
     \end{subfigure}
     \hfill
     \begin{subfigure}{\textwidth}
         \centering
         \includegraphics[width=\textwidth]{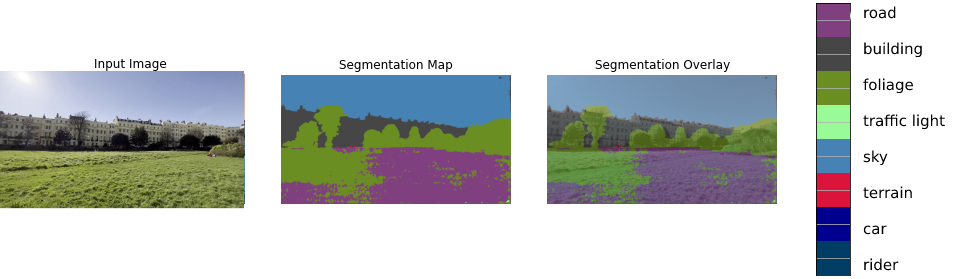}
         \caption{Example frame 2}
         \label{fig:brighton_semseg3}
     \end{subfigure}
     \caption[Brighton data-set DeepLabv3+ semantic segmentation results]{Brighton data-set DeepLabv3+ semantic segmentation results. (a) shows the DeepLab model was trained on suitable training data (CityScapes Data-set). While (b) shows the model doesn't generalise quite as well to nature scenes. }\label{fig:brighton_semseg4}
\end{figure}

The semantic segmentation algorithm DeepLabv3+ \cite{Chen2018Encoder-DecoderSegmentation} was used with the Brighton data-set and an extract of the semantic segmentation output is shown in Fig. \ref{fig:brighton_semseg2}, the model used was trained on the CityScapes data-set \cite{Chollet2017Xception:Convolutions}, the classes of which are shown in Table \ref{tab:cityscapesClasses}, and applied via transfer learning. The CityScapes data-set focuses on semantic understanding of urban street scenes, trained on 50 cities, throughout several months (Spring, Summer, Autumn) during daytime. The pre-trained model was applied via transfer learning, and generalised well to the street scenes seen in the Brunswick Square data-set. However, Fog. \ref{fig:brighton_semseg3} shows that the pre-trained model does not generalise as well to nature scenes, as grass is identified as road. This is a shortfall of using the CityScapes data-set, as even in real life, there is rarely a complete separation of street scenes and nature scenes. 
Semantic segmentation was performed on every frame of the Brighton Data-set video, after downsampling for computational ease, the results of which are posted on YouTube\footnote{The full video segmentation performed on Brighton Data-set using DeepLabv3+ pre-trained on CityScapes Data-set can be found at \url{https://www.youtube.com/watch?v=UwfRyR7IwWU}}.

To assess performance, the standard Jaccard Index, commonly known as the PASCAL Visual Object Classes (VOC) intersection-over-union metric (IoU) Eq. \ref{eq:IoU} \cite{Everingham2015TheRetrospective} is used, where $TP$, $FP$, $FN$ are the numbers of true positive, false positive and false negative pixels respectively. In which DeepLabv3+ achieves an $IoU_{class}=82.1$, $IoU_{category}=92.0$ on the CityScapes benchmark data-set in pixel level labelling.
\begin{equation}\label{eq:IoU}
    IoU= \frac{TP}{(TP+FP+FN)}
\end{equation}

\begin{figure}[H]
     \centering
     \begin{subfigure}{0.33\textwidth}
         \centering
         \includegraphics[width=\textwidth]{Figures/20220828_original.png}
         \caption{Original: 338,652 points}
         \label{fig:20220828_original}
     \end{subfigure}
     \begin{subfigure}{0.33\textwidth}
         \centering
         \includegraphics[width=\textwidth]{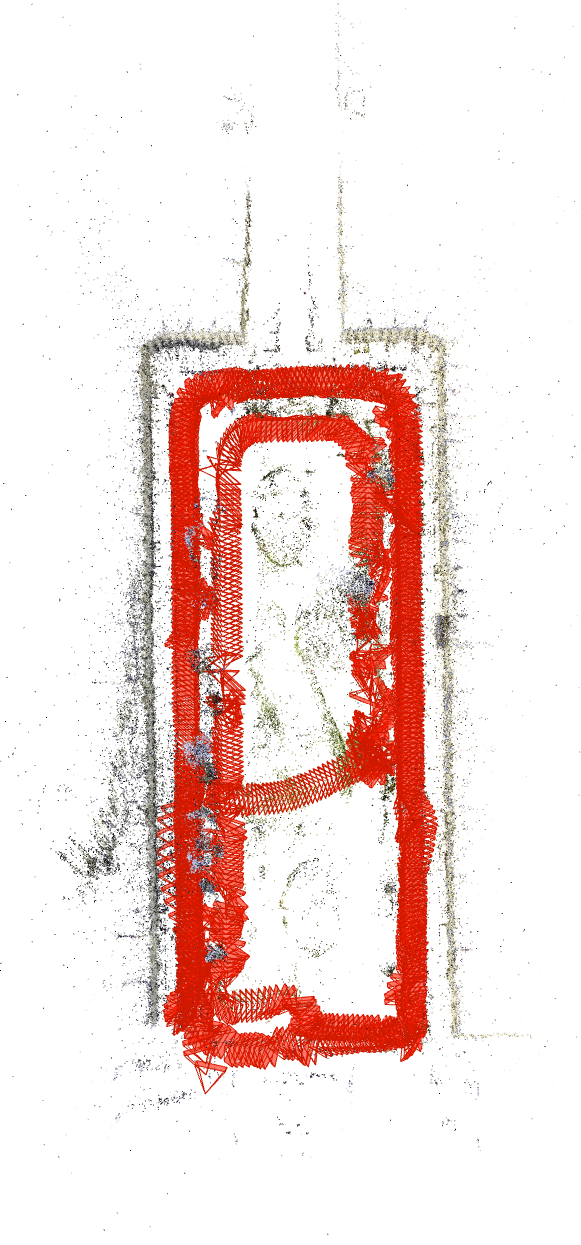}
         \caption{\centering Motion Removed: 315,650 points}
         \label{fig:20220828motion_removed}
     \end{subfigure}
    \caption[Motion removed Brunswick Square Sparse Models]{Motion removed Brunswick Square Sparse Models. 23,002 points associated with dynamic objects removed.}
        \label{fig:motionRemoved20220828}
\end{figure}

The segmentation map shows good segmentation of cars, buildings, fences and sidewalks, however struggles with sky and has a noisy result. This may cause errors in semantic consistency between detected feature points. Although sky is notoriously difficult for mapping as it introduces features that are temporally inconsistent and thus useless for static mapping. A useful application of applying semantic segmentation in SfM is also removing features identified on dynamic objects, so to improve the robustness to real-world dynamic scenarios. This motion removal was performed on the data-set to remove feature points labelled 'sky', 'person', 'car', 'truck', 'train', 'rider', 'motorcycle', 'bicycle'; any semantic class associated with motion. The removal of dynamic objects in mapping can allow for periodic mapping with robust change detection of the static environment.  The results demonstrated that the proposed motion removal approach was able to effectively improve SfM in otherwise challenging dynamic environments. The results of motion removal are shown in Tab. \ref{tab:BrightonMotionStat}, and one of the reconstructions shown in Fig. \ref{fig:motionRemoved20220828}. 

\begin{table}[H]
  \caption{\centering Brighton Data-set Reconstructions Statistics for Motion Removed Sparse Models}\label{tab:BrightonMotionStat}
  \small
  \centering
  \begin{tabular}{lccccr}
  \toprule[\heavyrulewidth]\toprule[\heavyrulewidth]
  \textbf{Statistic} & \textbf{1} & \textbf{2} & \textbf{3} & \textbf{4}  \\
  \midrule
  Cameras &  1102 & 1099 & 1102 & 1102  \\
  \hdashline
  Images  & 1102 & 1099 & 1102 & 1102  \\
  \hdashline
  Registered Images & 1102 & 1099, & 1102 & 1102  \\
  \hdashline
  Points  & 315,949 & 312,478  & 315,650 & 254,202  \\
  \hdashline
  Observations  & 1,783,849 & 1,745,884  & 1,803,042 & 1,506,703  \\
  \hdashline
  Mean Track Length & 6.04685& 5.98193& 6.1114 & 6.33812  \\
  \hdashline
  Mean Observations per Image & 1618.74 &1588.61 & 1636.15 & 1367.24\\
  \hdashline
  Mean Re-projection Error & 0.634635& 0.613145& 0.622501 & 0.635605  \\
    \hdashline
  Motion Points Removed & 23,080 & 22,944 & 23,002 & 17,815  \\
    \hdashline
  Observations Removed & 126,646 & 123,338 & 126,022 &  104,460 \\
  \bottomrule[\heavyrulewidth] 
  \end{tabular}
\end{table}
The Tab. \ref{tab:BrightonMotionStat} shows the results of each model having feature points associated with dynamic objects removed. Dynamic objects are user specified in the code, in this case they are 'sky', 'person', 'rider', 'car', 'truck', 'bus', 'train', 'motorcycle' or 'bicycle'. Although, the pipeline is capable of removal of any semantically labelled class on the input images for unwanted feature point removal. Each model has roughly $\approx 20,000$ points removed, with each point appearing in $\approx 5-6$ images/observations. The removal of points as a post-processing step is shown to increase the mean track length between points, as the resulting point cloud is sparser, as well as very slightly increasing the mean reprojection error.

\begin{table}[h]
  \caption{\centering CityScapes Data-set Classes \cite{Cordts2016CityscapesScenes} }\label{tab:cityscapesClasses}
  \small
  \centering
  \begin{tabular}{lcccr}
  \toprule[\heavyrulewidth]\toprule[\heavyrulewidth]
  
  \textbf{Class} & \textbf{Member} \\
  \midrule
  flat & road, sidewalk, parking, rail track\\
  \hdashline
  human & person, rider\\
  \hdashline
  vehicle & car, truck, bus, on rails, motorcycle, bicycle, caravan, trailer\\
  \hdashline
  construction & building, wall, fence, guard rail, bridge, tunnel \\
  \hdashline
  object & pole, pole group, traffic sign, traffic light\\
  \hdashline
  nature & vegetation, terrain\\
  \hdashline
  sky & sky\\
  \hdashline
  void & ground, dynamic, static\\
  \bottomrule[\heavyrulewidth] 
  \end{tabular}
\end{table}

\section{Planar Reconstruction}
\begin{figure}[H]
     \centering
     \begin{subfigure}{0.49\textwidth}
         \centering
         \includegraphics[width=\textwidth]{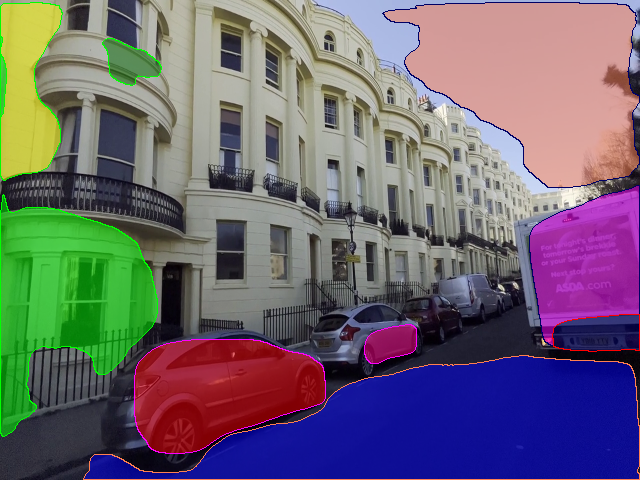}
         \caption{\centering Example frame 1}
         \label{fig:planar_segmentation}
     \end{subfigure}
     \hfill
     \begin{subfigure}{0.49\textwidth}
         \centering
         \includegraphics[width=\textwidth]{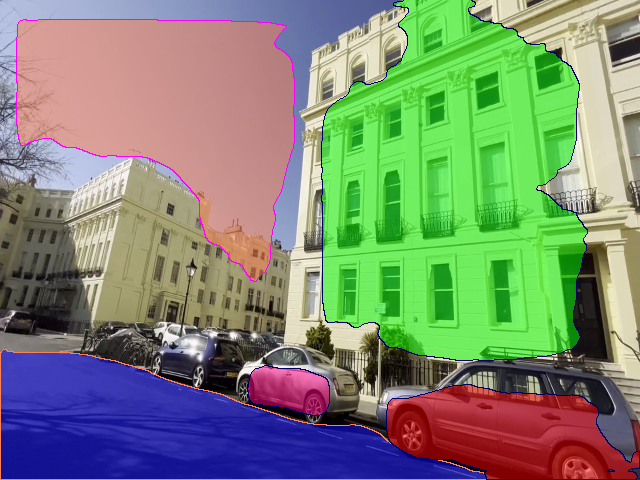}
         \caption{Example frame 2}
         \label{fig:planar_segmentation2}
     \end{subfigure}
     \caption{Example frames of the PlanRCNN planar segmentation CNN applied to Brunswick Square Brighton Data-set}\label{fig:planarseg}
\end{figure}

Planar reconstruction was attempted using PlanaRCNN \cite{Liu2018PlaneRCNN:Imageb} pre-trained on an indoor data-set, an extract of the results of Brunswick Square Brighton are shown in Fig. \ref{fig:planarseg}. The CNN was trained on annotated 3D reconstructions of indoor scenes data-set \cite{Dai2017ScanNet:Scenes}. Although it seems to segment road planes well, it does not generalize well on our outdoor data-set, as can be seen from Fig. \ref{fig:planar_segmentation}. The methodology intended was to determine planes in the input images, determine semantic label of the planes from DeepLab semantic segmentation pipeline trained on CityScapes Data-set, and then to ray trace each point in the SfM model from camera to determine if there are opaque planar occlusions between camera and point, if so, the points must be erroneous.
\par
\begin{figure}
     \centering
     \begin{subfigure}{0.33\textwidth}
         \centering
         \includegraphics[width=\textwidth]{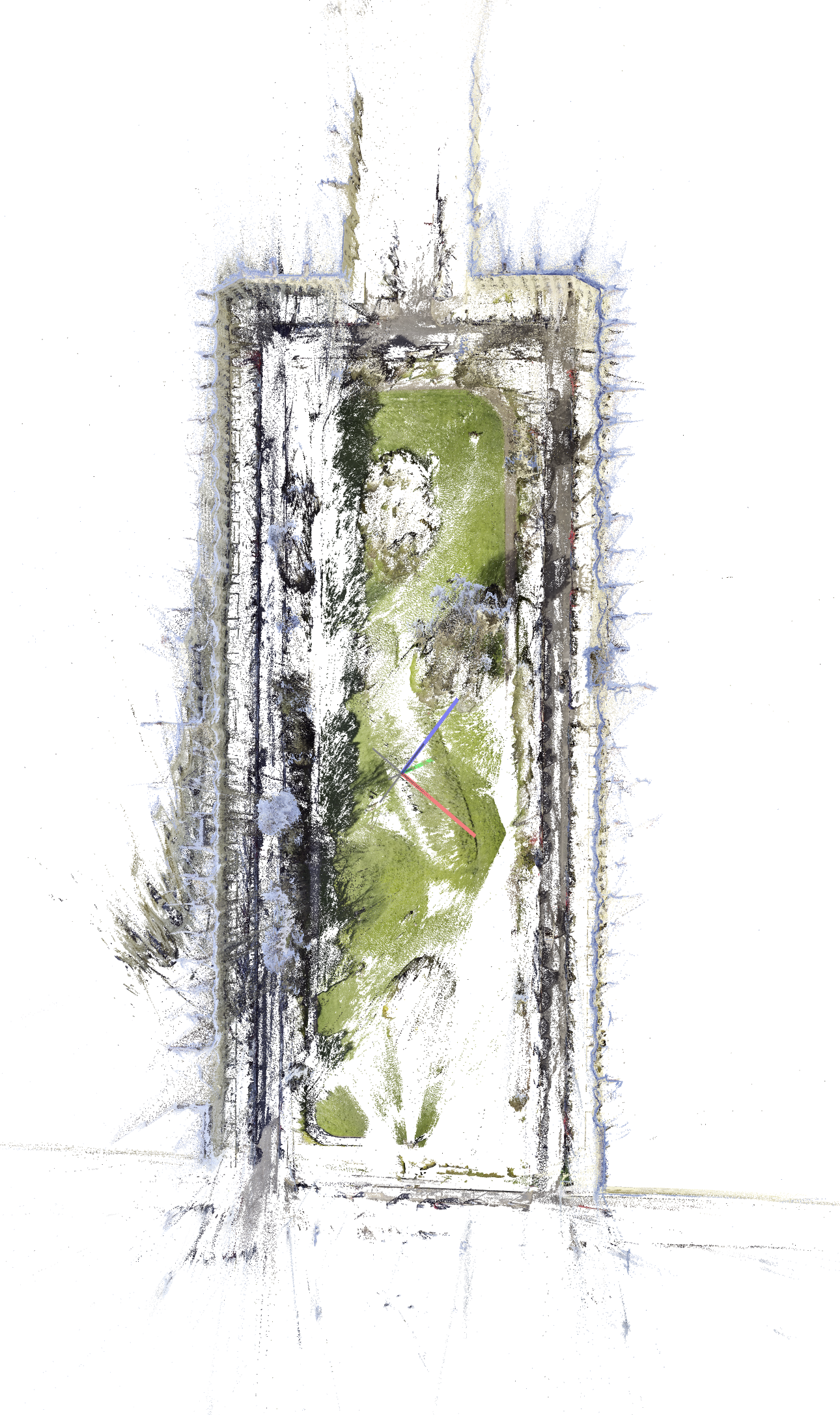}
         \caption{Fused reconstruction}
         \label{fig:fused}
     \end{subfigure}
     \begin{subfigure}{0.33\textwidth}
         \centering
         \includegraphics[width=\textwidth]{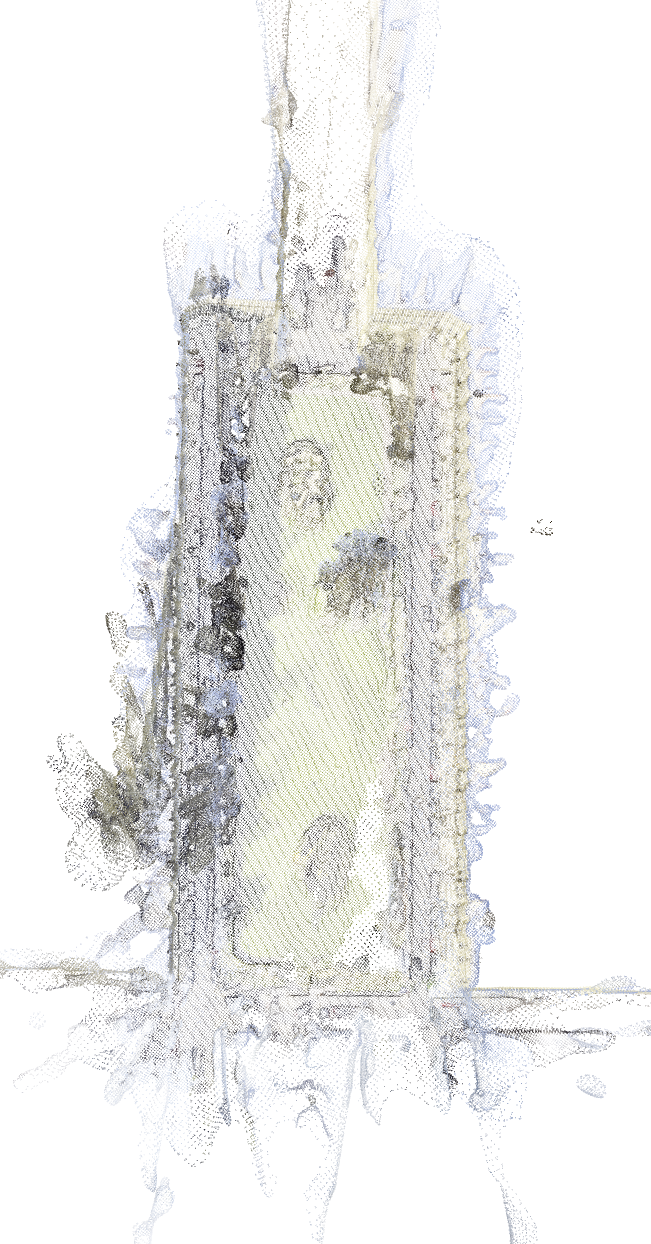}
         \caption{Poisson meshing}
         \label{fig:poisson}
     \end{subfigure}
     \begin{subfigure}{0.3\textwidth}
         \centering
         \includegraphics[width=\textwidth]{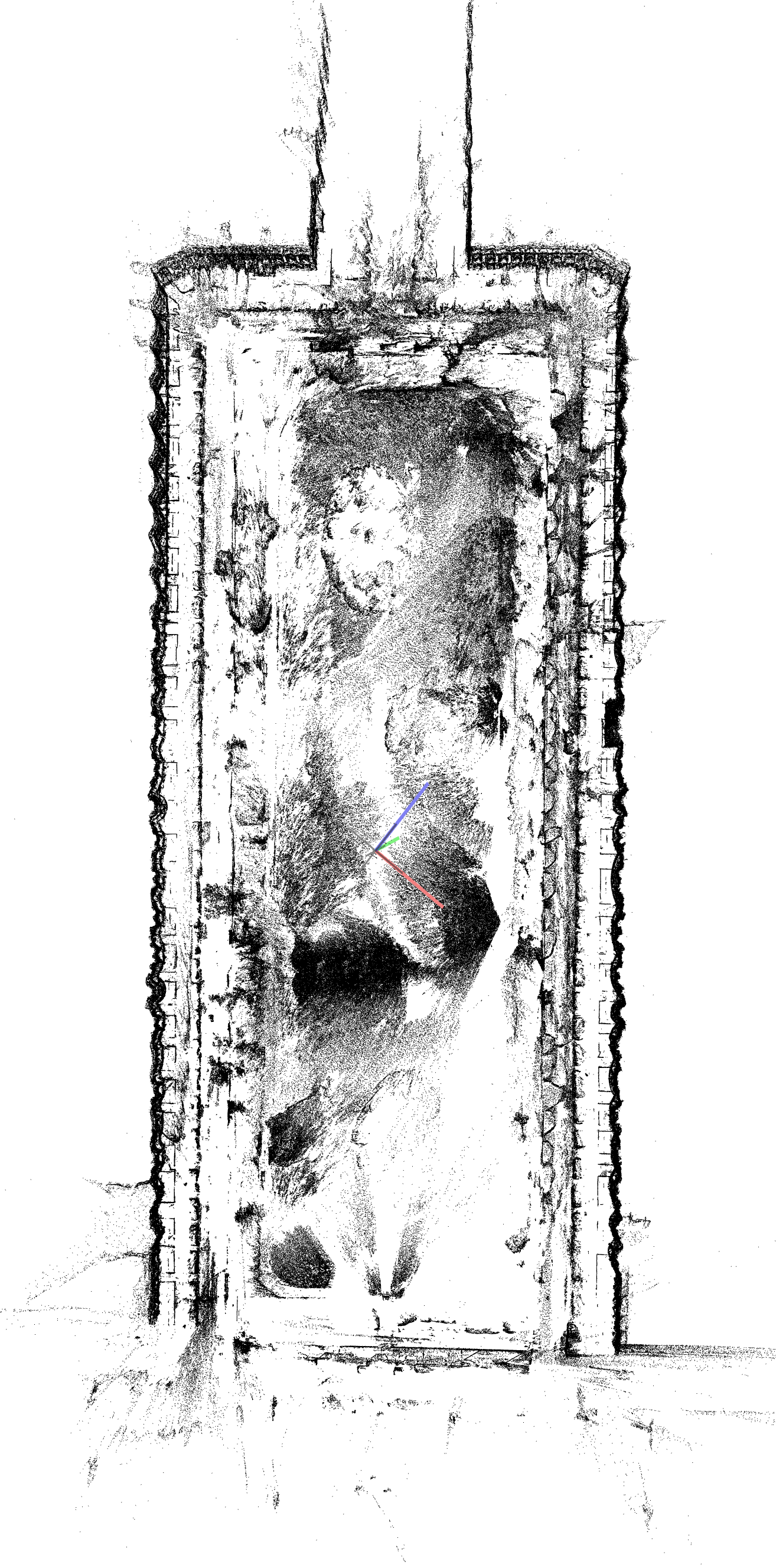}
         \caption{Delaunay triangulation}
         \label{fig:delaunay}
     \end{subfigure}
    \caption[Dense reconstruction results]{Dense reconstructions; Fused dense reconstruction, Poisson meshing \& Delaunay triangulation.}
        \label{fig:poissonDelaunay}
\end{figure}

Poisson meshing and Delaunay triangulation were also performed on the Brunswick Square Data-set, the results of which are shown in Fig. \ref{fig:poissonDelaunay}.
Poisson generated using meshing of the fused point cloud using Poisson surface reconstruction. Delaunay generated using meshing of the reconstructed sparse or dense point cloud using a graph cut on the Delaunay triangulation and visibility voting. Fused meshing is generated by fusion of dense 3D reconstructions and mapping using MVS after running image undistortion into a coloured point cloud.  
Delaunay triangulation is the most robust to outlier points in, and does not fit bad surfaces to outliers.
\chapter{Discussion} \label{Chap5}
%



SfM consists of three main steps, feature detection and matching, camera motion estimation, and recovery of 3D structure from estimated intrinsic and extrinsic parameters and features. The problems often faced in textureless scenes, or scenes with repetitive architecture are erroneous feature matching. In this project, we aimed to improve correctness of SfM models by integrating semantic understanding into SfM.
There is massive variability in point cloud models produced by COLMAP , due to the random element of RANSAC in the SIFT feature matching algorithm. The variability in the COLMAP SfM models, given the same input data necessitates a validation algorithm, so the models can be corrected, or even discarded. The variability in models can be reduced through enforcing constraints such as semantic consistency on matched feature points, as well as the constraint of ray tracing to find points occluded from the camera.
\par
Poisson meshes for surface reconstruction are not suitable for SfM models as they are not robust to outliers, and will fit to outliers, making bad surfaces. Poisson surface reconstruction can be used to mesh unstructured point clouds, however as this technique generates a mesh model by estimating the surface using surface normals of input points, an accurate surface normal is required, thus erroneous points can cause bad surfaces. Delaunay triangulation however, is suitable for surface reconstruction as it is a technique more robust to outliers.
\par
The semantic segmentation model DeepLab was pre-trained on the CityScapes data-set, which was trained on 50 cities, over several months, during daytime. Outdoor, camera-only localisation and semantic segmentation is difficult, due to the complex changes in the surrounding environment's appearance, such as those caused by changes in lighting, season, weather, and scene structure. The CityScapes data-set has a large number of dynamic objects and varying scene background and layout, thus is well suited for transfer learning. However, as the model is only trained on data acquired during daytime, a high dynamic range in the image data input to the pre-trained DeepLab model could decrease the IoU. As segmentation algorithms for road scenes based on deep learning are generally very dependant on the volume of images with pixel level annotations, and there is a scarcity of large scale nighttime data, the CityScapes data-set was determined to be the most fitting. Furthermore, SfM performs best in a data-set without extreme appearance change in terms of lighting, season, weather and scene structure.
A large assumption made in this project was that the semantic segmentation is correct, as the semantic labels are used to (in)validate the model. Due to the inherent working of the structure from motion algorithm; many features are identified on the edges of things, such as the edges of buildings. In the case of semantic segmentation, the edges of identified classes are where there is generally some noise, thus there is some assumption that these errors do not exist. The constraint violations are that the semantic labels are consistent between two view geometries, and if those two views are of the edge of the semantic class, the accuracy of the resulting model may not be ideal.

When creating the dense reconstruction, Poisson surface reconstruction was found to be impractical, as this method is not robust to outliers, and fit to outliers causing bad surfaces and dense models resembling nothing from this world. However, Delaunay triangulation offers a method more robust to outlier points and allows for dense reconstructions with more complete surface output.

An aspect of this project focused on ray tracing from camera location to 3D point, and determining if there's a planar occlusion between the point labelled as an opaque class such as wall or building. As a validation model, there is no assumption that either the plane, point, or camera are correct, as validation only determines if there is a constraint violation. However, if the planar intersection was used during the optimisation of the two view geometries, then the camera pose would be need to be assumed correct.
This project used semantic consistency to post-process the sparse reconstruction of the 3D model, by determining the most common semantic label in the observations in 2D, and discarding any observations different for each point. This aspect of semantic consistency acts as a correction of the 3D model SIFT feature matching, after the fact.
An interesting future work could be to integrate the semantic consistency into the feature matching optimisation, and would dramatically increase computation time.

\par

Windows are a common transparent object in daily life, however due to to their unique physical characteristics, they are often problematic in SfM. Due to their transparent and reflective nature, current sensors struggle to detect transparent obstacles \cite{Cui2021RecognitionLidar}. In particular, transparent materials do not follow the classical laws of geometric light paths used in SfM, this can cause mapping to be performed in through, behind or a reflection in the window. Specularities can also be identified as features and mapped erroneously. A limitation in the system created, is that the semantic segmentation model does not have a windows class, and cannot recognise windows from walls or buildings. Thus, any points mapped in through, or behind the window would be flagged as erroneous during the ray tracing from camera to point, planar intersection validation step. 

\section{Future Work}
In comparison to the large volume of research into static SfM, 3D reconstructions of dynamic scenes has thus far been investigated on smaller scale, shorter image sequences.
The integration of semantic segmentation into SfM can allow for identification of dynamic objects in 3D modelling (such as cars, people or clouds). In mapping in procedures such as SfM or SLAM, the solved problem revolves around an assumption of a static environment \cite{Sualeh2021SemanticsMODT}, something that is far from real-world. Semantic awareness in SfM and dynamic SLAM mapping can allow for improved mapping performance with fewer errors \cite{Yu2018DS-SLAM:Environments}. 
An interesting application of this could be the removal of semantically labelled dynamic objects in SLAM, enabling autonomous high level navigation tasks and obtain simple cognition and reasoning abilities. The removal of, or even just identification of dynamic objects can allow for robust mapping and localisation, as they can be treated as temporally inconsistent occlusions, rather than static features. This method also offers itself well towards the detection of change in scenes, where the problem is challenging as large changes make point-level correspondence establishment difficult, which in turn breaks the assumptions of vanilla SfM \cite{SGPUCL}.
Although the main impediment in the effective fusion of semantic understanding and SfM/SLAM mapping is the vast computational demands and time. At system level in SfM, the bottleneck is the global bundle adjustment of all past frames, a possible solution could be to limit the periodic bundle adjustments to a short time window, similar to the working of a visual SLAM system \cite{Ozden2010MultibodyPractice}.

To improve upon the existing system, pixel level semantic labelled data-set of urban scenes including a class of window should be acquired. Current methods of identification of glass obstacles are base on frame detection, and Semantic SfM would benefit from the recognition of transparent objects. To improve upon the CityScapes data-set and make the semantic segmentation DeepLab model more robust, the data-set should also be acquired in different image conditions: all seasons, and varying lighting. Enforcing label consistency across the matches through training on a greater dynamic range can improve robustness to seasonal changes and improve  object recognition IoU \cite{Larsson2019ASegmentation}. 
\chapter{Conclusion} \label{Chap6}
This research aimed to improve upon the existing SfM modelling correctness through correcting the semantically inconsistent 3D points, and leveraging the semantic and geometric properties of the objects recognised in the scene. Identifying points occluded by opaque structure such as  brick walls, allows for a validation model that can flag the point as erroneous if constraint violated. 
The fusion of semantic segmentation DCNN pipeline DeepLab was utilised to validate and correct COLMAP SfM models through semantic consistency, and ray tracing to show opaque occlusions.
This research has shown that SfM can benefit from the integration of semantic segmentation, to validate 3D models. The great inconsistency between SfM COLMAP results given the same input images, only further necessitates an autonomous validation pipeline. In addition, the fused methods provides rich semantic information that can be readily integrated into SfM.
\renewcommand{\bibname}{References}
\bibliographystyle{ieeetr}
\bibliography{Bibliography.bib}
\begin{appendices}
\chapter{Source Code} \label{System Requirements}
Source code for all of the methods implemented in Chap. \ref{Chap3} for the project can be found in the GitHub repository: \newline 
\url{https://github.com/joerowelll/COMP0132_RJXZ25.git}.\newline
Installation instructions for the C++17 SfM and MVS Pipeline COLMAP can be found here \cite{Schonberger2016Structure-from-MotionRevisited, AgarwalCeresLibrary}. \\
Project files can be found on Google Drive \url{https://drive.google.com/drive/folders/1CNxIw8gyTOldooBWsJqVdKNtusy9eF5a?usp=sharing}

\chapter{Project Introduction Video}\label{sec:projectIntroduction}
A short video presentation, introducing background, aims and organisation of the project, as of 30$^{th}$ June 2022: \newline
\url{https://youtu.be/hrHsb8gOGck}
\\
\\
\\
\begin{centering}
\href{https://youtu.be/hrHsb8gOGck?t=5}{\includegraphics[width = \textwidth]{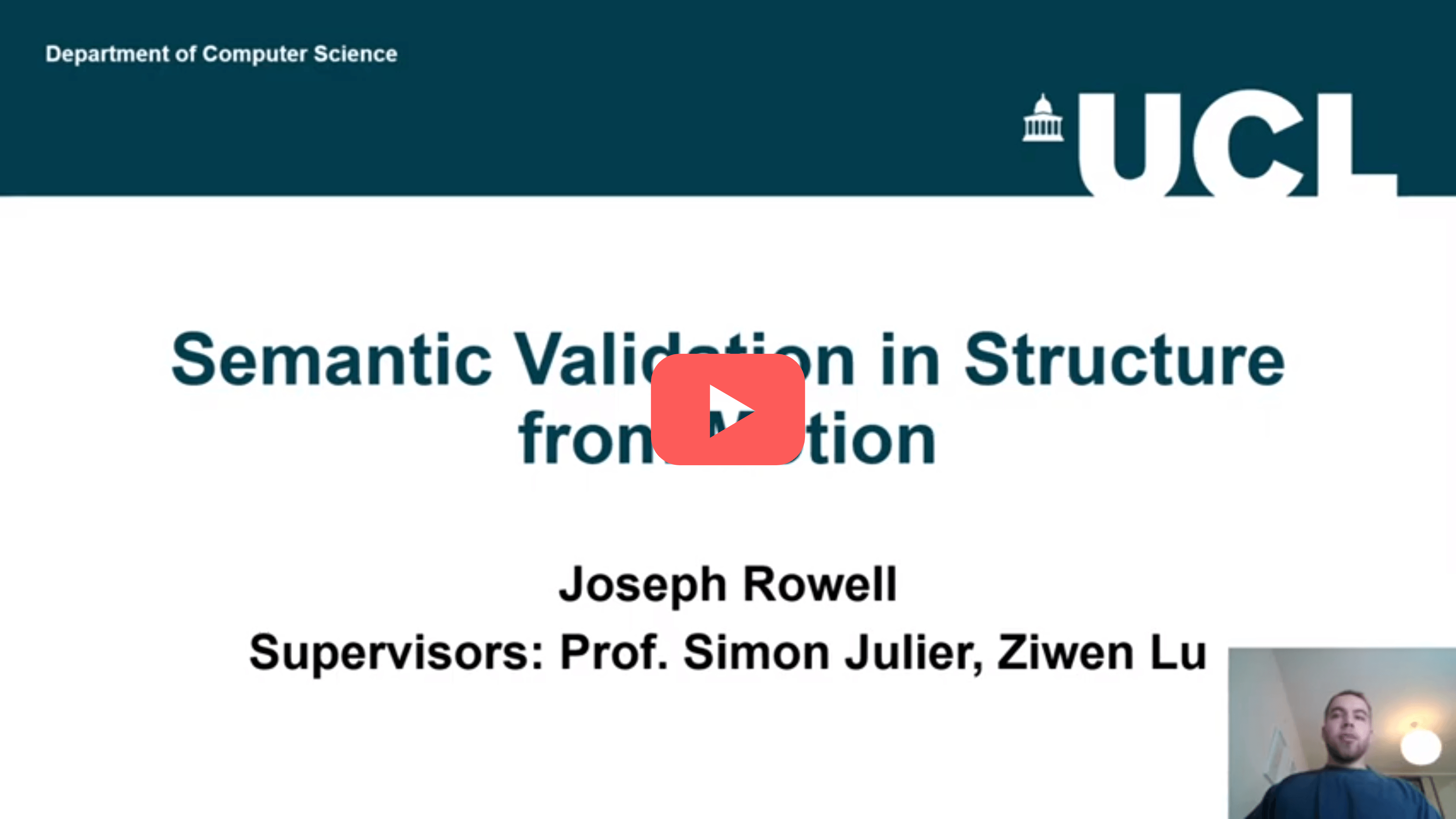}}
\end{centering}

\chapter{Semantically Labelled 3D Points}
An extract example of the comma separated values file showing semantically labelled 3D points is shown in Table \ref{tab:labelled3DPoints}. In this table, the "imageid" is the ordered index of the images input to the SfM and semantic segmentation pipeline. The "X2D", "Y2D", "X3D", "Y3D", "Z3D" are the coordinates of the two view geometry SIFT keypoints in the 2D input image and 3D output  model respectively. The "INTENSITY" is the pixel intensity in the DeepLab semantic segmentation output of the image, and the "SEMANTIC\_LABEL" is the label of the keypoint, at the given location, given by DeepLab semantic segmentation algorithm. These values were used to determine semantic consistency in the SfM model.

\begin{table}[H]
\begin{center}
\caption{\centering Semantically Labelled 3D Points \texttt{.csv} File Example Extract. }\label{tab:labelled3DPoints}
\begin{tabular}{ c c c c c c c c}
\hline \hline
\textbf{imageid}	& \textbf{X2D} &	\textbf{Y2D}	& \textbf{X3D} &\textbf{Y3D} & \textbf{Z3D} & \textbf{INTENSITY} & \textbf{SEMANTIC\_LABEL} \\
\hline
281	& 180.35 &	297.59 &	-0.036	 
& -0.33 &	-0.036	&2	&building \\
\hdashline
134	&252.37	& 311.10 &	0.10	&-0.30&	0.16&	3&	sky \\
\hdashline
1045	&138.74&	313.63	&-0.09	&-0.28	&-0.09&	4	& car\\
\hdashline
796	&240.27	&316.62&	0.08	&-0.28	&0.08	&1	&road \\
\hdashline
763&	105.79&	319.20	&-0.14&	-0.26 &	-0.14	&6	&foliage \\
\hdashline
415 &	156.90&321.13&	-0.06&	-0.26	&-0.06&	9	&dynamic \\
$\vdots$ & $\vdots$ & $\vdots$& $\vdots$& $\vdots$& $\vdots$& $\vdots$& $\vdots$\\
\hline
\end{tabular}
\end{center}
\end{table}
\end{appendices}
\clearpage


\end{document}